\RequirePackage{fix-cm}
\documentclass[smallextended]{svjour3}       
\smartqed  
\usepackage{graphicx}
\usepackage{subfigure}
\usepackage{amsfonts}
\usepackage{bm}
\usepackage{amsmath}
\usepackage{algorithm}
\usepackage{algorithmic,color}

\usepackage[numbers,sort&compress]{natbib}

\begin{document}

\title{
 A Registration-Based Star-Shape Segmentation Model and Fast Algorithms
  \thanks{Daoping Zhang is supported by National Natural Science Foundation of China (Nos. 12201320 and 12471484) and the
Fundamental Research Funds for the Central Universities, Nankai University (Nos. 63221039 and 63231144). Xue-Cheng Tai is supported by NORCE Kompetanseoppbygging program. Lok Ming Lui is supported by HKRGC GRF (Project ID: 14307622). }
}


\author{Daoping Zhang         \and
        Xue-Cheng Tai \and Lok Ming Lui 
}


\institute{Daoping Zhang \at
              School of Mathematical Sciences and LPMC, Nankai University, Tianjin 300071, People's Republic of China \\
              \email{daopingzhang@nankai.edu.cn}  
           \and
           Xue-Cheng Tai\at
              Norwegian Research Centre, Norway \\
              \email{xuechengtai@gmail.com}
              \and
           Lok Ming Lui \at
              Department of Mathematics , The Chinese Univerisity of Hong Kong, Shatin, NT, Hong Kong \\
              \email{lmlui@math.cuhk.edu.hk}
}

\date{Received: date / Accepted: date}

\maketitle

\begin{abstract}
Image segmentation plays a crucial role in extracting objects of interest and identifying their boundaries within an image. However, accurate segmentation becomes challenging  when dealing with occlusions, obscurities, or noise in corrupted images. To tackle this challenge, prior information is often utilized, with recent attention on  star-shape priors. In this paper, we propose a star-shape segmentation model based on the registration framework. By combining the level set representation with the registration framework and imposing constraints on the deformed level set function, our model enables both full and partial star-shape segmentation, accommodating single or multiple centers. Additionally, our approach allows for the enforcement of identified boundaries to pass through specified landmark locations. We tackle the proposed models using the alternating direction method of multipliers. Through numerical experiments conducted on synthetic and real images, we demonstrate the efficacy of our approach in achieving accurate star-shape segmentation.

\end{abstract}


\section{Introduction}

Image segmentation involves dividing a target image into multiple parts and extracting meaningful objects. As a fundamental task in image processing, image segmentation plays a pivotal role in applications like face recognition, object tracking, autonomous vehicles, and medical imaging \cite{elnakib2011medical,gould2009region,kamencay2013novel,kaymak2019brief,zaitoun2015survey}.

Traditional image segmentation methods encompass threshold-based methods \cite{cai2013two}, region growing methods \cite{pohle2001segmentation,tremeau1997region}, watershed methods \cite{hodneland2009four,mangan1999partitioning,tai2007level}, 
wavelet-based methods \cite{gavlasova2006wavelet}, garph-cut methods \cite{yi2012image}, max-flow and STD methods \cite{liu2022deep,yuan2014spatially}, and variational and PDE-based methods \cite{chan2007some}. In recent decades, the variational and PDE-based methods have gained significant attention from researchers due to the development of rigorous mathematical theories and highly efficient numerical algorithms. Therefore, this paper specifically concentrates on exploring variational and PDE-based methods.

The variational and PDE-based methods are generally categorized into two groups: edge-based and region-based methods. Among the notable edge-based techniques, Snakes \cite{kass1988snakes} stands out. It involves deforming a parametric curve using internal and external forces to identify boundaries of target objects. However, this approach is non-intrinsic as the cost functional depends on how the curve is parametrized and does not directly relate to the geometry of objects. Addressing this limitation, geodesic active contours \cite{caselles1997geodesic} were introduced to evolve active contours based on intrinsic geometric measures of the image, allowing for natural topology changes. To enhance the model's robustness to noise, \cite{li2007active} evolved active contours by introducing vector field convolutions as a new external force. On the other hand, in the realm of region-based methods, the Mumford-Shah model \cite{mumford1989optimal} is a prominent representative. It aims to approximate a given image by computing optimal piecewise-smooth or piecewise-constant functions. Addressing the piecewise-constant Mumford-Shah model, Chan and Vese \cite{chan2002active} utilized the level set method, representing object boundaries using the level set function. Furthermore, various works have extended the Chan-Vese model to handle diverse tasks, such as vector-valued images \cite{chan2000active}, textured images \cite{chan2002active}, and the segmentation of multiple objects \cite{lie2006variant,tai2007image}.

While methods mentioned above have achieved considerable success in various applications, they encounter significant challenges when dealing with corrupted, obscured, overexposed, or underexposed images \cite{siu2020image,zhang2021topology,zhang2021topology2}. To address these challenges, incorporating prior information about target objects into the segmentation process is a natural and effective approach. Cremers et al. proposed a variational approach that combined a level set formulation of the Mumford-Shah functional with shape priors to enhance segmentation accuracy \cite{cremers2003towards}. They extended their work by introducing a labeling function to identify regions where the shape prior was applied \cite{cremers2003towards}. Chan and Zhu also contributed to this area by introducing a labeling level set function to delineate regions where the prior shape should be considered for comparison \cite{chan2005level}. In the realm of segmenting multiple objects, \cite{thiruvenkadam2008segmentation} proposed a variational energy model that incorporated prior knowledge of object shapes. To ensure shape compactness, \cite{gui2017medical} innovatively combined the isoperimetric constraint with the level set method, presenting a novel segmentation model. These approaches signify a significant step forward in addressing the problem posed by imperfect or challenging image conditions.

In our daily lives, objects typically exhibit regular shapes, often displaying convex or star-like forms. {\color{black} In geometric terms, a convex domain refers to any shape where the straight line segment between any two points always remains entirely within the shape. This concept can be generalized to star-like (or star-shape) domains, which only require that every point in the shape can be connected to a fixed central point by a straight line segment that never exits the shape. Notably, star-like domains form a more inclusive category that encompasses various non-convex shapes.} Achieving an effective segmentation based on these shapes necessitates meticulous design of models and solvers to meet both theoretical and practical constraints. For convexity segmentation, \cite{strekalovskiy2011generalized} ensured a convex $n$-polygon by introducing at least $n \geq 3$ auxiliary surrounding regions. However, achieving a smooth object segmentation may require setting a larger $n$, potentially compromising algorithm efficiency. Alternatively, in \cite{gorelick2016convexity}, achieving the convexity relied on penalizing 1-0-1 configurations along all straight lines, represented as a sum of three-clique potentials. Leveraging the observation that the corresponding region of a convex level set function must be convex, 
\cite{yan2020convexity} utilized this function to represent convexity shape priors, extending it to multi-object segmentation \cite{luo2022convex,luo2019convex}. \cite{luo2020level} also proposed an efficient method for the convex-shape representation applicable across various object dimensions. Moreover, Chen et al. introduced convex geodesic models employing an orientation-lifting strategy, encoding specific curvature constraints to ensure convexity shape priors \cite{chen2022geodesic}. In the star-shape segmentation, \cite{veksler2008star} introduced a star-shape constraint within the graph-cut framework. Subsequently, Yuan et al. incorporated a variational constraint to uphold the star shape prior \cite{yuan2012efficient}. More recently, Liu et al. integrated the variational constraint of the star-shape prior with deep convolutional neural networks \cite{liu2022deep}. It is noteworthy that we can replace the graph-cut model by continuous max-flow and min-cut approaches as shown in  \cite{pock2008convex,Bae2010a,Bae2013a,Yuan2010,Yuan2010a},
which can further remove  metrication errors associated with boundary length regularization. 

The registration-based segmentation method, which involves performing image segmentation through image registration, offers a fresh perspective for exploring image segmentation \cite{le2011combined}. The essence of this approach lies in establishing a connection between boundaries of target objects in the reference and input image through the resultant transformation. This framework facilitates topology-preserving segmentation when the transformation is one-to-one. It can also be adapted for specific applications by incorporating constraints into the transformation \cite{siu2020image,zhang2021topology2}. Topology-preserving segmentation was assured in \cite{chan2018topology} through the Beltrami representation of a shape. Subsequently, \cite{siu2020image} extended the registration-based segmentation model proposed in \cite{chan2018topology} to achieve the convexity-preserving segmentation. This extension involved designing a dedicated convexity constraint based on the discrete conformality structures of the image mesh. However, it is important to note that models based on the Beltrami representation are limited to the 2D segmentation due to the representation's definition in complex space. To extend the registration-based segmentation into the 3D domain, Zhang and Lui combined a hyperelastic regularizer with the fitting term in the Chan-Vese model \cite{zhang2021topology}. Additionally, Zhang et al. introduced a 3D convex segmentation model based on the registration by incorporating the level set function to represent the convexity shape prior \cite{zhang2021topology2}. Nevertheless, as of our current knowledge, the star-shape segmentation model based on the registration framework remains unavailable.

Therefore, this paper focuses on generalizing the registration-based star-shape segmentation model. The main contributions of this paper are listed as follows:
\begin{itemize}
\item By combing the level set function with the registration framework and imposing constraints on the deformed level set function, we propose the star-shape segmentation model. We further extend the star-shape segmentation with respect to one center to the star-shape segmentation with respect to multiple centers, partial star-shape segmentation, and selective star-shape segmentation. In addition, by the advantage of the registration framework, we can also force the identified boundaries to pass through some specific landmark  locations.
\item We utilize the alternating direction method of multipliers (ADMM) to devise an efficient algorithm for solving the proposed model. Here, one subproblem is solved by the modified Newton method and the other has a closed-form solution.
\item Numerical tests demonstrate that our models accurately achieve the star-shape preservation in the segmentation, both for synthetic and real data.
\end{itemize}

The subsequent sections of this paper are structured as follows. In Section \ref{Pre}, we commence by reviewing pertinent preliminaries. Moving on to Section \ref{Model}, we introduce our registration-based segmentation model, designed to preserve star-shape features. Sections \ref{Alg} and \ref{Result} present the algorithm for our proposed models and showcase the results of numerical experiments, respectively. Finally, Section \ref{Con} concludes this paper.


\section{Preliminaries}\label{Pre} 

In this section, we comprehensively review key preliminaries for our paper, {\color{black} notably the level set function, the concept of a star-shape domain, image segmentation, image registration and  registration-based segmentation method}. These foundational elements set the stage for the subsequent discussions and applications in our work.

\subsection{Level set function and Star-shape domain}

\begin{definition}[Level set function, LSF]
Let $\mathbb{D}$ be a open subset of $\mathbb{R}^{d}$ and $\partial \mathbb{D}$ be its smooth boundary. A Lipschitz continuous function $\phi(\bm{x}): \mathbb{R}^{d}\rightarrow\mathbb{R}$ is called a level set function of the region $\mathbb{D}$ if it satisfies the following conditions:
\begin{equation*}
\left\{
\begin{split}
&\phi(\bm{x})>0 \quad \mathrm{if}\ \bm{x}\in\mathbb{D},\\
&\phi(\bm{x})=0 \quad \mathrm{if}\ \bm{x}\in\partial\mathbb{D},\\
&\phi(\bm{x})<0 \quad \mathrm{if}\ \bm{x} \in \bar{\mathbb{D}}^{c}, \\
\end{split}\right.
\end{equation*}
where $d$ is the dimension of the space and $\bm{x} = (x_{1},\cdots,x_{d})$ is the coordinate. In this paper, we focus on the case of $d=2$.
\end{definition}

\begin{remark}
It is important to note that a region $\mathbb{D}$ can have multiple corresponding level set functions. For instance, considering $B_{r}(\bm{x})$ as a ball centered at the origin with radius $r>0$, both $r^{2}-x_{1}^{2}-\cdots-x_{d}^{2}$ and $r-\sqrt{x_{1}^{2}+\cdots+x_{d}^{2}}$ serve as valid level set functions for this region.
\end{remark}

\begin{definition}[Star-shape domain]\label{Star_Shape_Def}
A domain $\mathbb{D}\subset\mathbb{R}^{d}$ is classified as a star-shape domain if there exists  a point $\bm{c}\in \mathbb{D}$, the line segment connecting any point $\bm{x}\in \mathbb{D}$ to $\bm{c}$ lies entirely within the domain $\mathbb{D}$. In this context, $\bm{c}$ is referred to as the center point.

Alternatively, in a more formal mathematical definition, let $\mathbb{D}\subset\mathbb{R}^{d}$ be a bounded closed domain with a simple closed boundary curve $\partial\mathbb{D}$. If for a  given center point $\bm{c}\in\mathbb{D}$, the domain $\mathbb{D}$ is such that $\{\bm{y}|\bm{y} = (1-\lambda)\bm{x}+\lambda\bm{c},\lambda\in (0,1)\}\subset\mathbb{D}$ for any $\bm{x}\in \partial\mathbb{D}$, then $\mathbb{D}$ is considered as a star-shape domain with respect to the center point $\bm{c}$.
\end{definition}

\begin{figure}[hthp]
\centering
\subfigure[]{
\includegraphics[width=1.4in,height=1.1in]{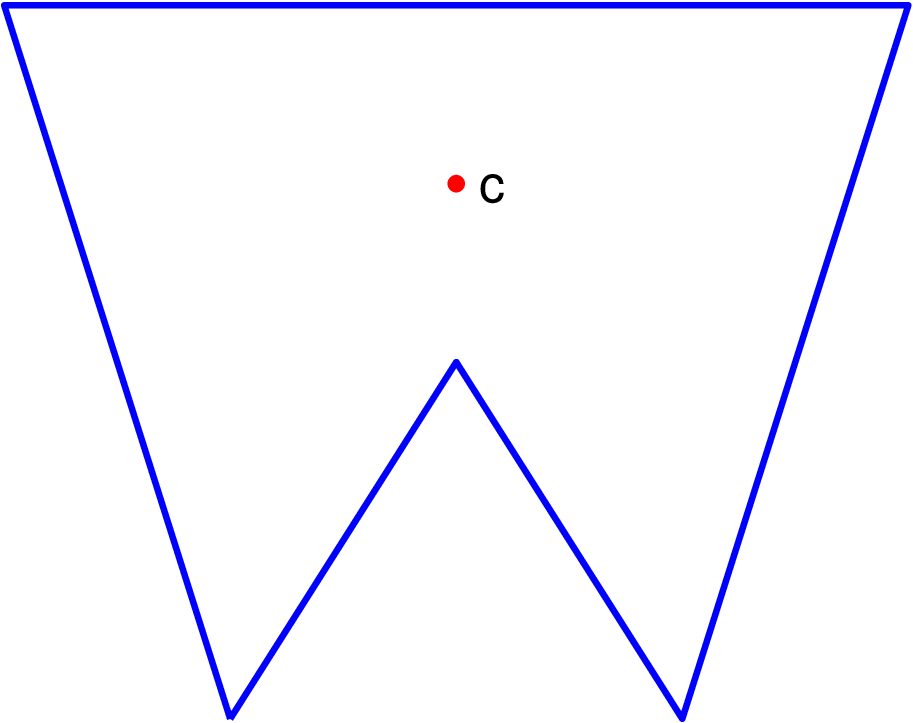}
\includegraphics[width=1.4in,height=1.1in]{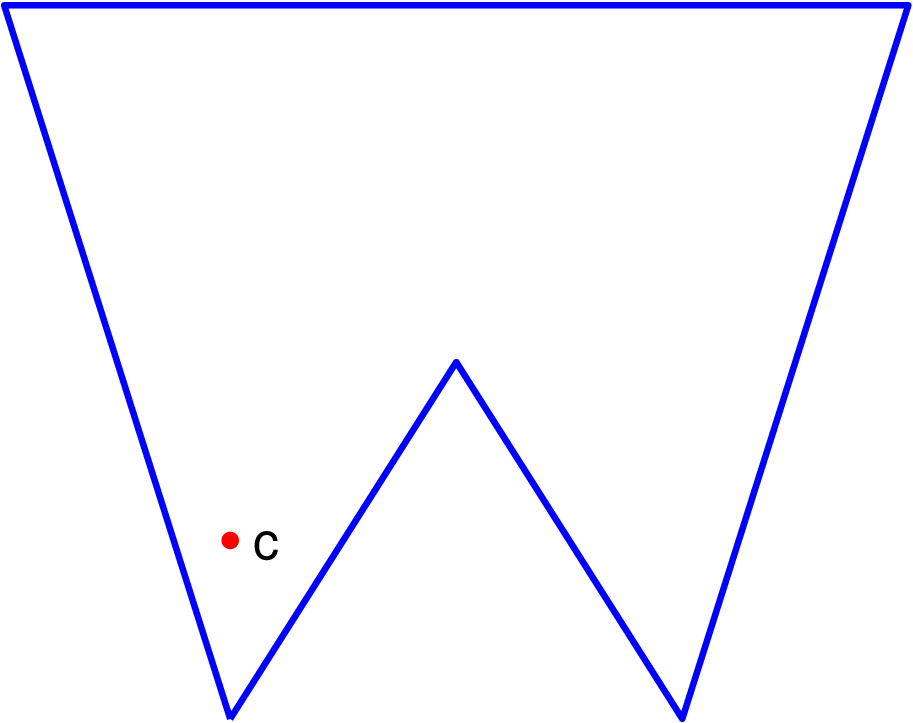}}
\subfigure[]{
\includegraphics[width=1.4in,height=1.1in]{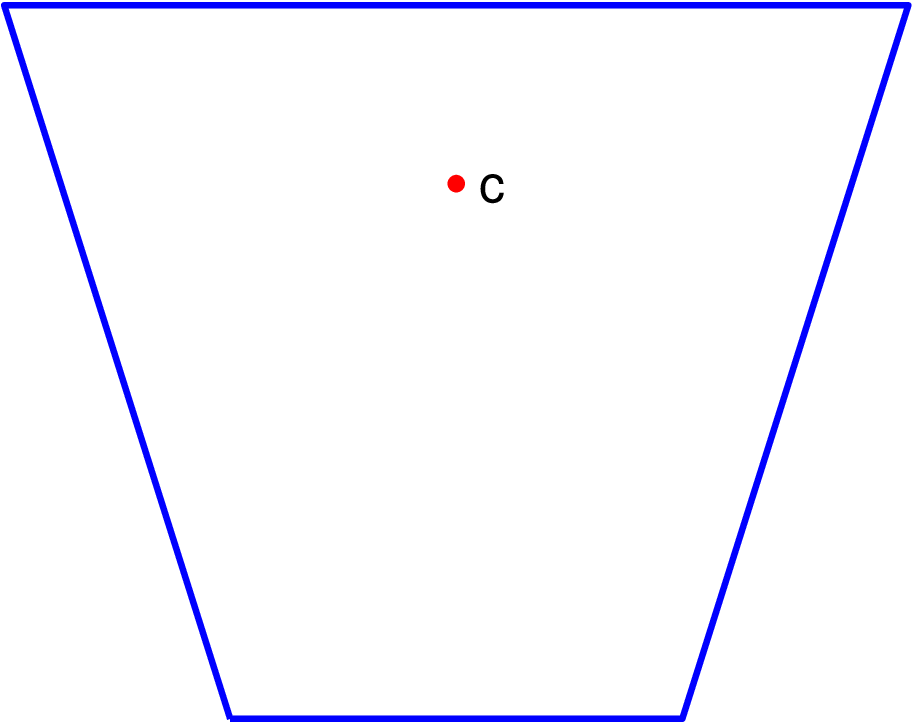}
\includegraphics[width=1.4in,height=1.1in]{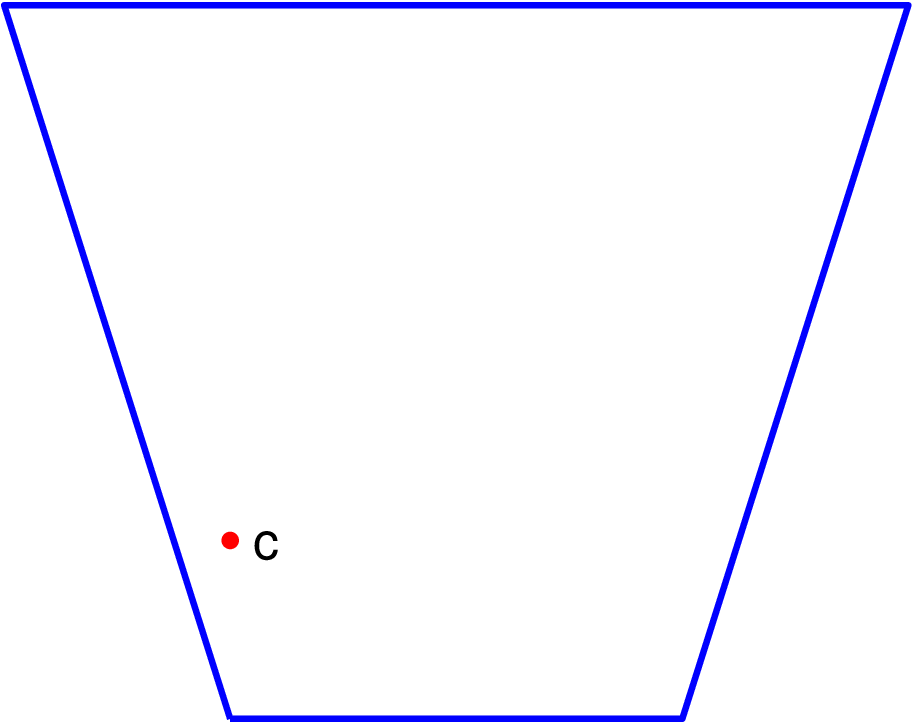}} 
\caption{The nature of a star-shape domain is contingent upon the placement of the center point $\bm{c}$. (a) The left diagram illustrates a star-shape domain with center point $\bm{c}$. In contrast, the right diagram  does not qualify as a star-shape domain with  alternative center point $\bm{c}$. 
(b) In the case of a convex domain, it inherently qualifies as a star-shape domain regardless of the location of the center point $\bm{c}$,  as long as it is inside the domain. }\label{fig_star_shape_domain}
\end{figure}

According to Definition \ref{Star_Shape_Def}, the configuration of a star-shape domain is influenced by  the center point $\bm{c}$ (Fig. \ref{fig_star_shape_domain}(a)). Additionally, if a domain is convex, it is inherently a star-shape domain, regardless of the location of the center point $\bm{c}$ (Fig. \ref{fig_star_shape_domain}(b)). From this perspective, we can consider the star-shape property as a generalization of convexity.
Moreover, observe that a domain satisfying the condition where any radial line from the center point $\bm{c}$ to any $\bm{x} \in \mathbb{D}$ intersects the boundary only once is a star-shape domain. Motivated by this characteristic, Veksler introduced a pairwise star-shape constraint term in \cite{veksler2008star} that is trying to use the integral of the following
penalty function to force the represented object to be star-shape:

\begin{equation}\label{pairwise_star_shape_cons}
\mathcal{P}_{0}(u(\bm{x}),u(\bm{y})) = 
\left\{
\begin{split}
0,  \quad \quad\ \ \ &\mbox{if}\ u(\bm{y}) = u(\bm{x}),\\
\beta\geq 0,        \quad\ &\mbox{if}\ u(\bm{y}) = 1\ \mbox{and}\ u(\bm{x}) = 0,\\
+\infty, \quad\quad &\mbox{if}\ u(\bm{y}) = 0\ \mbox{and}\ u(\bm{x}) = 1.
\end{split}\right.
\end{equation}
Here, $u: \Omega\rightarrow\{0,1\}$ is the indicator function of $\mathbb{D}$ which takes value 1 inside $\mathbb{D}$ and 0 outside. 
When minimizing $\mathcal{P}_0$, the neighborhood structure of $u$ with $1-0-1$ will be avoided. Especially, for any point $\bm{x}$ within the object, if $\bm{x}$ is assigned label $1$, then every point $\bm{y}$ on the line from the center point $\bm{c}$ to $\bm{x}$ must also be assigned label $1$. Thus force the domain $\mathbb{D}$ be star-shape.

The pairwise star-shape constraint term \eqref{pairwise_star_shape_cons} operates within a spatially discrete framework as was done in \cite{veksler2008star}. To extend this concept to ensure that a spatially continuous model produces a star-shape domain, the variational formulation for the star-shape prior was proposed in \cite{yuan2012efficient,Yuan2012b}. This formulation shows that the following constraint is enough to guarantee star-shape for a domain $\mathbb{D}$:
\begin{equation}\label{varia_cons}
\langle \nabla u(\bm{x}), e(\bm{x})\rangle \leq 0, \quad \forall \bm{x}\in \Omega,
\end{equation}
where $u$ is the indicator function of $\mathbb{D}$  and $e(\bm{x}) = \nabla d_{\bm{c}}(\bm{x})$, $d_{\bm{c}}(\bm{x}) = |\bm{x}-\bm{c}|^{2}$ denotes the distance map from the center point $\bm{c}$, and $\langle\cdot,\cdot \rangle$ signifies the inner product operation. This equation ensures that for all points $\bm{x}$ within the domain $\Omega$, the inner product between the gradient of the function $u(\bm{x})$ and the gradient of the distance map $d_{\bm{c}}(\bm{x})$ is nonpositive.

In a recent work by Liu et al. detailed in \cite{liu2022deep}, they extended the formulation by permitting the indicator function 
$u(\bm{x})$ in \eqref{varia_cons} to be continuously differentiable level set function (LSF) of $\mathbb{D}$. Subsequently, they provided the following sufficient conditions to ensure that $\mathbb{D}$  is  a star-shape domain.

\begin{theorem}[Sufficient conditions for star-shape domain \cite{liu2022deep}]\label{suff_con_star_shape}
Let $u \in C^{1}(\Omega)$. Assume that its $\gamma$-super level set $u^{-1}[\gamma,+\infty):=\{x\in\Omega | u(\bm{x})\geq \gamma\}$ is closed and has a simple closed boundary curve. In addition, we also assume $\bm{c}=(c_1,c_2)\in u^{-1}[\gamma,+\infty)$ and $\bm{c}\notin \partial u^{-1}[\gamma,+\infty)$. If $u\in \{u | \langle \nabla u(\bm{x}),\bm{x}-\bm{c}\rangle \leq 0, \mbox{a.e.}\ \forall \bm{x}\in \Omega\}$, then we have that any $\gamma$-super level set $u^{-1}[\gamma,+\infty)$ is a star-shape domain with respect to the center point $\bm{c}$.
\end{theorem}

The application of Theorem \ref{suff_con_star_shape} offers a method to utilize the level set function in characterizing star-shape domains. For instance, if $u\in C^{1}(\Omega)$ serves as a level set function for a simply connected domain $\mathbb{D}$ and satisfies $\langle \nabla u(\bm{x}),\bm{x}-\bm{c}\rangle \leq 0,\ \mbox{a.e.}\ \forall \bm{x}\in \Omega$,
then according to Theorem \ref{suff_con_star_shape}, the domain $\mathbb{D}$ is guaranteed to be a  star-shape domain. In this scenario, the parameter $\gamma$ in Theorem \ref{suff_con_star_shape} is set to $0$.

\subsection{Image segmentation}
Image segmentation is a critical process involving the partition of an image into distinct segments to extract relevant objects. In variational settings, segmentation methods typically fall into two categories: edge-based methods, exemplified by works such as \cite{caselles1997geodesic,kass1988snakes,li2007active,roberts2019convex}, and region-based methods, represented by models like \cite{chan2002active,chan2001active,mumford1989optimal,lie2006variant}. In this context, we focus on reviewing a prominent region-based model, namely, the Chan-Vese (CV) model \cite{chan2001active}. The CV model has gained widespread recognition and serves as a pivotal reference in the field of the region-based image segmentation. This review aims to provide a foundational understanding of the CV model, laying the groundwork for the subsequent discussion of the proposed model.

The variational formulation of the CV model \cite{chan2001active} aims to segment the target image $I(\bm{x}): \Omega\subset\mathbb{R}^{d}\rightarrow\mathbb{R}$ into foreground and background segments. The model is expressed as:
\begin{equation}\label{CV_model}
\min_{\Gamma, a_{1}, a_{2}} \mathcal{D}^{\mathrm{CV}}(\Gamma, a_{1}, a_{2})+\mathcal{R}^{CV}(\Gamma),
\end{equation}
\begin{equation*}
\begin{split}
&\mathcal{D}^{\mathrm{CV}}(\Gamma, a_{1}, a_{2}) := \int_{\mathrm{in}(\Gamma)}(I(\bm{x})-a_{1})^{2}\mathrm{d}\bm{x}+\int_{\mathrm{out}(\Gamma)}(I(\bm{x})-a_{2})^{2}\mathrm{d}\bm{x}, \\&\mathcal{R}^{CV}(\Gamma) := \mu\mathrm{Length}(\Gamma),
\end{split}
\end{equation*}
where $\mathcal{D}^{\mathrm{CV}}$ and $\mathcal{R}^{CV}$ are the fidelity term and regularization term, $\Gamma$ is a closed curve, $\mathrm{in}(\Gamma)$ and $\mathrm{out}(\Gamma)$ are regions inside and outside of the curve $\Gamma$, $a_{1}$ and $a_{2}$ are unknown constants to represent mean values of the image $I$ in regions $\mathrm{in}(\Gamma)$ and $\mathrm{out}(\Gamma)$, $\mathrm{Length}(\Gamma)$ is the length of the curve $\Gamma$, and $ \mu$ is a positive parameter to control the balance of these terms. 
This formulation seeks to optimize the curve 
$\Gamma$ and constants $a_{1}$ and $a_{2}$ to achieve an effective segmentation of the image, delineating foreground and background regions.

To solve this model \eqref{CV_model}, in \cite{chan2001active}, Chan and Vese introduced the level set function and Heaviside function to convert \eqref{CV_model} into the following equivalent formulation:
\begin{equation}\label{Equ_CV_model}
\min_{\phi, a_{1},a_{2}} \int_{\Omega}(I(\bm{x})-a_{1})^{2}H(\phi(\bm{x}))\mathrm{d}\bm{x}+\int_{\Omega}(I(\bm{x})-a_{2})^{2}(1-H(\phi(\bm{x})))\mathrm{d}\bm{x} + \mu\int_{\Omega}|\nabla H(\phi(\bm{x}))|\mathrm{d}\bm{x}, 
\end{equation}
where $\phi(\bm{x})$ is a level set function and $H(z) = 1 \ \mathrm{if}\ z\geq 0 \ \mathrm{or}\ 0 \ \mathrm{if}\ z < 0$. To solve \eqref{Equ_CV_model}, an alternating direction method is implemented in \cite{chan2001active}, namely, first fix $\phi(\bm{x})$ to find $a_{1}$ and $a_{2}$ then fix $a_{1}$ and $a_{2}$ to solve $\phi(\bm{x})$. 

\subsection{Image registration}

Image registration involves determining a plausible spatial transformation to deform one image in order to align it with another corresponding image. In the conventional setting, two images, denoted as $T(\bm{x})$ and $R(\bm{x}):\Omega\subset \mathbb{R}^{d} \rightarrow \mathbb{R}$, are provided. Here, $T(\bm{x})$ and $R(\bm{x})$ are referred to as the template and reference, respectively. The primary objective of image registration is to derive a transformation function $\bm{y}(\bm{x}):\mathbb{R}^{d} \rightarrow \mathbb{R}^{d}$ such that the deformed template $T(\bm{y}(\bm{x}))$  closely resembles with the reference $R(\bm{x})$ in some meaningful way. Consequently, the variational formulation for image registration is constructed as follows:
\begin{equation}\label{Registration_framework}
\min_{\bm{y}} \mathcal{D}(T\circ\bm{y}(\bm{x}),R)+\alpha\mathcal{R}(\bm{y}),
\end{equation}
where $\mathcal{D}$ represents a fidelity term, $\mathcal{R}$ serves as a regularizer to ensure the well-posedness of the problem, and $\alpha>0$ is a positive parameter used to balance the weighting between the two terms.

In various applications, a diverse range of fidelity terms \cite{haber2006intensity,haber2007intensity,maes1997multimodality,modersitzki2009fair} and regularizers \cite{broit1981optimal,burger2013hyperelastic,christensen1996deformable,droske2004variational,fischer2002fast,fischer2003curvature,lee2016landmark,zhang2018novel,zhang20203d,zhang2015variational} have been proposed for image registration. In the context of this paper, the focus is primarily on the sum of squared differences (SSD)  \cite{modersitzki2009fair} and the diffusion regularizer \cite{fischer2002fast}:
\begin{equation*}
\mathcal{D}^{\mathrm{SSD}}(T\circ\bm{y},R):= \frac{1}{2}\int_{\Omega}(T(\bm{y}(\bm{x}))-R(\bm{x}))^2\mathrm{d}\bm{x}, \ \mathcal{R}^{\mathrm{Diff}}(\bm{y}):=\int_{\Omega}|\nabla(\bm{y}(\bm{x})-\bm{x})|^{2}\mathrm{d}\bm{x}.
\end{equation*}
Here, we can see that the diffusion regularizer is to make the displacement smooth.

\subsection{Registration-based segmentation method}

The registration-based segmentation method, introduced in \cite{le2011combined}, leverages image registration to perform image segmentation. The model proposed in \cite{le2011combined} is 
\begin{equation}\label{reg-seg_model}
\min_{\bm{y}, a_{1},a_{2}} \int_{\Omega}(I-a_{1})^{2}H(\phi_0(\bm{y}))+(I-a_{2})^{2}(1-H(\phi_0(\bm{y})))\mathrm{d}\bm{x} + \alpha\mathcal{R}(\bm{y}),
\end{equation}
where $\phi_0$ is a given level set function. 
Set $\phi_0$ as the level set function of the region $\mathbb{D}_{1}$. Define
\begin{equation*}
J(a_1,a_2,\bm{x}) = 
\left\{
\begin{split}
&a_{1}, \quad \ \mbox{if} \ \bm{x}\in \mathbb{D}_{1}, \\
&a_{2}, \quad \ \mbox{if} \ \bm{x}\in \mathbb{D}_{2},
\end{split}
\right.
\end{equation*}
where $\Omega = \mathbb{D}_{1}\cup \mathbb{D}_{2}$ and $\mathbb{D}_{1}\cap\mathbb{D}_{2} = \emptyset$.
 By employing the indicator function:
\begin{equation*}
\mathcal{X}_{\Omega}(\bm{x}) = 
\left\{
\begin{split}
&1, \quad \ \mbox{if} \ \bm{x}\in \Omega, \\
&0, \quad \ \mbox{if} \ \bm{x}\notin \Omega,
\end{split}
\right.
\end{equation*}
then $J(a_1,a_2,\bm{x})$ can be rewritten into the following equivalent formulation:
\begin{equation*}\label{prior_image}
J(a_1,a_2,\bm{x}) = \sum_{l=1}^{2} a_{l}\mathcal{X}_{\mathbb{D}_{l}}(\bm{x}).
\end{equation*}
Hence, the model \eqref{reg-seg_model} can be rebuilt as the following model \cite{zhang2021topology2}:
\begin{equation}\label{Reg_bas_seg_model}
\min_{\bm{y},a_1,a_2}   \int_{\Omega}(J(a_1,a_2,\bm{y})-I)^{2}\mathrm{d}\bm{x} + \alpha\mathcal{R}(\bm{y}). 
\end{equation}

Although the models \eqref{reg-seg_model} and \eqref{Reg_bas_seg_model} are mathematically equivalent, they can be understood from different views. For the model \eqref{reg-seg_model}, it is derived by deforming the level set function with a transformation instead of involving the level set function directly. However, taking $J(a_1,a_2,\bm{x})$ as the template and $I(\bm{x})$ as the reference, the model \eqref{Reg_bas_seg_model} is to find a transformation to link the boundaries of the objects in $J(a_1,a_2,\bm{x})$ and $I(\bm{x})$. Since $J(a_1,a_2,\bm{x})$ is defined by users and the corresponding information is known, then the boundaries of the objects in $I(\bm{x})$ can be located. 

Here, compared with the CV model \eqref{Equ_CV_model}, since the models \eqref{reg-seg_model} and \eqref{Reg_bas_seg_model} both solve the transformation rather than finding a level set function, they can avoid the reinitialization. In addition, solving the models \eqref{reg-seg_model} and \eqref{Reg_bas_seg_model} usually need to do the interpolation. But for some special given level set functions $\phi_0$, such as $\phi_0(\bm{x}) = r^2-x_{1}^2-x_{2}^2$, solving the model \eqref{reg-seg_model} does not involve the interpolation. 

\begin{remark}
Consider that if $a_1,a_2$ remain constant in both \eqref{reg-seg_model} and \eqref{Reg_bas_seg_model}, these models will simplify to the conventional variational models for image registration as denoted in \eqref{Registration_framework}. Conversely, the fidelity terms in \eqref{reg-seg_model} and \eqref{Reg_bas_seg_model} incorporate the merits of the CV model presented in \eqref{CV_model}. From this standpoint, the registration-based segmentation models \eqref{reg-seg_model} and \eqref{Reg_bas_seg_model} can be regarded as a fusion of registration and segmentation models.
\end{remark}

Subsequently, we will leverage Theorem \ref{suff_con_star_shape} to construct a segmentation model that preserves the star-shape domain. This model integrates a registration-based segmentation approach with the star-shape constraint, aiming to enhance the segmentation performance. Then we extend the proposed model to some other variants, including star-shape segmentation with respect to multiple centers, partial star-shape segmentation, and selective star-shape segmentation.


\section{The star-shape segmentation models}\label{Model}

In this section, we will delve into the intricacy of integrating the star-shape constraint with the registration-based segmentation model, thereby introducing a novel star-shape segmentation model. This integration aims to capitalize on the strength of both approaches, offering a robust segmentation model that not only benefits from the registration-based methodology but also ensures that the resulting segmented region maintains the desirable star-shape property.

%

\begin{figure}[hthp]
\centering
\subfigure[Target image $I$]{
\includegraphics[width=1.4in,height=1.4in]{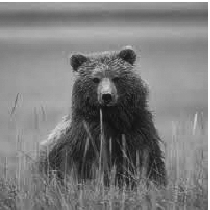}}
\subfigure[Prior image $J$]{
\includegraphics[width=1.4in,height=1.4in]{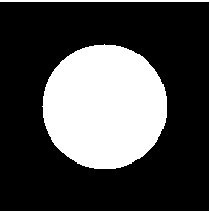}}
\subfigure[Level set function $\phi_0$]{
\includegraphics[width=1.4in,height=1.4in]{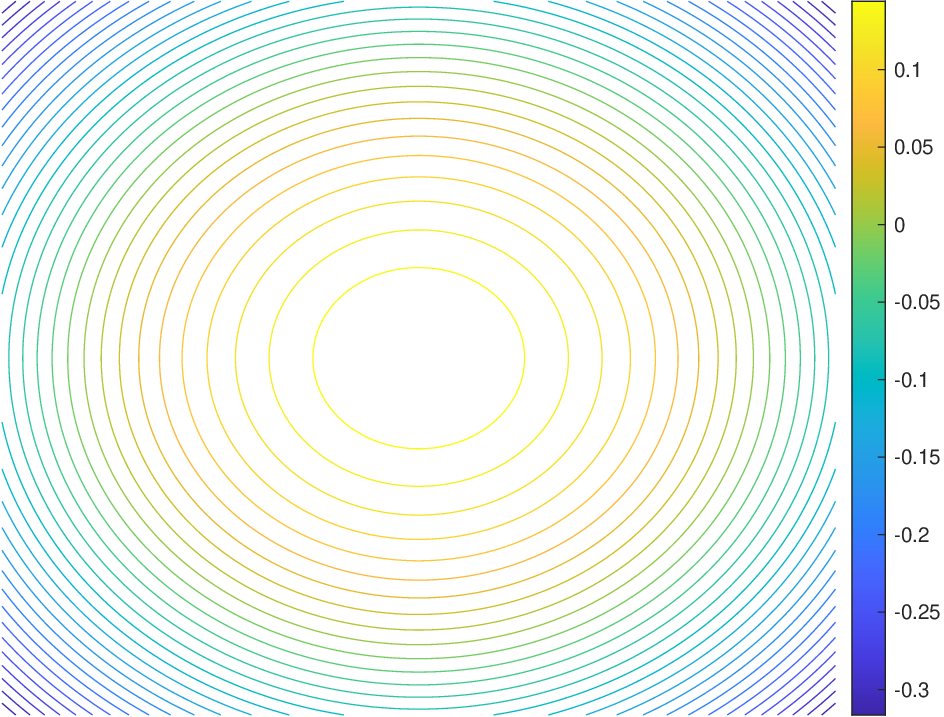}} \\
\subfigure[Segmentation result]{
\includegraphics[width=1.4in,height=1.4in]{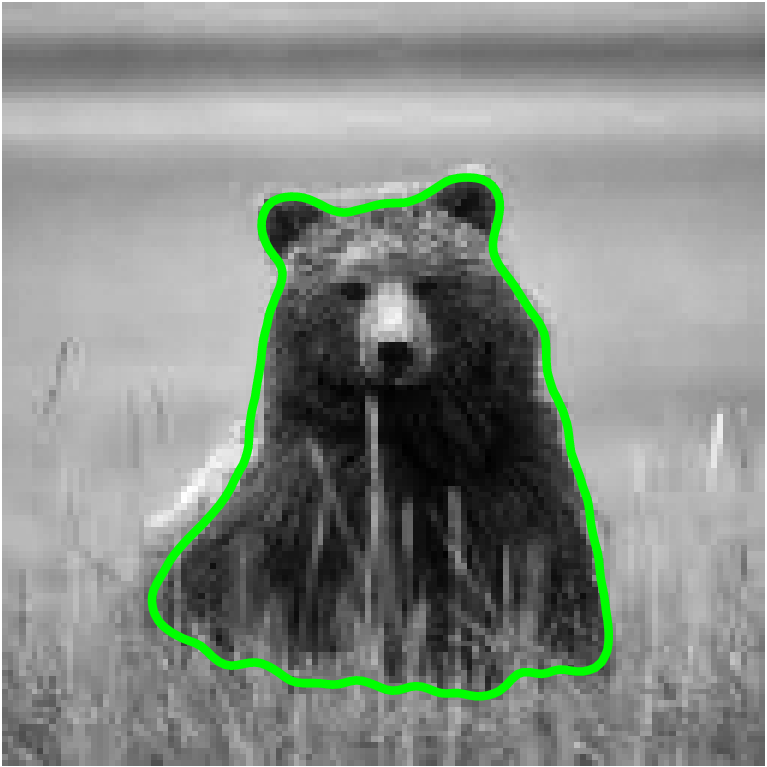}}
\subfigure[Transformation $\bm{y}$]{
\includegraphics[width=1.4in,height=1.4in]{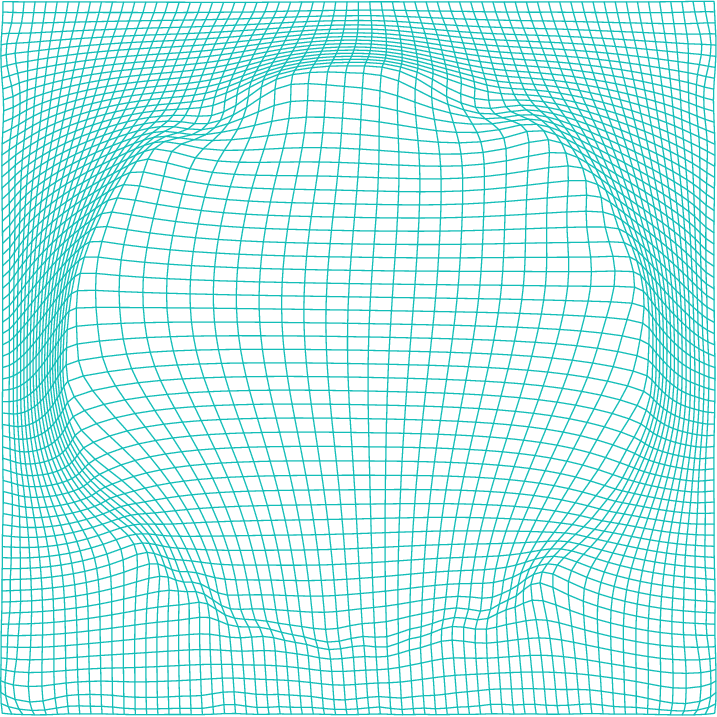}}
\subfigure[Level set function $\phi_0\circ\bm{y}$]{
\includegraphics[width=1.4in,height=1.4in]{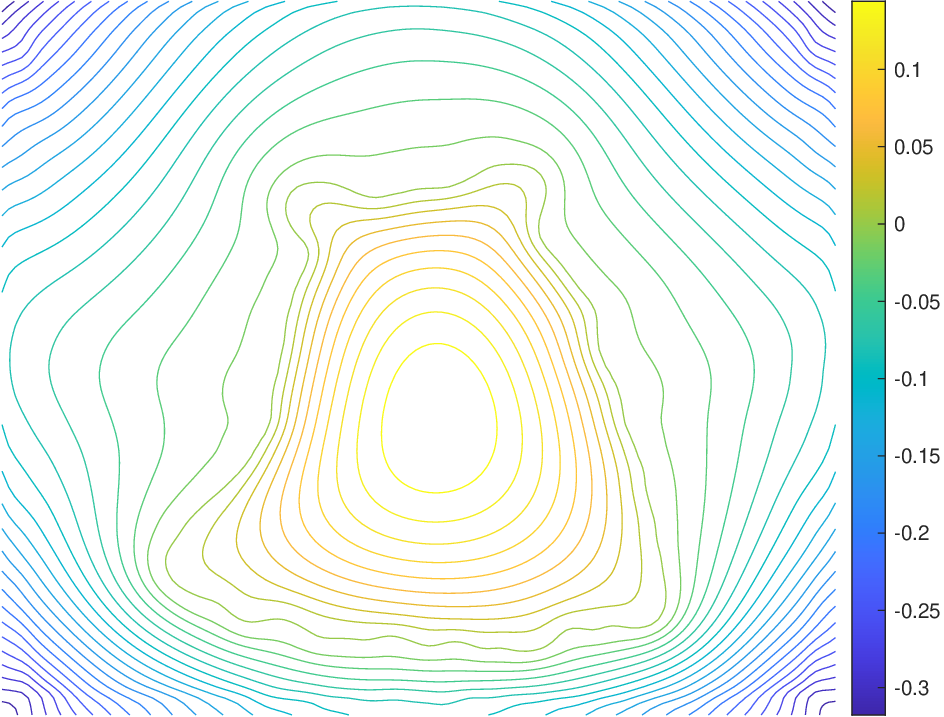}} 
\caption{ {\color{black} An example based on our prior work \cite{zhang2021topology} to demonstrate the registration-based segmentation process.} (a) Target image $I$ (b) Prior image $J$ given by the structure of the target image (c) Level set function $\bm{y}$ with respect to the prior image (d) Segmentation result (e) Transformation $\bm{y}$ (f) Deformed level set function $\phi_0\circ\bm{y}$}\label{fig1}
\end{figure}

In the context of the registration-based segmentation model, a user-prescribed prior image is employed. Within the object domain $\mathbb{D}$ of the prior image, a corresponding level set function $\phi_0$ can be defined. If $\bm{y}$ represents a mapping, then $\phi_0\circ\bm{y}$ becomes the level set function for the deformed domain $\tilde{\mathbb{D}}$, where $\tilde{\mathbb{D}} = \{\bm{x}|\bm{y}(\bm{x})\in\mathbb{D}\}$ denotes the deformed domain associated with $\mathbb{D}$. In Fig. \ref{fig1}, the target image $I$ features a bear as the object of interest. To segment the entire object, a prior image $J$ is provided, depicting a unit disk, along with its corresponding level set function $\phi_0$, such as $1-x_{1}^{2}-x_{2}^{2}$. Following the registration process, a transformation $\bm{y}$ is obtained, and the bear can be effectively represented by the deformed level set function $\phi_0\circ\bm{y}$. {\color{black} Specifically, the segmentation result is obtained by our previous work \cite{zhang2021topology}.}

{
\color{black} Building on the discussion, we can further refine the segmentation model by incorporating suitable constraints on the deformed level set function. A key insight arises from the relationship between convex level set functions and convex regions: if a twice differentiable convex function is a level set function of a region, then the region is a convex region \cite{yan2020convexity}. This property allows us to enforce convexity in the segmentation process. Specifically, if the deformed level set function $\phi_0\circ\bm{y}$ is convex over $\Omega$, the resulting segmented domain $\tilde{\mathbb{D}}$ is guaranteed to be convex. Leveraging this, we formulate a convexity-preserving segmentation model \cite{zhang2021topology2}:
\begin{equation*}
\begin{split}
& \min_{\bm{y}, a_{1},a_{2}} \int_{\Omega}(I-a_{1})^{2}H(\phi_0(\bm{y}))+(I-a_{2})^{2}(1-H(\phi_0(\bm{y})))\mathrm{d}\bm{x} + \alpha\mathcal{R}(\bm{y}), \\
&\ \  \mathrm{s.t.} \ \phi_0\circ\bm{y}\ \mbox{is convex with respect to}\ \Omega.
\end{split}
\end{equation*}
}

{\color{black} Since star-shape domains generalize the notion of convexity, it is natural to extend the above convexity-preserving framework to a star-shape segmentation model.} Inspired by the aforementioned concept and guided by Theorem  \ref{suff_con_star_shape}, we can establish the following proposition: if a center point $\bm{c}$ resides within the interior of $\tilde{\mathbb{D}}$, $\phi_0\circ\bm{y}\in C^{1}(\Omega)$, and $\langle \nabla (\phi_0\circ\bm{y}),\bm{x}-\bm{c}\rangle \leq 0,\ \mbox{a.e.}\ \forall x\in \Omega$, then the domain $\tilde{\mathbb{D}}$ is inherently a star-shape domain relative to the center point $\bm{c}$. Consequently, building upon this rationale, we obtain the subsequent registration-based variational segmentation framework that specifically preserves the star-shape structure:
\begin{equation}\label{basic_model}
\begin{split}
& \min_{\bm{y}, a_{1},a_{2}} \int_{\Omega}(I-a_{1})^{2}H(\phi_0(\bm{y}))+(I-a_{2})^{2}(1-H(\phi_0(\bm{y})))\mathrm{d}\bm{x} + \frac{\alpha}{2}\int_{\Omega}|\nabla (\bm{y}-\bm{x})|^{2}\mathrm{d}\bm{x}, \\
&\ \  \mathrm{s.t.} \ \langle \nabla (\phi_0\circ\bm{y}),\bm{x}-\bm{c}\rangle \leq 0,\ \mbox{a.e.}\ \forall \bm{x}\in \Omega.
\end{split}
\end{equation}

The fitting term (piecewise-constant approximation) used in \eqref{basic_model} is proficient when dealing with visually smooth regions exhibiting homogenous intensity values. However, challenges arise when dealing with intricate geometries or areas distinguished by uneven intensity values. To address this limitation, we first recall a more general model, the so-called Potts model, then modify the above model \eqref{basic_model} to present the main models in this paper. The Potts model for the multiphase image segmentation is to minimize the following functional \cite{tai2023potts}: 
\begin{equation}\label{Potts_model}
\min_{\{\mathbb{D}_{k}\}_{k=1}^{K}}\sum_{k=1}^{K}\int_{\mathbb{D}_{k}}f_{k}(\bm{x})\mathrm{d}\bm{x}+\mathcal{R}(\{\mathbb{D}_{k}\}_{k=1}^{K}),
\end{equation}
where $\{\mathbb{D}_{k}\}_{k=1}^{K}$ is a partition of $\Omega$ such that $\cup_{k=1}^{K}\mathbb{D}_{k} = \Omega$ and $\mathbb{D}_{k}\cap \mathbb{D}_{k'} = \emptyset$ for $k\neq k'$. Here, $f_{k}(\bm{x})$ and $\mathcal{R}(\{\mathbb{D}_{k}\}_{k=1}^{K})$ in \eqref{Potts_model} are the region force function and the regularization to measure the geometry properties of the boundaries of $\{\mathbb{D}_{k}\}_{k=1}^{K}$, respectively. By using level set functions and the Heaviside function, we can equivalently convert the Potts model \eqref{Potts_model} into the following formulation \cite{zhao1996variational}: 
\begin{equation*}
\begin{split}
& \min_{\phi_{1}(\bm{x}),\cdots,\phi_{K}(\bm{x})}\sum_{k=1}^{K}\int_{\Omega}f_{k}(\bm{x})H(\phi_{k}(\bm{x}))\mathrm{d}\bm{x}+\mu\sum_{k=1}^{K}\int_{\Omega}|\nabla H(\phi_{k})(\bm{x})|\mathrm{d}\bm{x}, \\
&\ \qquad \mathrm{s.t.}\ \sum_{k=1}^{K}H(\phi_{k}(\bm{x})) = 1,
\end{split}
\end{equation*}
where we regularize the length of the boundaries of $\{\mathbb{D}_{k}\}_{k=1}^{K}$ and $\phi_k$ is a level set function for $\mathbb{D}_{k}$. Thus, we can naturally get the registration-based segmentation method derived from the Potts model:
\begin{equation*}
\begin{split}
& \min_{\bm{y}}\sum_{k=1}^{K}\int_{\Omega}f_{k}(\bm{x})H(\phi_{k}(\bm{y}))\mathrm{d}\bm{x}+\frac{\alpha}{2}\int_{\Omega}|\nabla (\bm{y}-\bm{x})|^{2}\mathrm{d}\bm{x}, \\
&\ \mathrm{s.t.}\ \sum_{k=1}^{K}H(\phi_{k}(\bm{y})) = 1.
\end{split}
\end{equation*}
Inspired from the approach outlined in \cite{wei2018newregion}, we choose the following region force function:
\begin{equation*}
f_{k}(\bm{x}) = -\log(p_{k}(\bm{x}))+\log(1-p_{k}(\bm{x}),
\end{equation*}
where $p_{k}(\bm{x})$ is the probability of the point $\bm{x}$ belonging to specific region $\tilde{\mathbb{D}}_{k} = \{\bm{x}|\bm{y}(\bm{x})\in\mathbb{D}_{k}\}$ and computed by $\frac{\exp(-|I(\bm{x})-a_{k}|/2\tau^{2})}{\sum_{k'=1}^{K}\exp(-|I(\bm{x})-a_{k'}|/2\tau^{2})}$. In practice, $a_{k}$ is computed by $k$-means {\color{black} clustering method} and $\tau$ is set to one.  The application of this multiphase model is demonstrated in (\ref{proposed_model4}) for $k=3$. 


Now, we are ready to give our main models. {\color{black} Here, we present a hierarchical framework that systematically extends the basic star shape to address increasingly complex real-world scenarios, where objects often exhibit partial, multi-center, or hybrid star-shape properties.
\begin{enumerate}
\item The {\textbf{Basic Star-Shape Segmentation Model}}:
\begin{equation}\label{proposed_model1}
\begin{split}
& \min_{\bm{y}} \int_{\Omega}f_{1}(\bm{x})H(\phi_0(\bm{y}))+f_{2}(\bm{x})(1-H(\phi_0(\bm{y})))\mathrm{d}\bm{x} + \frac{\alpha}{2}\int_{\Omega}|\nabla (\bm{y}-\bm{x})|^{2}\mathrm{d}\bm{x}, \\
&\ \mathrm{s.t.}  \ \langle \nabla (\phi_0\circ\bm{y}),\bm{x}-\bm{c}\rangle \leq 0,\ \mbox{a.e.}\ \forall \bm{x}\in \Omega, 
\end{split}
\end{equation}
is our foundation, which enforces global visibility from a single center point $\bm{c}$. This is particularly valuable for segmenting idealized star-shape structures.
\item The {\textbf{Partial Star-Shape Segmentation Model}}:
\begin{equation}\label{proposed_model2}
\begin{split}
& \min_{\bm{y}} \int_{\Omega}f_{1}(\bm{x})H(\phi_0(\bm{y}))+f_{2}(\bm{x})(1-H(\phi_0(\bm{y})))\mathrm{d}\bm{x} + \frac{\alpha}{2}\int_{\Omega}|\nabla (\bm{y}-\bm{x})|^{2}\mathrm{d}\bm{x}, \\
&\ \mathrm{s.t.}  \ \langle \nabla (\phi_0\circ\bm{y}),\bm{x}-\bm{c}\rangle \leq 0,\ \mbox{a.e.}\ \forall \bm{x}\in \bar\Omega,
\end{split}
\end{equation}
where $\bar\Omega$ is the constrained region prescribed by users, restricts constraints to user-defined subregions $\bar{\Omega}$ and allows free-form deformation elsewhere. This model is the relaxation of the above model \eqref{proposed_model1} and can address many anatomical structures only exhibiting star-shape properties locally. It is critical for handling occlusions or natural variations while maintaining physiologically plausible shapes in key areas.
\item The {\textbf{Multi-Center Star-Shape Segmentation Model}}:
\begin{equation}\label{proposed_model3}
\begin{split}
& \min_{\bm{y}} \int_{\Omega}f_{1}(\bm{x})H(\phi_0(\bm{y}))+f_{2}(\bm{x})(1-H(\phi_0(\bm{y})))\mathrm{d}\bm{x} + \frac{\alpha}{2}\int_{\Omega}|\nabla (\bm{y}-\bm{x})|^{2}\mathrm{d}\bm{x}, \\
&\ \mathrm{s.t.} \left\{
\begin{split}
& \ \langle \nabla (\phi_0\circ\bm{y}),\bm{x}-\bm{c}_{1}\rangle \leq 0,\ \mbox{a.e.}\ \forall \bm{x}\in \Omega_{1}, \\
&\quad \cdots\cdots\cdots\cdots  \\
& \ \langle \nabla (\phi_0\circ\bm{y}),\bm{x}-\bm{c}_{K}\rangle \leq 0,\ \mbox{a.e.}\ \forall \bm{x}\in \Omega_{K}, 
\end{split}\right.
\end{split}
\end{equation}
where $\bm{c}_{1},\cdots,\bm{c}_{K}$ are different center points and $\Omega_{1},\cdots,\Omega_{K}$ are the constrained regions prescribed by users, is the generalization of the above model \eqref{proposed_model2}, which addresses complex structures with multiple ``core regions". Instead of requiring the entire shape to be star-shape around a single center, this model allows different parts of the object to organize around their own natural centers $\bm{c}_{k}$. This flexibility makes it suitable for segmenting irregular but partially structured objects, where global star-shape assumptions would fail.

\item The {\textbf{Selective Star-Shape Segmentation Model}}:
\begin{equation}\label{proposed_model4}
\begin{split}
&
\begin{split}
& \min_{\bm{y}} \int_{\Omega}f_{1}(\bm{x})H(\phi_1(\bm{y}))+f_{2}(\bm{x})H(\phi_2(\bm{y}))+f_{3}(\bm{x})(1-H(\phi_1(\bm{y}))-H(\phi_2(\bm{y})))\mathrm{d}\bm{x} \\
&\qquad \qquad \qquad \qquad \qquad \qquad \qquad \qquad \qquad\qquad \qquad \qquad \qquad + \frac{\alpha}{2}\int_{\Omega}|\nabla (\bm{y}-\bm{x})|^{2}\mathrm{d}\bm{x}, 
\end{split} \\
&\ \mathrm{s.t.}  \ \langle \nabla (\phi_1\circ\bm{y}),\bm{x}-\bm{c}\rangle \leq 0,\ \mbox{a.e.}\ \forall \bm{x}\in \Omega, \\
\end{split}
\end{equation}
where $\phi_{1}(\bm{x})$ and $\phi_{2}(\bm{x})$ are level set functions with respect to $\mathbb{D}_{1}$ and $\mathbb{D}_{2}$ and $\mathbb{D}_{1}\cap\mathbb{D}_{2}=\emptyset$, respectively, are motivated by that real-world scenes often contain mixed topologies. This model simultaneously segments star-shape ($\phi_{1}\circ\bm{y}$) and non-star-shape ($\phi_{2}\circ\bm{y}$) regions via separate level sets, enabling applications like isolating starfish ($\phi_{1}\circ\bm{y}$) from irregular rocks ($\phi_{2}\circ\bm{y}$) in marine imagery. The optional constraint on $\phi_{2}\circ\bm{y}$ provides flexibility to enforce star shape in both regions when needed.
\item The {\textbf{Landmark-Constrained Star-Shape Segmentation Model}}: 
\begin{equation}\label{proposed_model5}
\begin{split}
& \min_{\bm{y}} \int_{\Omega}f_{1}(\bm{x})H(\phi_0(\bm{y}))+f_{2}(\bm{x})(1-H(\phi_0(\bm{y})))\mathrm{d}\bm{x} + \frac{\alpha}{2}\int_{\Omega}|\nabla (\bm{y}-\bm{x})|^{2}\mathrm{d}\bm{x}, \\
&\ \mathrm{s.t.}  \ \langle \nabla (\phi_0\circ\bm{y}),\bm{x}-\bm{c}\rangle \leq 0,\ \mbox{a.e.}\ \forall \bm{x}\in \Omega, \qquad \bm{y}(p_{k}) = q_{k},\ k=1,\cdots,K,
\end{split}
\end{equation}
where $p_{k}$ are some points that are passed through by the identified contour and $q_{k}$ are some points that are located on $\{\bm{x}|\phi_{0}(\bm{x}) = 0\}$, integrates anatomical priors by fixing boundary points $p_{k}\rightarrow q_{k}$, crucial for images where specific features must be preserved. This combines the robustness of star-shape constraints with precision of landmark matching, addressing a key limitation of purely shape-prior driven approaches.

\end{enumerate}
}

\begin{remark}
We just list several representative models \eqref{proposed_model1}-\eqref{proposed_model5} under our framework. Users can flexibly combine the above components, including center points, prescribed regions, and landmarks, to build the suitable model for specific tasks. 
\end{remark}

\begin{remark}
We can also apply the binary image to rewrite the objective functionals in the proposed models \eqref{proposed_model1}-\eqref{proposed_model5}. For example, set $\mathcal{X}_{\mathbb{D}_{1}}(\bm{x})$ and $\mathcal{X}_{\mathbb{D}_{2}}(\bm{x})$ as the indicator functions of $\mathbb{D}_{1}$ and $\mathbb{D}_{2}$, respectively. Here $\mathbb{D}_{1} \cap \mathbb{D}_{2} = \emptyset$ and $\mathbb{D}_{1} \cup \mathbb{D}_{2} = \Omega$. Then the proposed model \eqref{proposed_model1} is equivalent to the following formulation:
\begin{equation*}
\begin{split}
& \min_{\bm{y}} \int_{\Omega}f_{1}(\bm{x})\mathcal{X}_{\mathbb{D}_{1}}(\bm{y})+f_{2}(\bm{x})\mathcal{X}_{\mathbb{D}_{2}}(\bm{y})\mathrm{d}\bm{x} + \frac{\alpha}{2}\int_{\Omega}|\nabla (\bm{y}-\bm{x})|^{2}\mathrm{d}\bm{x}, \\
&\ \mathrm{s.t.}  \ \langle \nabla (\phi_0\circ\bm{y}),\bm{x}-\bm{c}\rangle \leq 0,\ \mbox{a.e.}\ \forall \bm{x}\in \Omega,
\end{split}
\end{equation*}
where $\phi_{0}(\bm{x})$ is a prescribed level set function for the domain $\mathbb{D}_{1}$. For other proposed models, they can also be similarly converted to equivalent formulations involving the binary image.
\end{remark}

It is crucial to note that, despite the involvement of the level set function in the proposed models \eqref{proposed_model1}-\eqref{proposed_model5}, our approach focuses on identifying a transformation that deforms the level set function, which is distinct from directly evolving the level set function. Consequently, in contrast to the traditional level set method, our framework eliminates the need for reinitialization to obtain the signed distance function throughout the solving process. This key distinction streamlines the computational process and enhances the efficiency of our approach. In addition, since our model is based on the registration, it is easy to force the deformed contour to pass through some landmarks.


\section{The proposed algorithm}\label{Alg}
In this section, our focus is on developing an algorithm specifically tailored for solving the proposed model \eqref{proposed_model1}. Notably, for the models \eqref{proposed_model2}-\eqref{proposed_model5}, a seamless extension of the same algorithmic framework can be effortlessly applied.

In devising a fast algorithm for solving the proposed model \eqref{proposed_model1}, we adopt the alternating direction method of multipliers (ADMM). ADMM stands as a variant of the augmented Lagrangian method. The process involves initially formulating the augmented Lagrangian functional corresponding to the model and subsequently resolving the variables alternately. 

To apply ADMM to \eqref{proposed_model1}, we first introduce an auxiliary variable $q(\bm{x})$ such that $\langle \nabla (\phi_0\circ\bm{y}(\bm{x})),\bm{x}-\bm{c}\rangle = q(\bm{x})$. So the proposed model \eqref{proposed_model1} has the following equivalent formulation:
\begin{equation*}
\begin{split}
&  \min_{\bm{y},q}\int_{\Omega}f_{1}H(\phi_0(\bm{y}))+f_{2}(1-H(\phi_0(\bm{y})))\mathrm{d}\bm{x} + \frac{\alpha}{2}\int_{\Omega}|\nabla (\bm{y}-\bm{x})|^{2}\mathrm{d}\bm{x}, \\
&\ \mathrm{s.t.} \ \langle \nabla (\phi_0\circ\bm{y}),\bm{x}-\bm{c}\rangle = q,\ q(\bm{x}) \leq 0, \mbox{a.e.}\ \forall \bm{x}\in \Omega.
\end{split}
\end{equation*}
Next, we introduce an indicator functional $\delta_{\Omega}(q(\bm{x}))$:
\begin{equation*}
\delta_{\Omega}(q(\bm{x})) := 
\left\{
\begin{split}
& 0 \quad\quad  \mathrm{if}\ q(\bm{x})\leq 0 \ \forall \bm{x} \in \Omega, \\
& +\infty \ \mathrm{if}\ q(\bm{x}) > 0 \ \exists \bm{x} \in \Omega.
\end{split}\right.
\end{equation*}
Thus, the augmented Lagrangian functional of the proposed model \eqref{proposed_model1} is built as:
\begin{equation*}
\begin{split}
\mathcal{L}_{A}(\bm{y}(\bm{x}),q(\bm{x}),\lambda(\bm{x}),\sigma) := & \int_{\Omega}f_{1}H(\phi_0(\bm{y}))+f_{2}(1-H(\phi_0(\bm{y})))\mathrm{d}\bm{x}  \\
&+ \frac{\alpha}{2}\int_{\Omega}|\nabla (\bm{y}-\bm{x})|^{2}\mathrm{d}\bm{x}+ \int_{\Omega}\lambda(\langle \nabla (\phi_0\circ\bm{y}),\bm{x}-\bm{c}\rangle - q)\mathrm{d}\bm{x} \\
&+ \frac{\sigma}{2}\int_{\Omega}(\langle \nabla (\phi_0\circ\bm{y}),\bm{x}-\bm{c}\rangle - q)^{2}\mathrm{d}\bm{x}+\delta_{\Omega}(q),
\end{split}
\end{equation*}
where $\lambda(\bm{x})$ is the Lagrangian multiplier and $\sigma>0$ is a penalty parameter. Now, we are ready to give the $l$-th iteration of the ADMM scheme for the proposed model \eqref{proposed_model1}:
\begin{equation}\label{ADMM_scheme}
\left\{
\begin{split}
\bm{y}^{l}(\bm{x}) &:= \mathrm{arg}\min_{\bm{y}(\bm{x})} \mathcal{L}_{A}(\bm{y}(\bm{x}),q^{l-1}(\bm{x}),\lambda^{l-1}(\bm{x}),\sigma); \\
q^{l}(\bm{x}) & := \mathrm{arg}\min_{q(\bm{x})} \mathcal{L}_{A}(\bm{y}^{l}(\bm{x}),q(\bm{x}),\lambda^{l-1}(\bm{x}),\sigma); \\
\lambda^{l}(\bm{x}) &:= \lambda^{l-1}(\bm{x}) + \sigma(\langle \nabla (\phi_0\circ\bm{y}^{l}(\bm{x})),\bm{x}-\bm{c}\rangle - q^{l}(\bm{x})).
\end{split}\right.
\end{equation}

Next, we first give the discretization of above subproblems, and then show details of how to solve these subproblems.

For simplicity, we consider the square domain $\Omega = [0,1]^2$. We use the cell-centered grid to define a spatial partition
\begin{equation*}
\Omega_{h} = \left\{\bm{x}^{i,j}\in\Omega| \bm{x}^{i,j} = (x_{1}^{i},x_{2}^{j}) = \left(\left(i-\frac{1}{2}\right)h,\left(j-\frac{1}{2}\right)h\right), 1\leq i\leq n, 1\leq j\leq n)\right\},
\end{equation*}
where $h = \frac{1}{n}$. Then we discretize the transformation $\bm{y}$, the auxiliary variable $q$, and the Lagrangian multiplier $\lambda$ on the cell-centered grid, namely $\bm{y}^{i,j} = (y_{1}^{i,j},y_{2}^{i,j}) = (y_{1}(x_{1}^{i},x_{2}^{j}),y_{2}(x_{1}^{i},x_{2}^{j}))$, $q^{i,j} = q(x_{1}^{i},x_{2}^{j})$, and $\lambda^{i,j} = \lambda(x_{1}^{i},x_{2}^{j})$.
According to the lexicographical ordering, we reshape
\begin{equation*}
X = (X_1^{t},X_2^{t})^{t} = (x_{1}^{1},\cdots,x_{1}^{n},\cdots,x_{1}^{1},\cdots,x_{1}^{n},x_{2}^{1},\cdots,x_{2}^{n},\cdots,x_{2}^{n},\cdots,x_{2}^{n})^{t},
\end{equation*}
\begin{equation*}
Y = (Y_1^{t},Y_2^{t})^{t} = (y_{1}^{1,1},\cdots,y_{1}^{n,1},\cdots,y_{1}^{1,n},\cdots,y_{1}^{n,n},y_{2}^{1,1},\cdots,y_{2}^{n,1},\cdots,y_{2}^{1,n},\cdots,y_{2}^{n,n})^{t},
\end{equation*}
\begin{equation*}
Q = (q^{1,1},\cdots,q^{n,1},\cdots,q^{1,n},\cdots,q^{n,n})^{t},
\end{equation*}
\begin{equation*}
\Lambda = (\lambda^{1,1},\cdots,\lambda^{n,1},\cdots,\lambda^{1,n},\cdots,\lambda^{n,n})^{t},
\end{equation*}
and
\begin{equation*}
\Phi_0\circ Y = (\phi_0(\bm{y}^{1,1}),\cdots,\phi_0(\bm{y}^{n,1}),\cdots,\phi_0(\bm{y}^{n,1}),\cdots,\phi_0(\bm{y}^{n,n}))^{t}.
\end{equation*}
By the forward difference and Neumann boundary condition, the discrete gradient operator can be defined as
\begin{equation*}
A = \begin{pmatrix}
I_{2}\otimes A_1 \\
I_{2}\otimes A_2
\end{pmatrix}, \ A_1 =  I_{n}\otimes \partial_{n}^{1,h}, \ A_2 =  \partial_{n}^{1,h}\otimes I_{n}, \
\partial_{n}^{1,h} = \frac{1}{h}
\begin{pmatrix}
-1 & 1 & & & \\
    & -1 & 1 & & \\
    & \cdots  &\cdots &\cdots  & \\ 
    & & &-1 & 1 \\
    &   & &  & 0
\end{pmatrix}\in\mathbb{R}^{n\times n},
\end{equation*}
where $I_{n}$ is a $n\times n$ identity matrix and $\otimes$ represents a Kronecker product. 


\paragraph{Subproblem $\bm{y}$.} The subproblem $\bm{y}$ has the following formulation:
\begin{equation*}
\begin{split}
\min_{Y} h^{2}\sum_{i=1}^{n}\sum_{j=1}^{n}f_{1}(\bm{x}^{i,j})H(\phi_0(\bm{y}^{i,j}))&+f_{2}(\bm{x}^{i,j})(1-H(\phi_0(\bm{y}^{i,j}))) \\
&+ \frac{\alpha h^2}{2}(Y-X)A^{t}A(Y-X)+\frac{\sigma h^2}{2}S^{t}S.
\end{split}
\end{equation*}
where $S = W\cdot(\Phi_0\circ Y)-Q^{l-1}+\Lambda^{l-1}/ \sigma$, $W = \mathrm{Diag}(X_1-c_1)A_1+\mathrm{Diag}(X_2-c_2)A_2$, and $\mathrm{Diag}(v), v\in\mathbb{R}^{n^2 \times 1}$ is a diagonal matrix whose diagonal element $(\mathrm{Diag}(v))_{k,k}$ is $v_{k}$, for $k=1,\cdots, n^{2}$. However, since the Heaviside function $H$ is discontinuous, we may use a smooth function 
$H_{\epsilon}(v) = \frac{1}{2}\left(1+\frac{2}{\pi}\arctan\left(\frac{v}{\epsilon}\right)\right)$ to make the approximation. Then in practice, we solve the following approximated problem:
\begin{equation}\label{discretization_subproblem_y}
\begin{split}
\min_{Y} F(Y):= h^{2}\sum_{i=1}^{n}\sum_{j=1}^{n}f_{1}(\bm{x}^{i,j})H_\epsilon(\phi_0(\bm{y}^{i,j}))&+f_{2}(\bm{x}^{i,j})(1-H_\epsilon(\phi_0(\bm{y}^{i,j}))) \\
&+ \frac{\alpha h^2}{2}(Y-X)A^{t}A(Y-X)+\frac{\sigma h^2}{2}S^{t}S.
\end{split}
\end{equation}

Here, we choose a modified Newton method to solve \eqref{discretization_subproblem_y}. The gradient and Hessian of \eqref{discretization_subproblem_y} are
\begin{equation*}
d_{F} = h^2 g +\alpha h^2 A^{t}A(Y-X)\ \ \mathrm{and} \ \
H_{F} = h^{2}M + \alpha h^2 A^{t}A,
\end{equation*}
respectively, where $g$ and $M$ are shown in Appendix \ref{appendix1}. However, since $H_{F}$ is usually not positive definite,  then solving the following Newton equation
\begin{equation}\label{Newtonequation}
H_{F}p = -d_{F}
\end{equation}
may not obtain a descent direction $p$. Thus, we need to make a modification about $M$ to guarantee that solving \eqref{Newtonequation} indeed leads to a descent direction $p$. {\color{black} In Appendix \ref{appendix2}, we give the details of how to do this modification to derive a semi-positive definite system. While a strictly positive definite system (e.g., via adding a small identity perturbation, as in Levenberg-Marquardt method) would theoretically guarantee descent, our numerical experiments demonstrate that the modified semi-positive definite system already yields descent directions without explicit regularization.} Hence, the iteration scheme to solve \eqref{discretization_subproblem_y} is
\begin{equation*}
\tilde{Y} = Y+\eta \delta Y,
\end{equation*}
where $\delta Y$ is obtained by solving \eqref{Newtonequation} with a modification $M$ and $\eta$ is the step length derived by the Armijo line search satisfying the sufficient descent condition. During the numerical implementation, the iteration process for solving the subproblem $\bm{y}$ concludes when any of the following conditions is met: 1) the number of iterations reaches $5$; 2) $\|d_{F}^{m}\|_{2}\leq 10^{-1}\times \|d_{F}^{0}\|_{2}$; 3) $\|d_{F}^{m}\|_{2} \leq 10^{-3}$. This stopping criteria ensure a balance between computational efficiency and achieve a satisfactory solution for the subproblem.

\paragraph{Subproblem $\bm{q}$.} The subproblem $q$ is equivalent to the following constrained optimization problem:
\begin{equation}\label{subproblem_p}
\begin{split}
&\min_{Q}  \frac{\sigma h^{2}}{2} (W\cdot(\Phi_0\circ Y^{l})-Q+\Lambda^{l-1}/ \sigma)^{t}(W\cdot(\Phi_0\circ Y^{l})-Q+\Lambda^{l-1}/ \sigma), \\
&\ \mathrm{s.t. \ all\ the\ components\ of}\ Q\ \mathrm{are\ nonpositive.} 
\end{split}
\end{equation}
By the projection, we can easily get the closed-form solution of \eqref{subproblem_p}:
\begin{equation*}
Q_{k} = 
\left\{
\begin{split}
&(W\cdot(\Phi_0\circ Y^{l})+\Lambda^{l-1}/ \sigma)_k \ \mathrm{if}\ (W\cdot(\Phi_0\circ Y^{l})+\Lambda^{l-1}/ \sigma)_k \leq 0, \\
&0\ \qquad\qquad\qquad\qquad\qquad\ \ \ \mathrm{if} \ (W\cdot(\Phi_0\circ Y^{l})+\Lambda^{l-1}/ \sigma)_k > 0,
\end{split}\right.
\end{equation*}
for $k=1,\cdots, n^{2}$.

\paragraph{The choice of the penalty parameter $\bm{\sigma}$.} While the ADMM scheme \eqref{ADMM_scheme} typically involves a fixed penalty parameter $\sigma$, it is worth noting that in practice, the convergence can be notably influenced by dynamically adjusting the penalty parameter $\sigma$ throughout the iterations \cite{boyd2011distributed}. Here, we simply use the following choice:
\begin{equation*}
\sigma^{l+1} = 
\left\{
\begin{split}
&10\sigma^{l} \ \mathrm{if} \ \|r^{l+1}\|_{\infty}\geq 0.95\times\|r^{l}\|_{\infty}, \\
&\sigma^{l} \quad\ \mathrm{if} \ \mathrm{otherwise},
\end{split}\right.
\end{equation*}
where $r^{l} = W\cdot(\Phi_0\circ Y^{l})-Q^{l}$ is the residual of the $l$-th iteration. In other words, we will expand $\sigma$ by a factor $10$ if the infinity norm of the $l$-th residual does not sufficiently decrease.

\paragraph{Multilevel Strategy.} {\color{black} The nonconvex model \eqref{proposed_model1} requires careful initialization to achieve good solutions. We address this using a multilevel strategy \cite{modersitzki2009fair} that progressively refines the registration from coarse to fine resolutions. First, we build an image pyramid through $L$ levels of downsampling, preserving key structural features at each coarser level. Optimization starts at the coarsest level, where lower dimensionality enables efficient computation. The solution is then upsampled and refined at each subsequent level, with each stage providing better initialization to avoid local minima. When reaching the original resolution, this process yields an accurate registration. The method combines computational efficiency (from coarse-level optimization) with robustness (by using global features to guide local refinements), making it ideal for solving nonconvex optimization problem \eqref{proposed_model1}.}

%

Now, we can summarize ADMM with the multilevel strategy to solve \eqref{proposed_model1} in Algorithm \ref{Alg1}. Here, the stopping criteria is set as $\|r^{l}\|_{\infty} \leq 10^{-2}$, namely, when the infinity norm of the residual is smaller than a prescribed threshold, the algorithm will terminate. 

\begin{algorithm}[h!]
\caption{ADMM with the multilevel strategy to solve the proposed model \eqref{proposed_model1}.}
\label{Alg1}
\begin{algorithmic}
\STATE{Compute the largest possible number of levels based
                         on the size of $I(\bm{x})$: $L=$ Maxlevel.\\
                         Define the coarsest level as level $1$ and the finest level as level $L$.\\
                       Work out the multilevel representation of the given image $I(\bm{x})$: $I^{1}(\bm{x}),\cdots,I^{l}(\bm{x}),\cdots,I^{L}(\bm{x})$.}
\STATE{Set the level set function $\phi_0(\bm{x})$.}
\FOR{$level = 1:L$ }
\STATE{Set $l=0$;}
\IF{$level=1$}
\STATE{Input $I^{1}(\bm{x}), \sigma^{1,0}$. Set $Y^{1,0}$, and $Q^{1,0}$;}
\ELSE
\STATE{Input $I^{level}(\bm{x}), \sigma^{level,0}$. Compute $Y^{level,0}$ and $Q^{level,0}$ by interpolating $Y^{level-1,\star}$ and $Q^{l-1,\star}$;}
\ENDIF
\WHILE{$\|r^{level,l}\|_{\infty}\leq 10^{-2}$}
\STATE{Update $Y^{level,l+1}$ by solving \eqref{discretization_subproblem_y};}
\STATE{Update $Q^{level,l+1}$ by solving \eqref{subproblem_p};}
\STATE{Update $\Lambda^{level,l+1}$ by $\Lambda^{level,l+1}= \Lambda^{level,l} + \sigma^{level,l}(T\cdot(\Phi_0\circ Y^{level,l+1})-Q^{level,l+1})$;}
\IF{$\|r^{level,l+1}\|_{\infty}\geq$$0.95\times$$\|r^{level,l}\|_{\infty}$}
\STATE{Update $\sigma^{level,l+1}$ by $\sigma^{level,l+1} = 10\sigma^{level,l}$;}
\ELSE
\STATE{Set $\sigma^{level,l+1} = \sigma^{level,l}$;}
\ENDIF
\STATE{Set $l = l+1$;}
\ENDWHILE
\STATE{Output $Y^{level,\star}$ and $Q^{level,\star}$.}
\ENDFOR
\STATE{Output $Y^{\star} = Y^{L,\star}$ and $Q^{\star}=Q^{L,\star}$.}
\end{algorithmic}
\end{algorithm}


\section{Numerical experiments}\label{Result}
In this section, the proposed models \eqref{proposed_model1}-\eqref{proposed_model5} are tested to using 2D synthetic and real images to showcase their advantages. The implementation is carried out using Matlab R2022a on a MacBook Pro with an Apple M1 Pro processor and 16 GB RAM. The tested images are all rescaled into the size of $256\times 256$ and their corresponding intensity values are normalized to the range $[0,1]$. For the approximation function $H_{\epsilon}(v) = \frac{1}{2}(1+\frac{2}{\pi}\arctan(\frac{v}{\epsilon}))$, we set $\epsilon$ as $0.01$ for all the tests. \textcolor{black}{To enable a comprehensive comparison, we employ the Matlab built-in function \textit{activecontour} with the default parameter (smooth factor is set to $0$) to generate segmentation results using the CV model \cite{chan2001active}. For further benchmarking, we also conduct comparative experiments with the convexity-preserving registration-based segmentation model proposed in \cite{zhang2021topology2}}. For all the tests in this section, we adopt a four-level strategy, where we downscale the target images to four scales: $32\times 32$, $64\times 64$, $128\times 128$, and $256\times 256$.

\subsection{The performance of the proposed models \eqref{proposed_model1}-\eqref{proposed_model5} on synthetic images} 

In this part, we test the proposed models \eqref{proposed_model1}-\eqref{proposed_model5} on some synthetic images. For the parameter $\alpha$ in the proposed models, we just set $0.001$. 

\begin{figure}[htbp!]
\centering
\subfigure[Target images with different initial contours]{
\includegraphics[width=1.15in,height=1.15in]{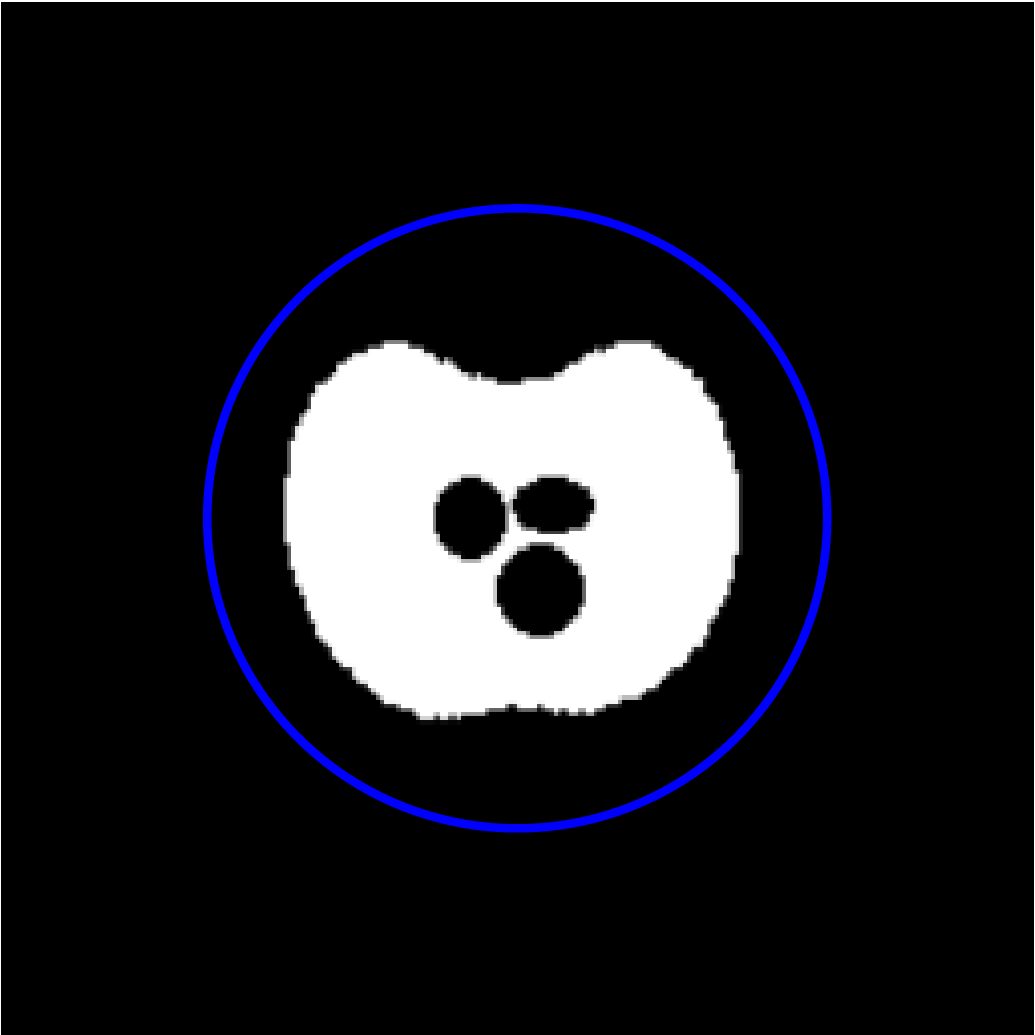}
\includegraphics[width=1.15in,height=1.15in]{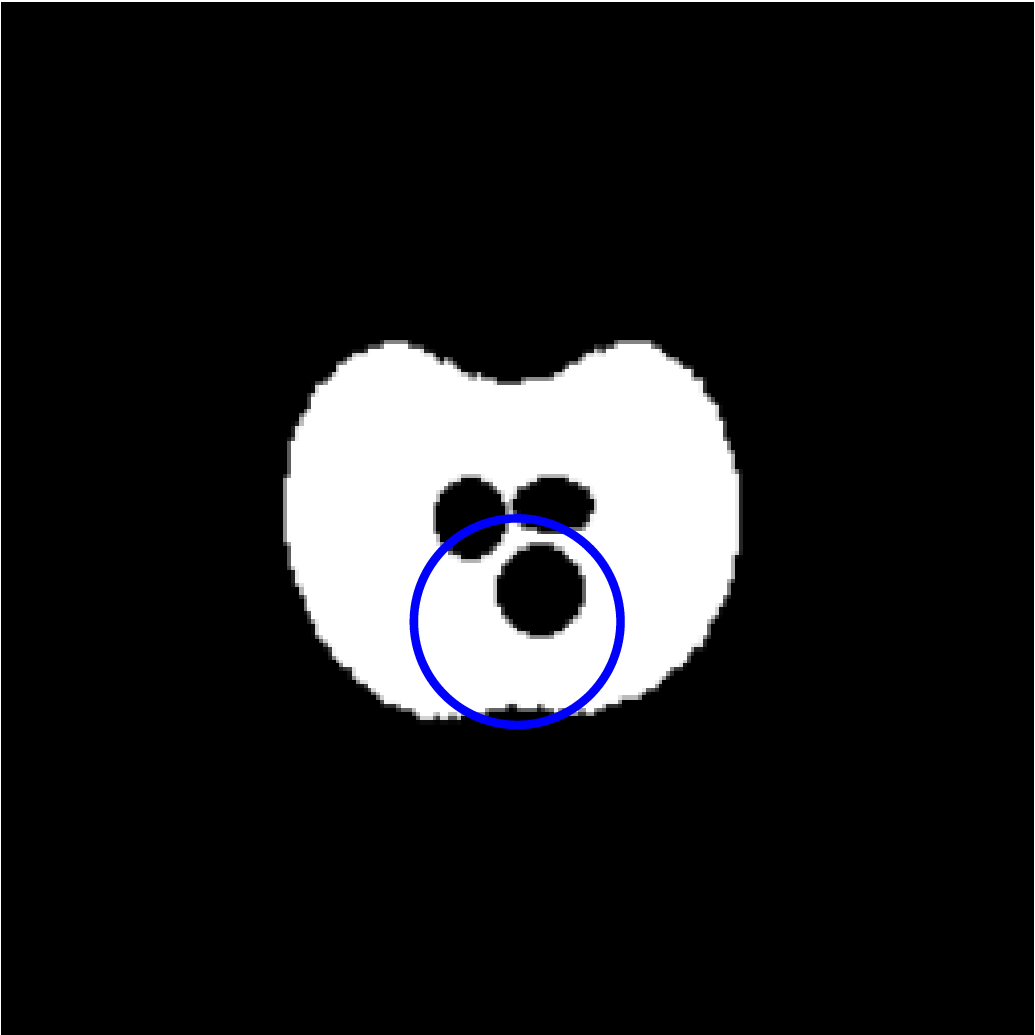}
\includegraphics[width=1.15in,height=1.15in]{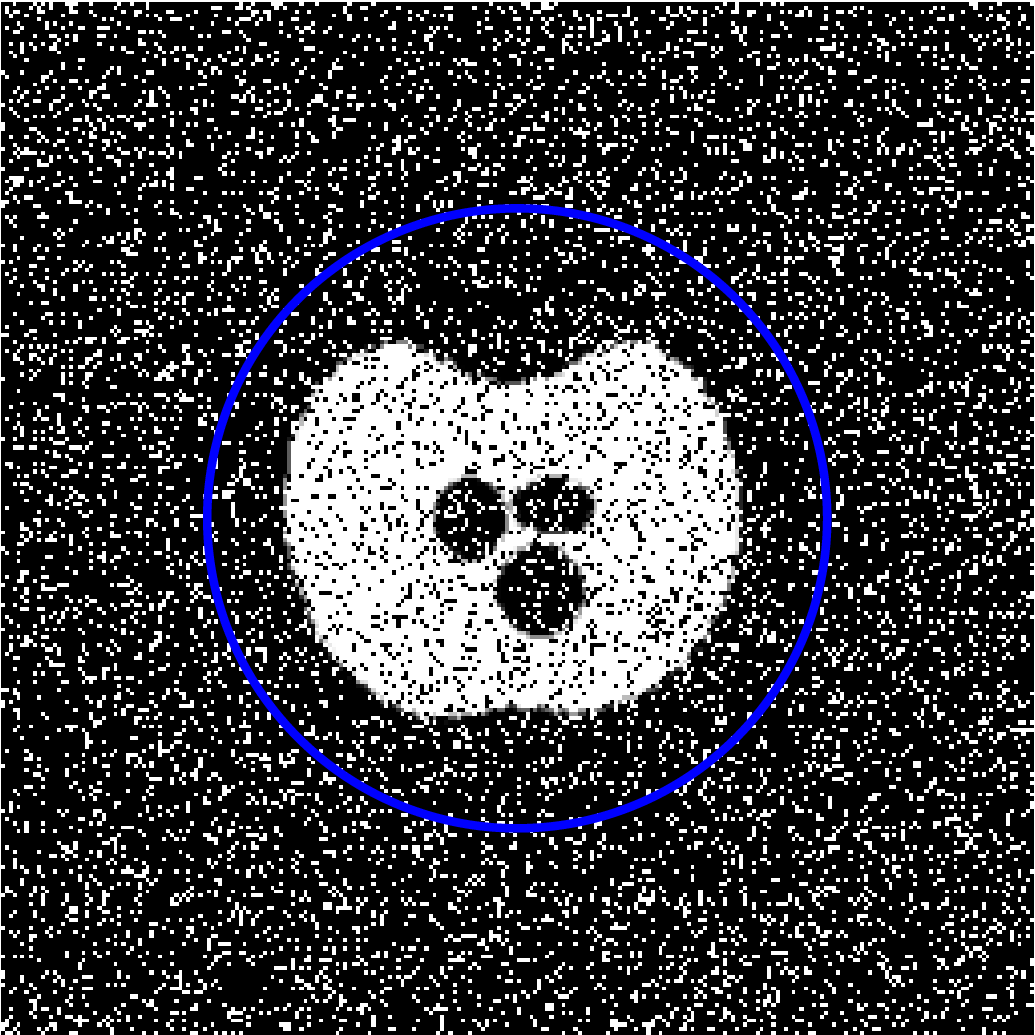}
\includegraphics[width=1.15in,height=1.15in]{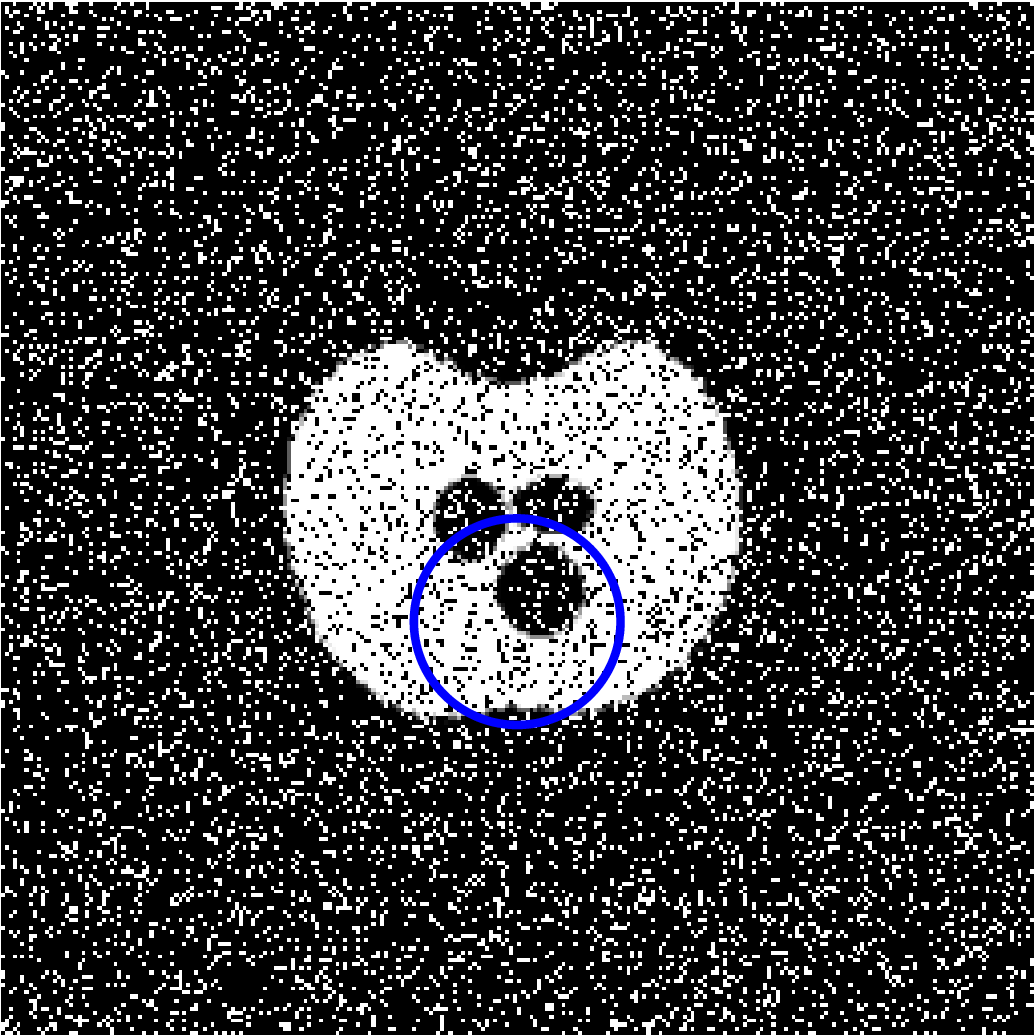}}\\
\subfigure[Segmentation results by CV. \textcolor{black}{The running times are 2.9 s, 3.1 s, 5.6 s, and 4.6 s.} ]{
\includegraphics[width=1.15in,height=1.15in]{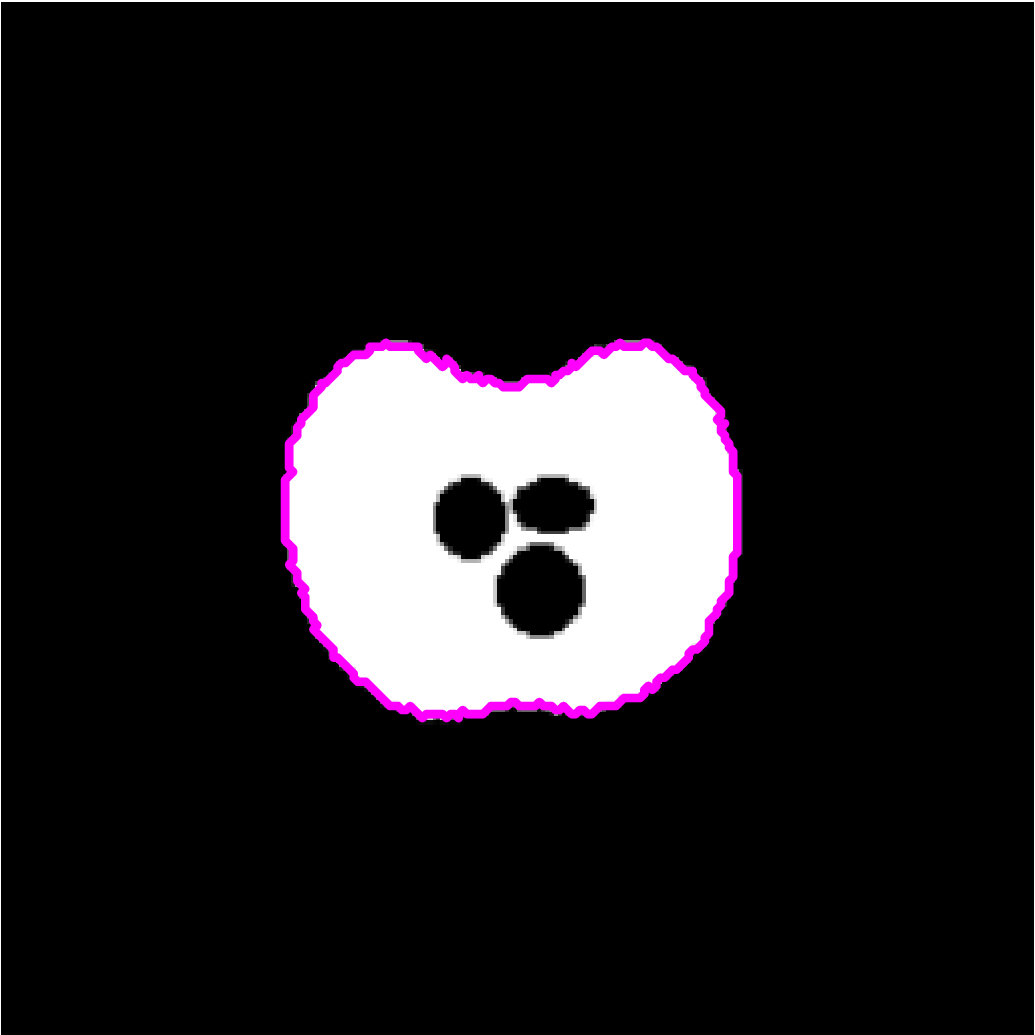}
\includegraphics[width=1.15in,height=1.15in]{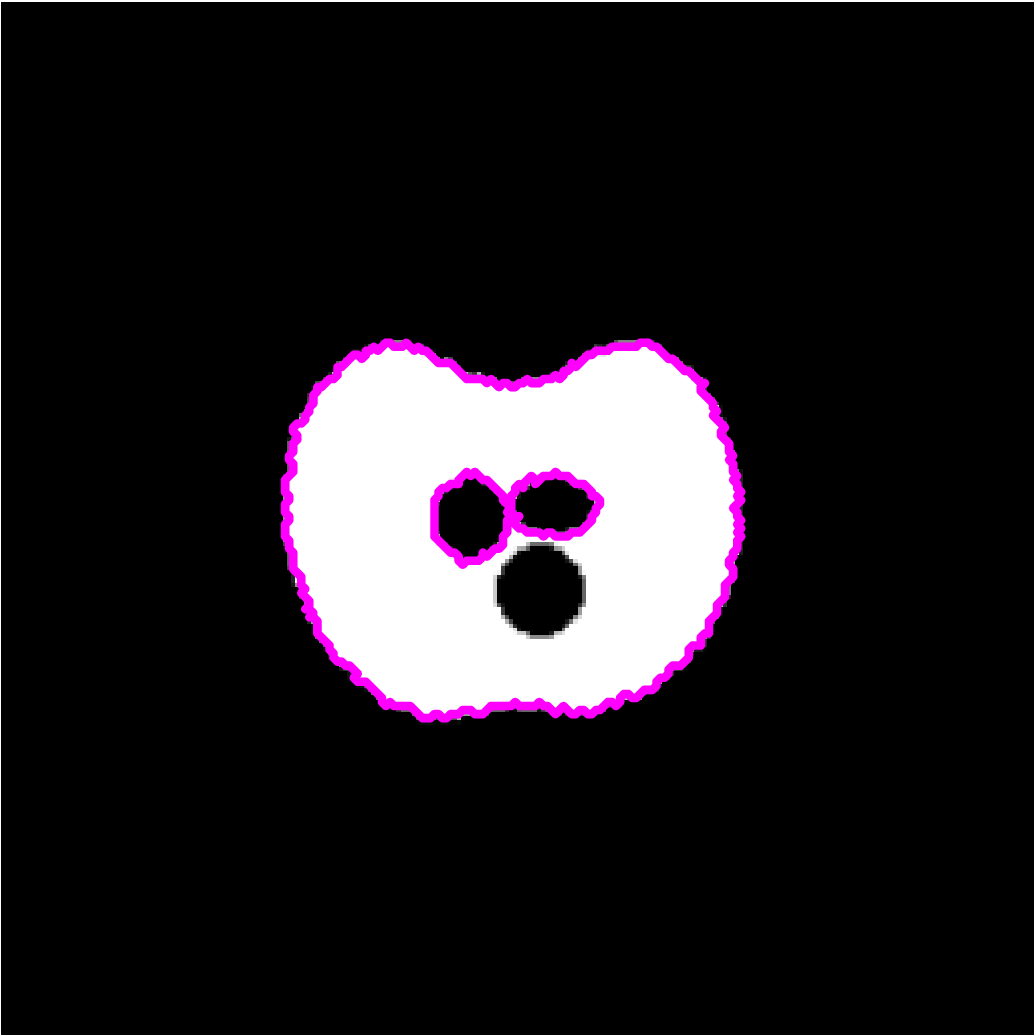}
\includegraphics[width=1.15in,height=1.15in]{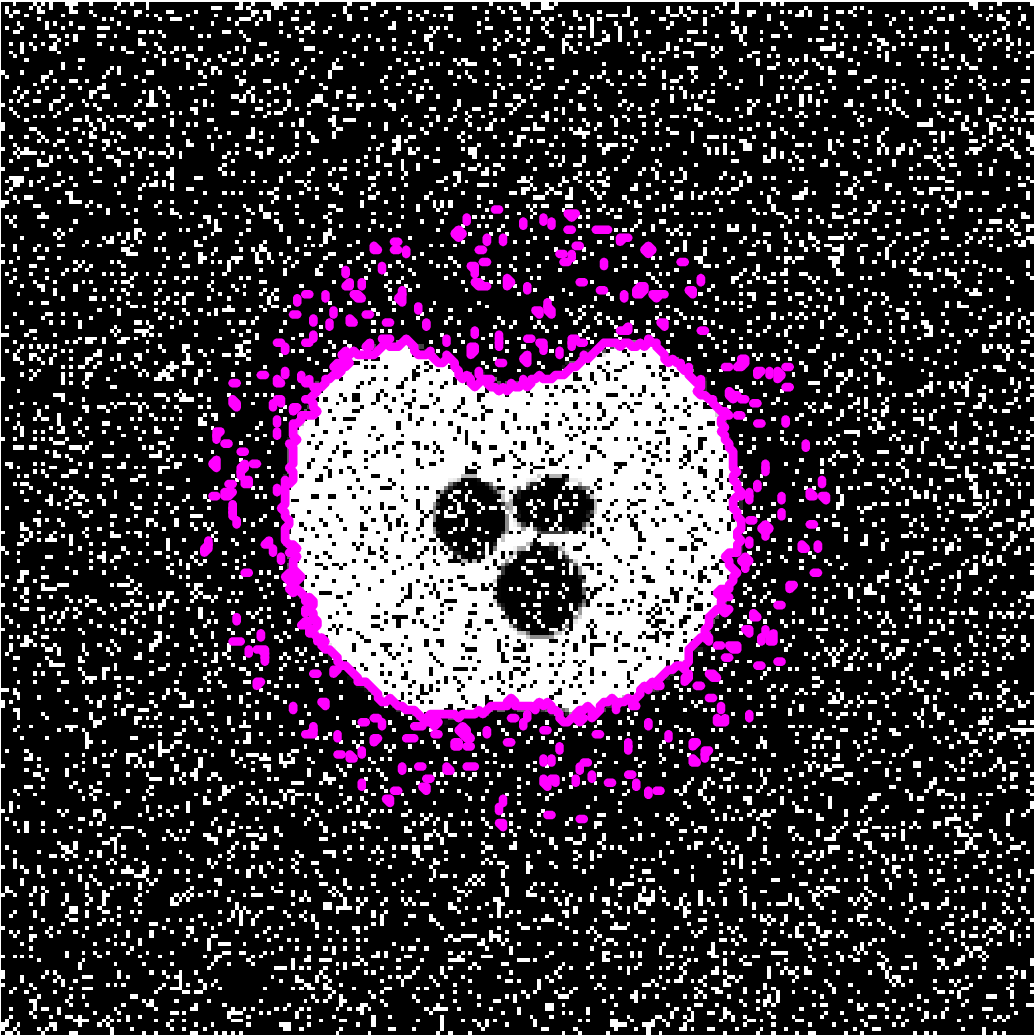} 
\includegraphics[width=1.15in,height=1.15in]{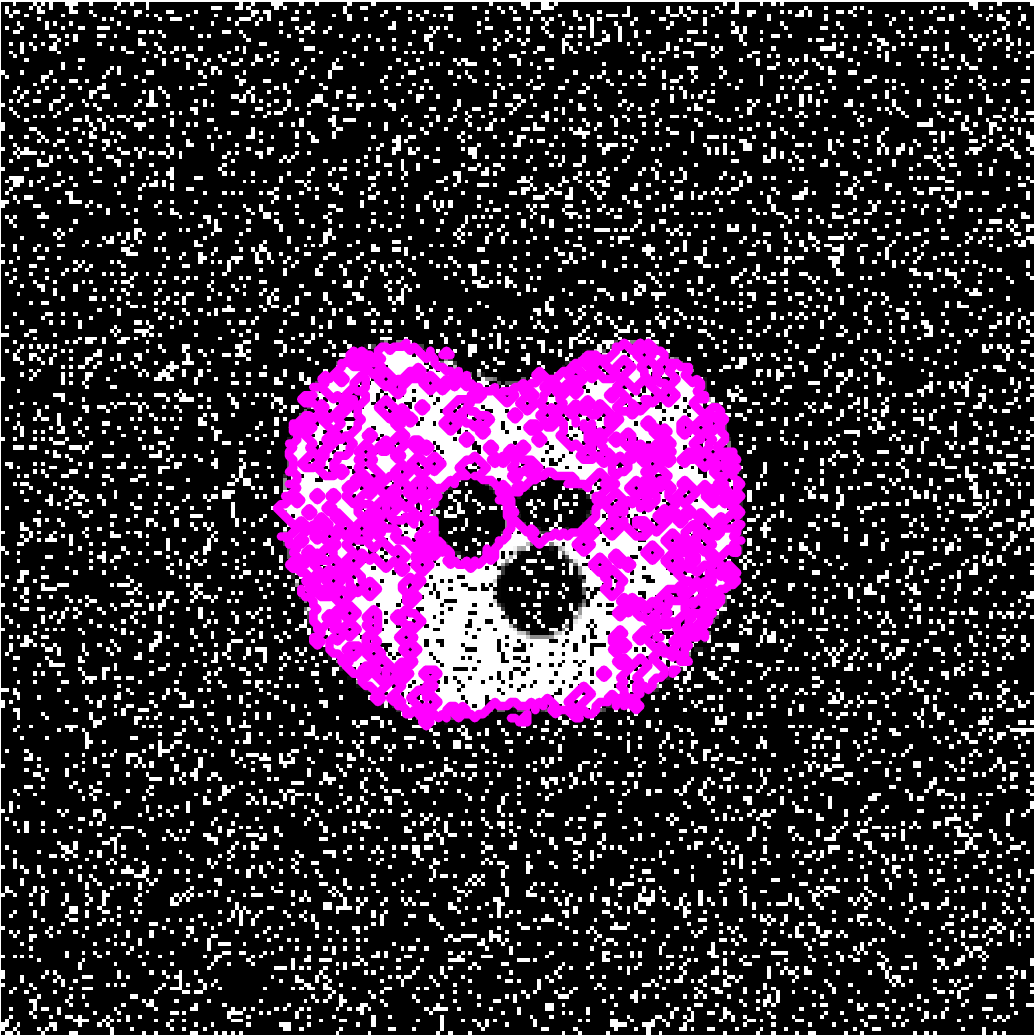}}\\
\subfigure[\textcolor{black}{Segmentation results by the convexity-preserving model.  The running times are 8.0 s, 9.6 s, 3.4 s, and 3.0 s.}]{
\includegraphics[width=1.15in,height=1.15in]{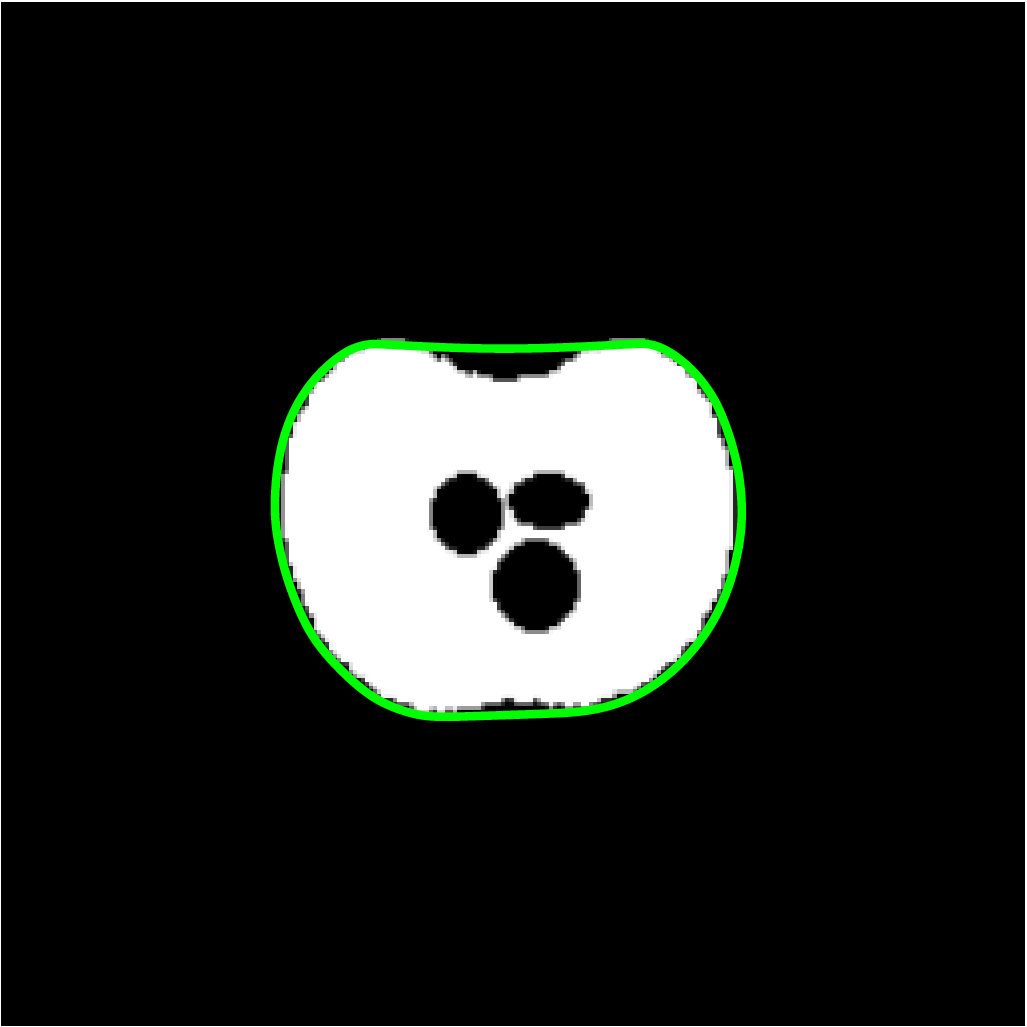}
\includegraphics[width=1.15in,height=1.15in]{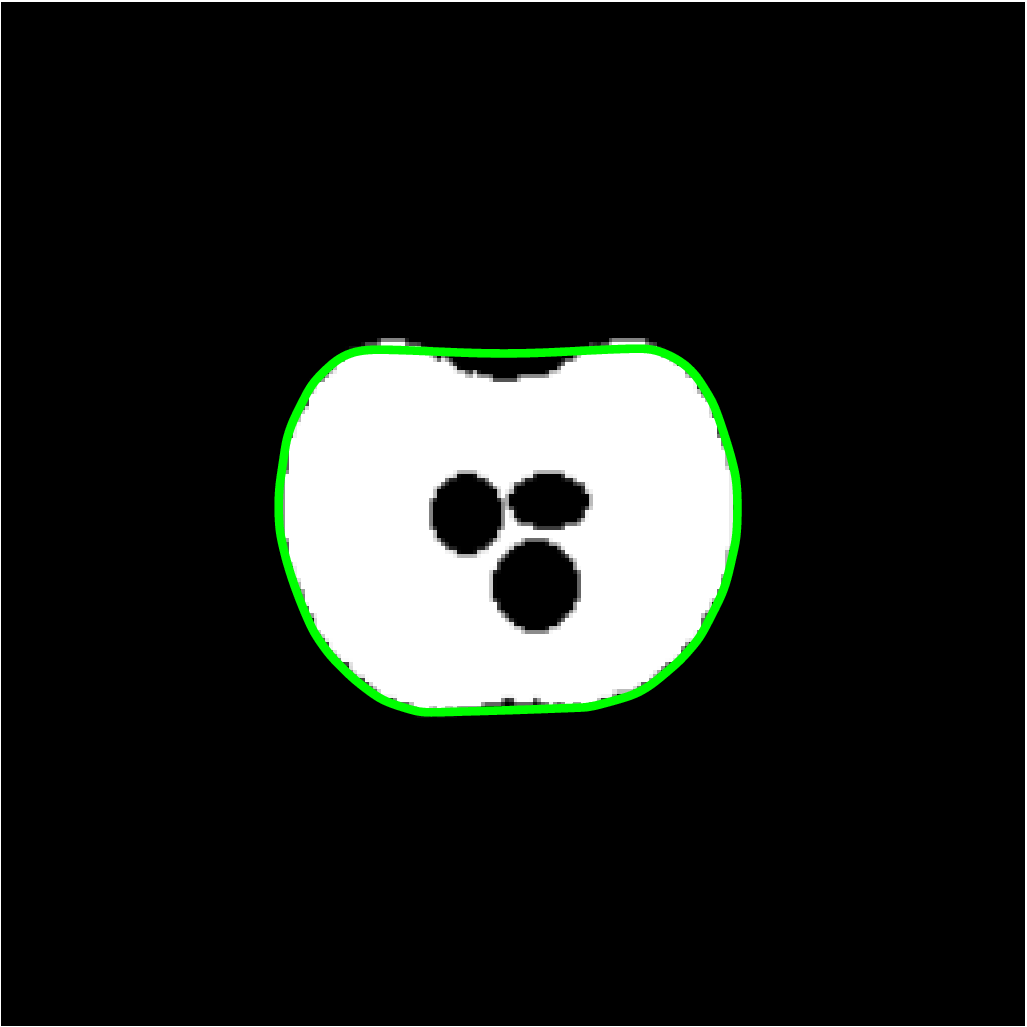}
\includegraphics[width=1.15in,height=1.15in]{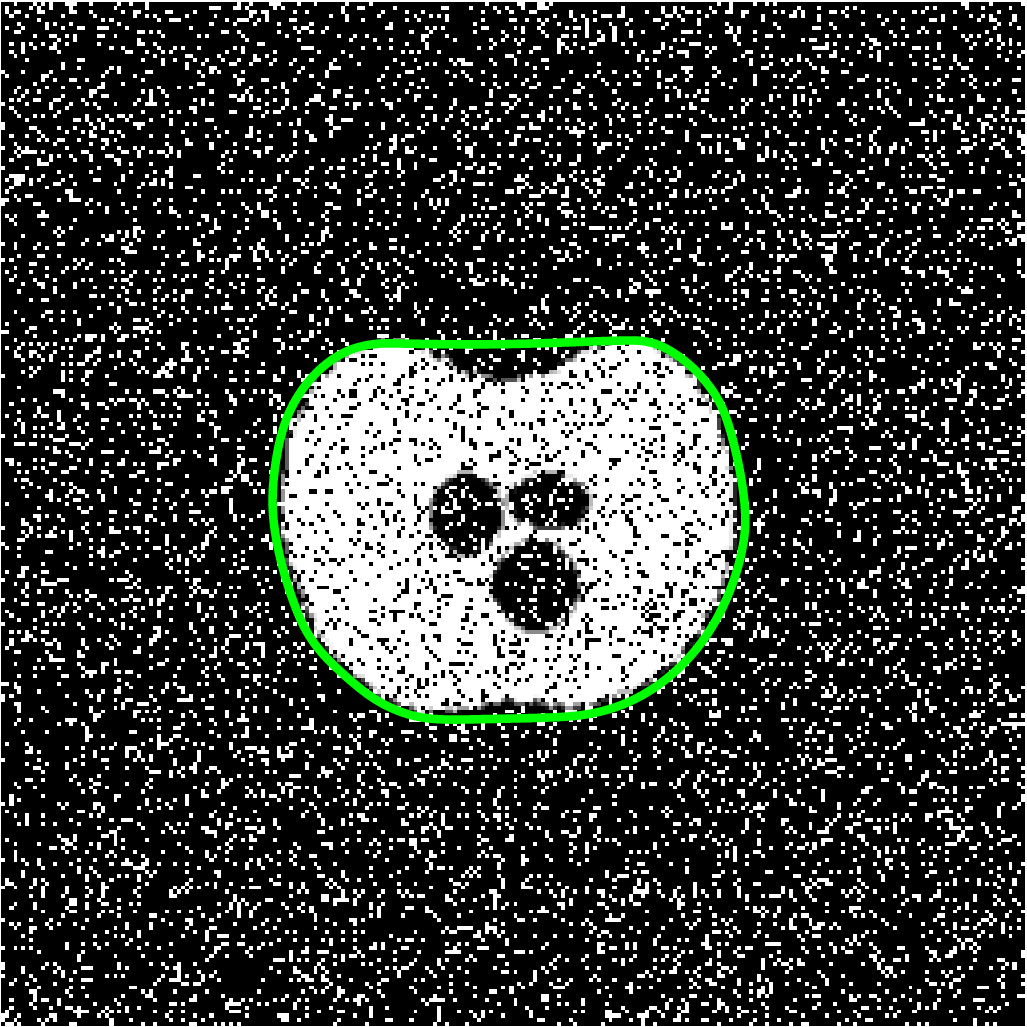}
\includegraphics[width=1.15in,height=1.15in]{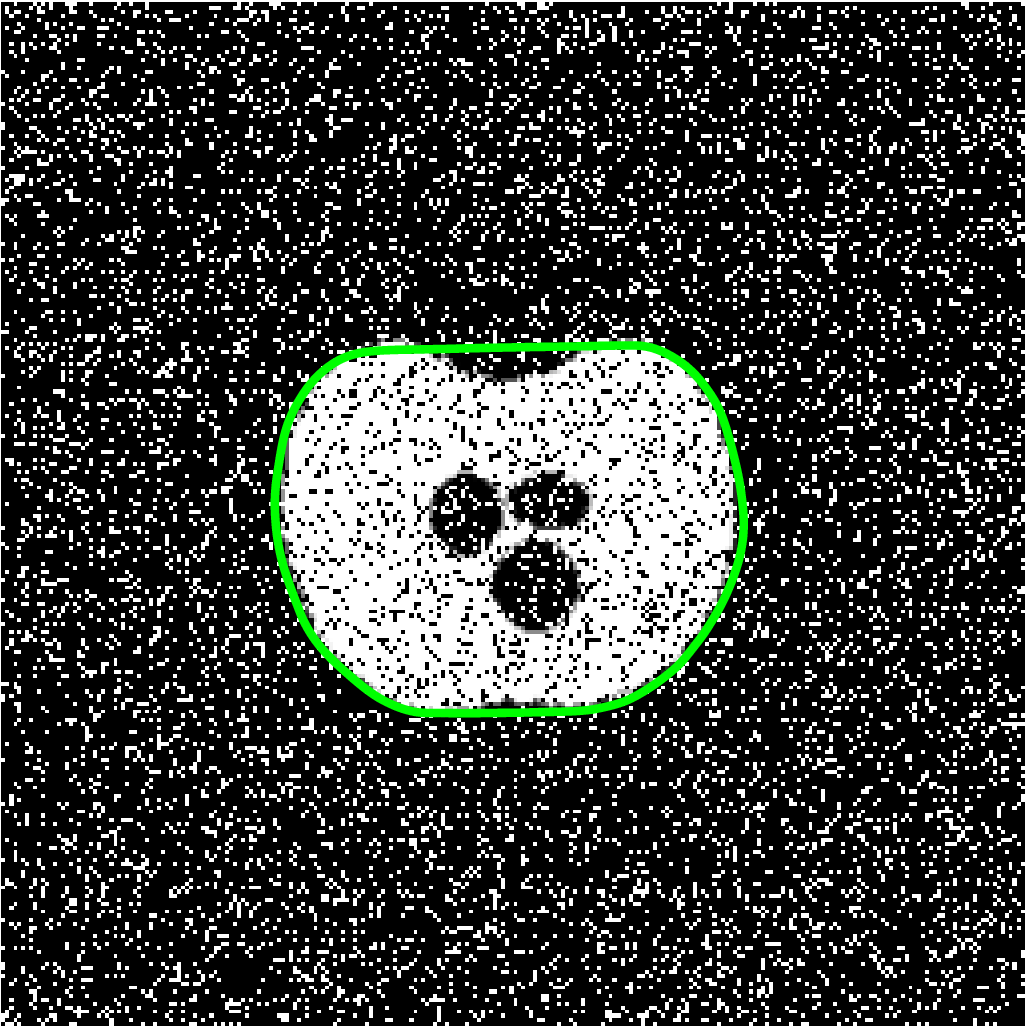}}\\
\subfigure[Segmentation results by the proposed model \eqref{proposed_model1}. \textcolor{black}{The running times are 1.1 s, 6.5 s, 1.5 s, and 7.0 s.}]{
\includegraphics[width=1.15in,height=1.15in]{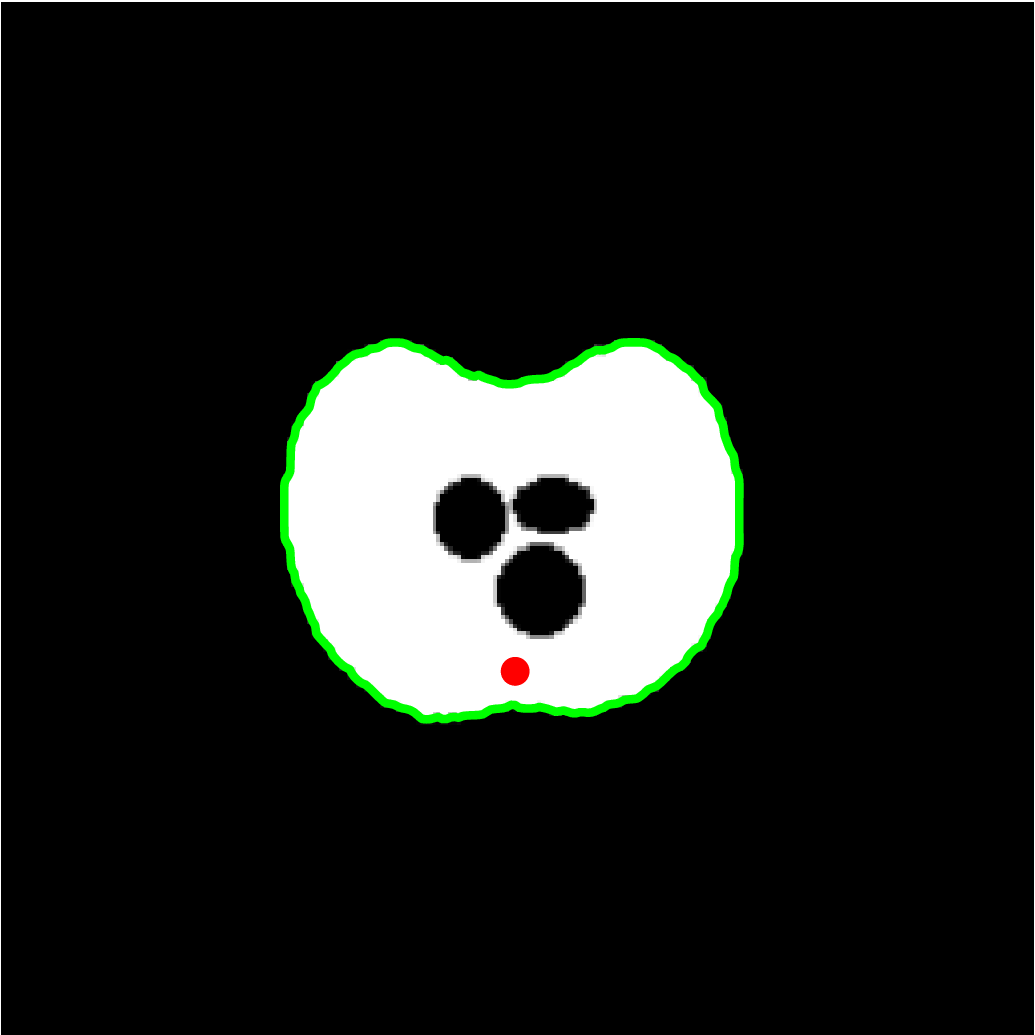}
\includegraphics[width=1.15in,height=1.15in]{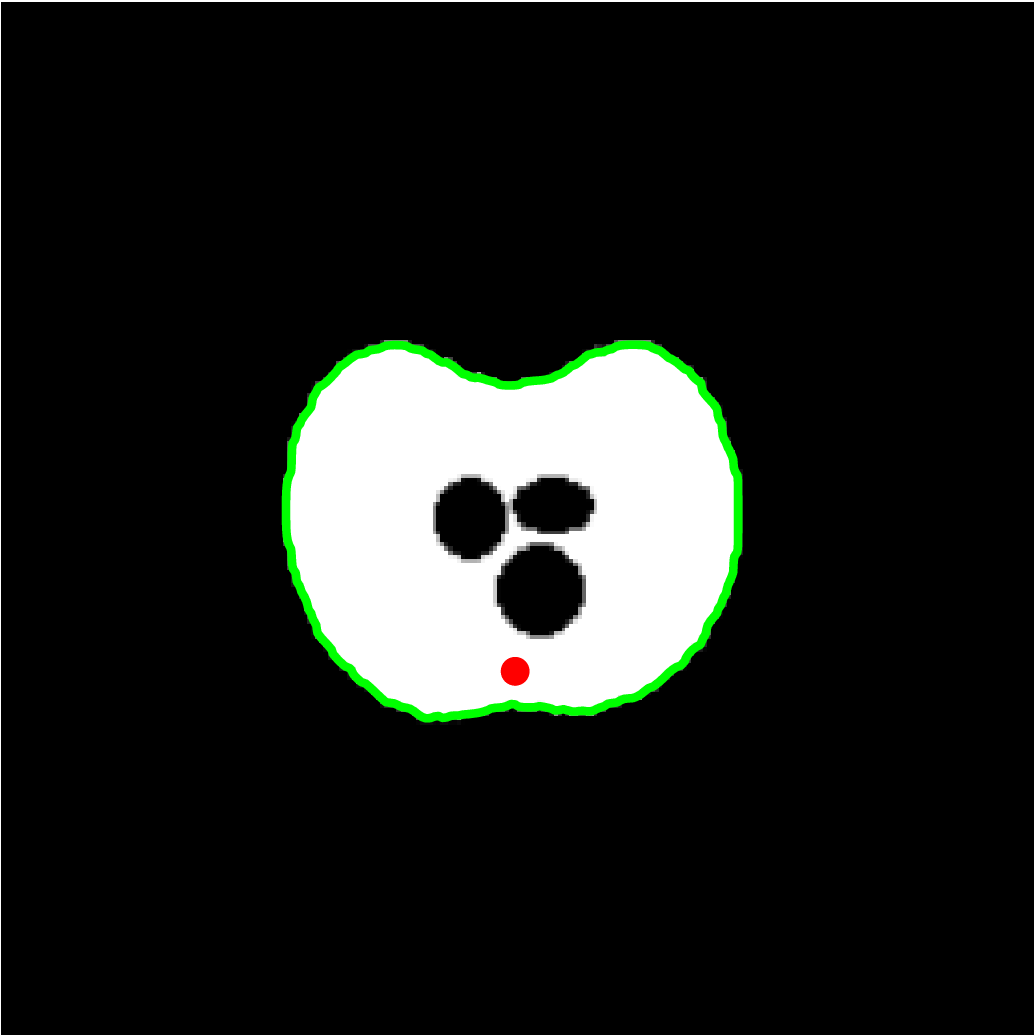}
\includegraphics[width=1.15in,height=1.15in]{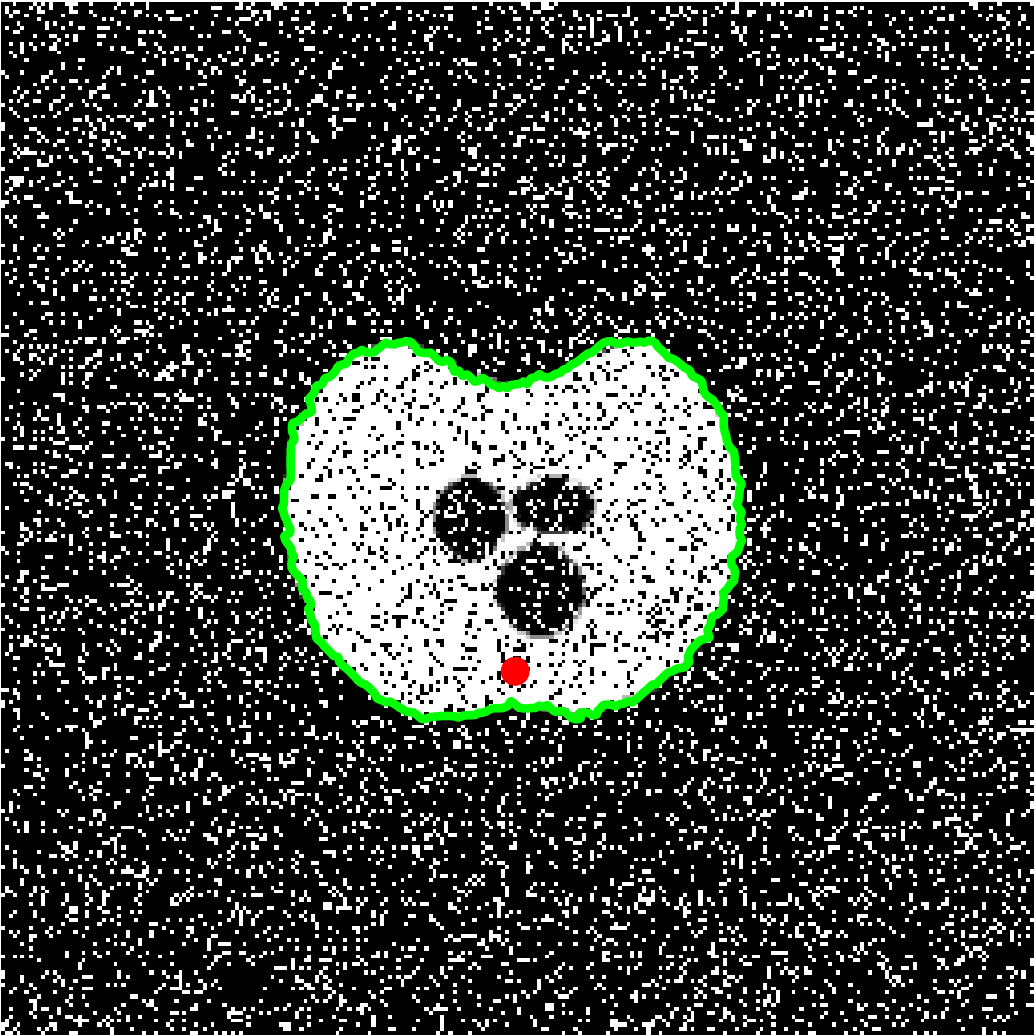}
\includegraphics[width=1.15in,height=1.15in]{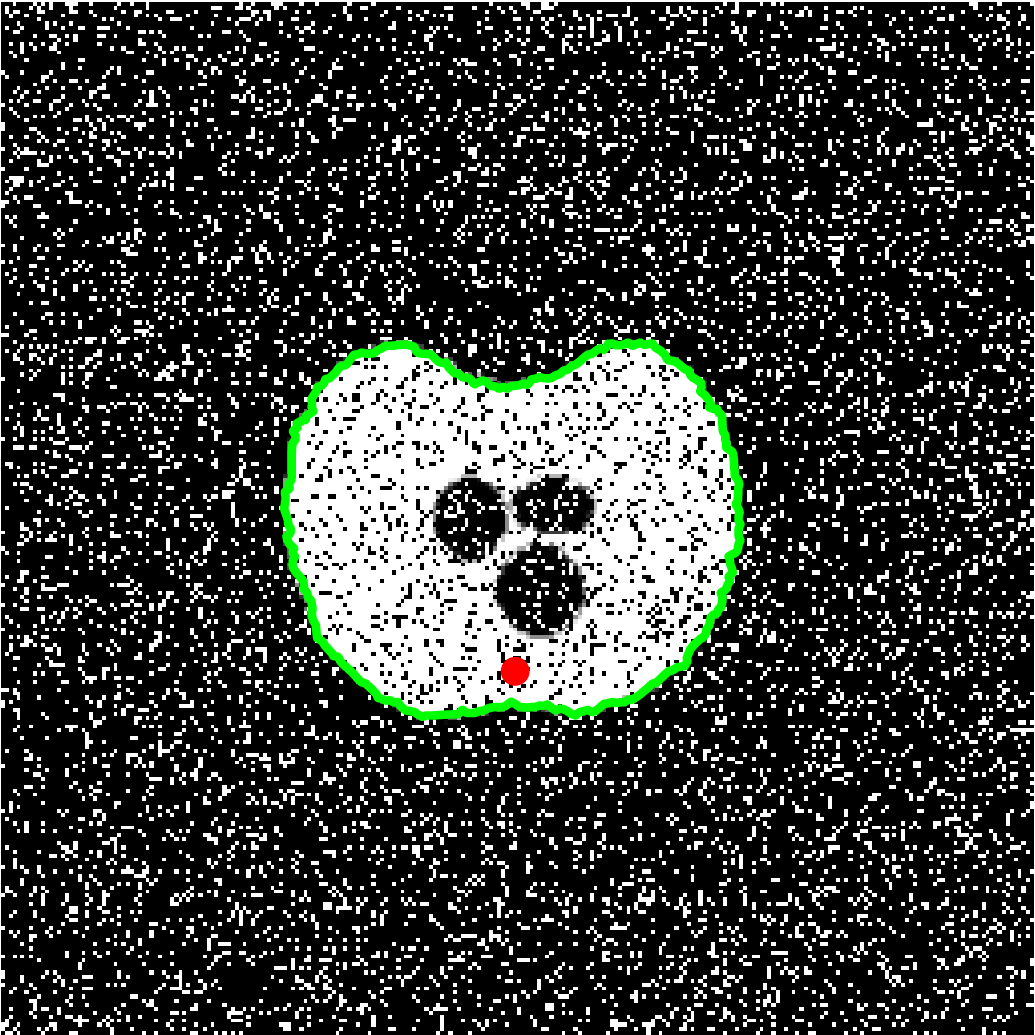}}
\caption{Test on the proposed model \eqref{proposed_model1}. 
The first row shows the target images with different initial contours, where one is a clean image and the other one contains noise. The second row gives the segmentation results by the CV model. \textcolor{black}{The third row shows the segmentation results by the convexity-preserving model \cite{zhang2021topology2}.} The  \textcolor{black}{fourth} row displays the segmentation results by the proposed model \eqref{proposed_model1}. Here, the red point is the given center point. \textcolor{black}{The running time is measured in seconds.}}\label{Exp1_fig1}
\end{figure}

We first apply the proposed model \eqref{proposed_model1} to two straightforward examples, shown in the first row of Fig. \ref{Exp1_fig1}. Here, one image is artificially generated and the other one is added with 30$\%$ impulsive noise by the Matlab built-in function \textit{imnoise}. For each case, we choose two different initial contours. From the second row of Fig. \ref{Exp1_fig1}, we can see that for the clean image, the CV model gives different segmentation results and for the noise image, it generates worse segmentation results, seriously affected by the noise. 
\textcolor{black}{The convexity-preserving segmentation model \cite{zhang2021topology2} inherently produces whole-part segmentation results, as demonstrated in the third row of Fig. \ref{Exp1_fig1}. Notably, this model effectively avoids generating outliers, particularly in noisy images. However, this convexity-preserving property results in the slightly reduced segmentation accuracy. In contrast, our proposed model \eqref{proposed_model1} achieves accurate segmentation regardless of initial contour placement, consistently producing star-shape domains centered around the specified point (fourth row of Fig. \ref{Exp1_fig1}). The proposed model maintains its robustness with noisy images, successfully segmenting the entire target region without outlier generation. Furthermore, our model demonstrates superior computational efficiency compared to the convexity-preserving approach \cite{zhang2021topology2}. This performance advantage stems from our formulation's reliance on first-order derivatives, while the convexity-preserving model requires computationally more expensive second-order derivative calculations.
}

\begin{figure}[htbp!]
\centering
\subfigure[Energy and residual for the noise image: Case 1]{
\includegraphics[width=2.0in,height=2.0in]{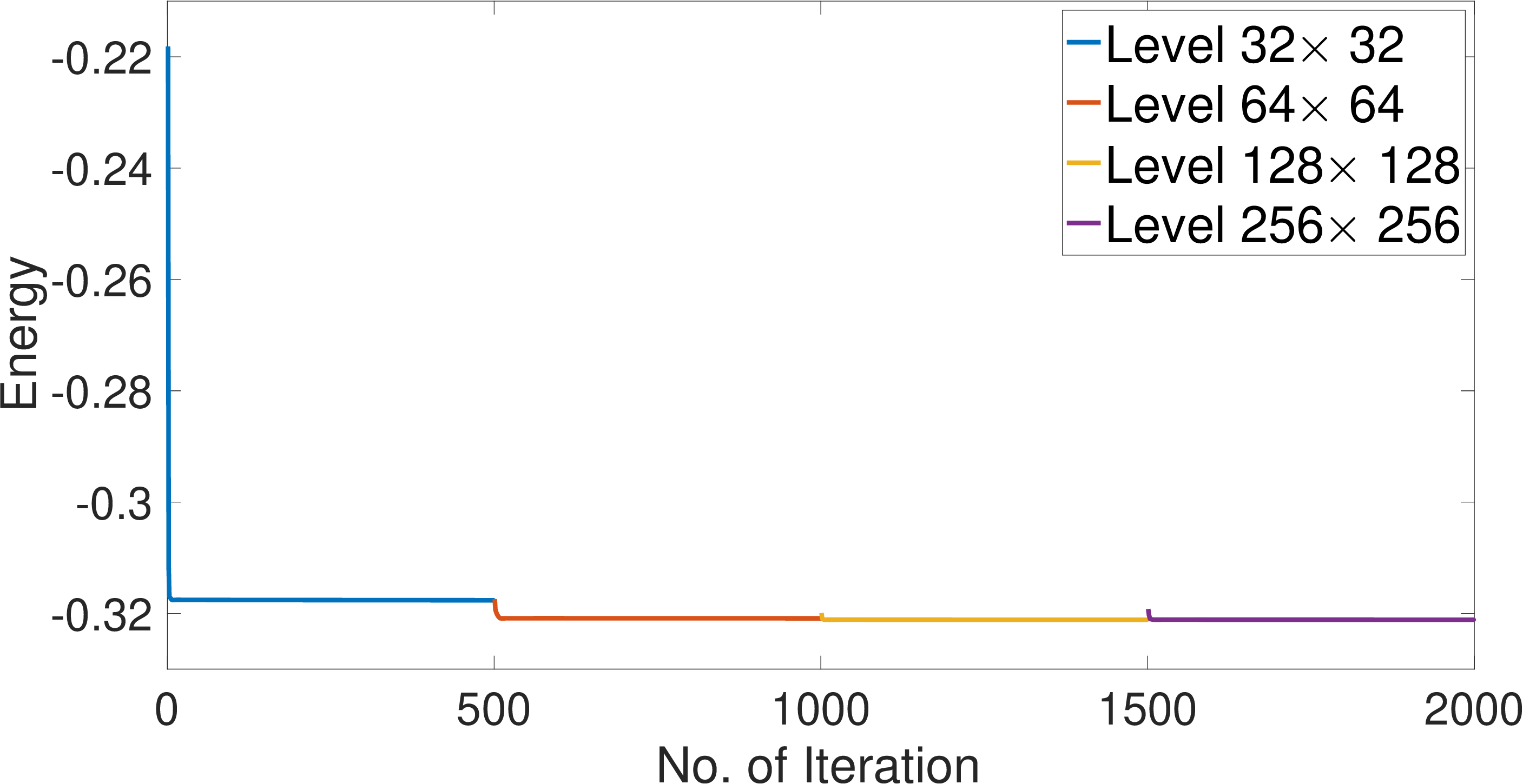}\hspace{0.5cm}
\includegraphics[width=2.0in,height=2.0in]{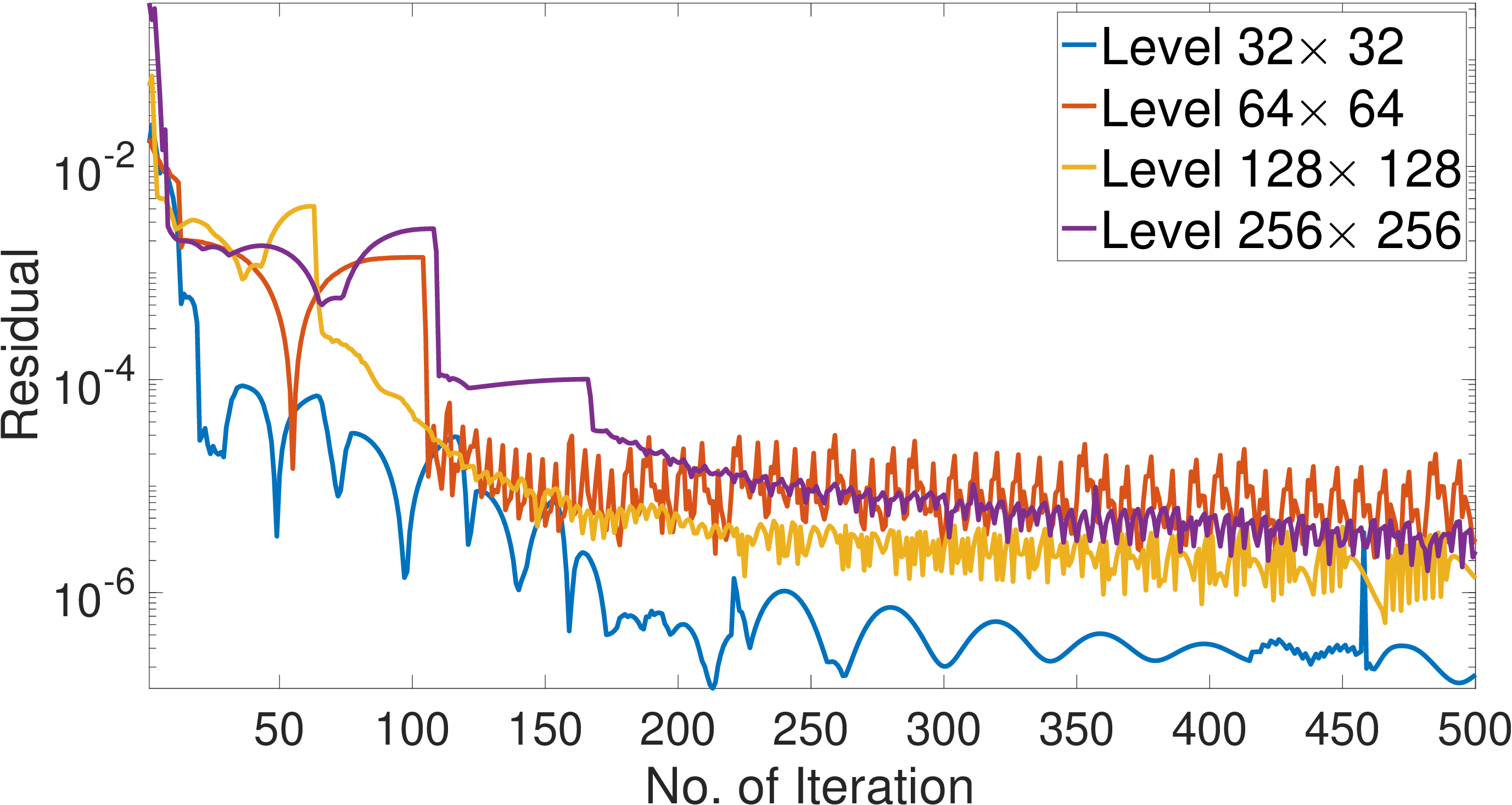}}\\
\subfigure[Energy and residual for the noise image: Case 2 ]{
\includegraphics[width=2.0in,height=2.0in]{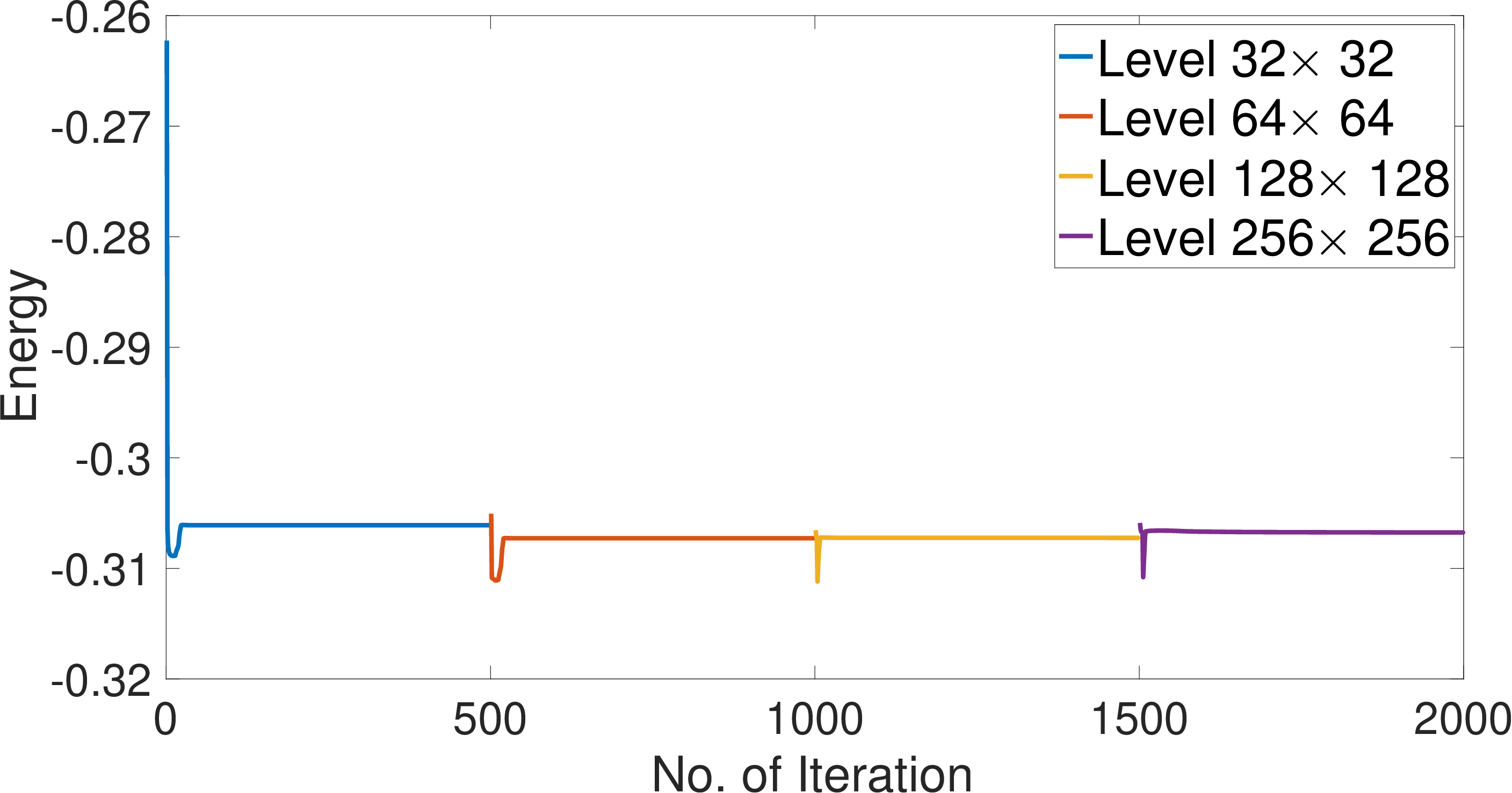}\hspace{0.5cm}
\includegraphics[width=2.0in,height=2.0in]{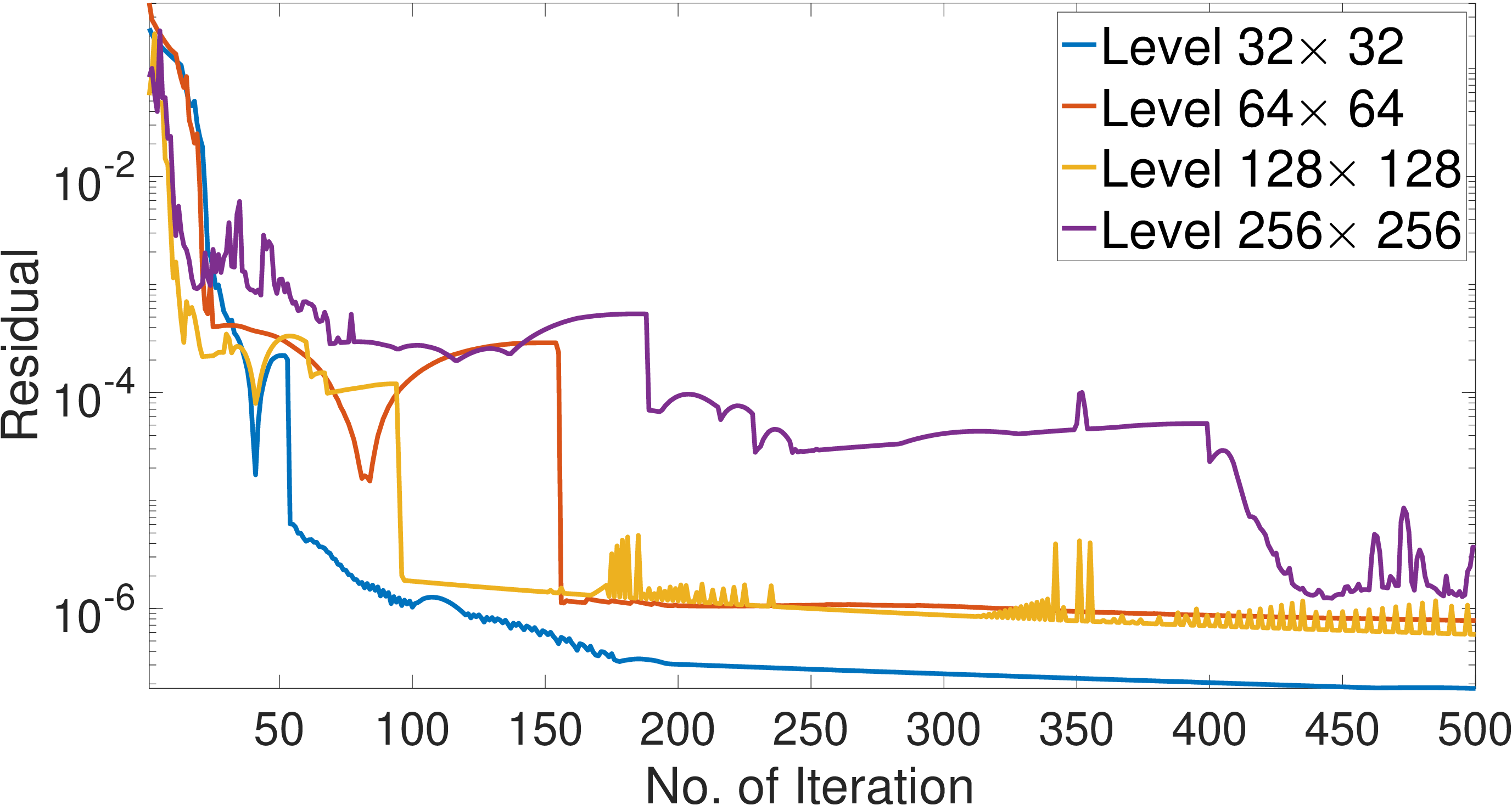}}
\caption{Energy and residuals versus iterations {\color{black} on per level} by segmenting the above noisy example with two different initial contours (two right subfigures in the first row of Fig. \ref{Exp1_fig1}).}\label{Exp1_convergence}
\end{figure}

{\color{black} To rigorously evaluate the convergence of Algorithm \ref{Alg1}, we conduct experiments on the above noisy image using two distinct initial contours, enforcing 500 outer iterations per level. Fig. \ref{Exp1_convergence} illustrates the evolution of both the objective energy and the residual across all levels. The results demonstrate clear numerical convergence: the energy stabilizes rapidly at each level, with only minor fluctuations observed beyond the coarsest level ($32\times32$). This stability is attributed to the multilevel strategy, which progressively refines the solution by leveraging coarse-level approximations as initial guesses for finer levels. Additionally, the residuals remain consistently below the strict threshold of $10^{-4}$ throughout all iterations, further confirming the algorithm’s  convergence under the tested conditions. These observations collectively provide strong empirical evidence for the reliability and efficiency of Algorithm \ref{Alg1}.}


\begin{figure}[htbp!]
\centering
\subfigure[Target image with the initial contour]{
\includegraphics[width=1.13in,height=1.13in]{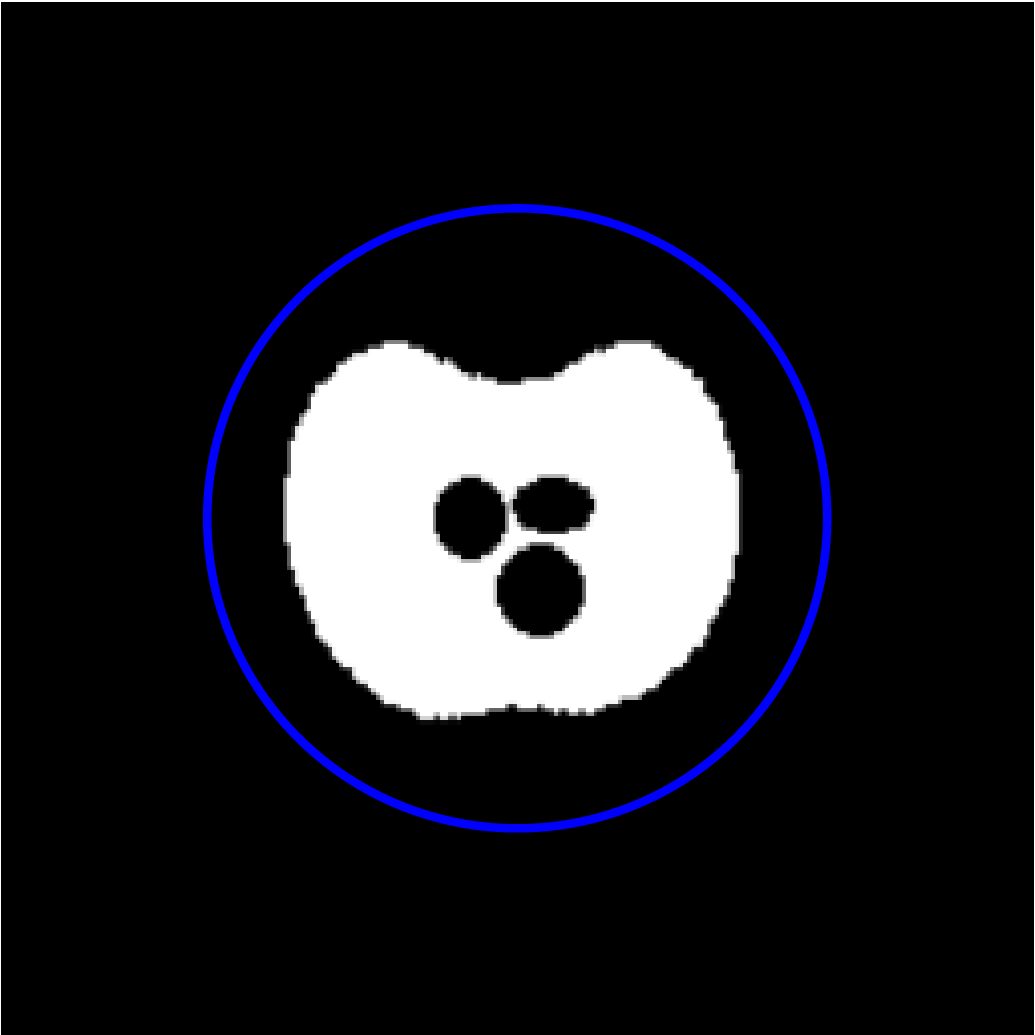}}
\hspace{0.5cm}
\subfigure[Center and restricted domain (left) and segmentation result (right). \textcolor{black}{Running time 2.8 seconds.}]{
\includegraphics[width=1.13in,height=1.13in]{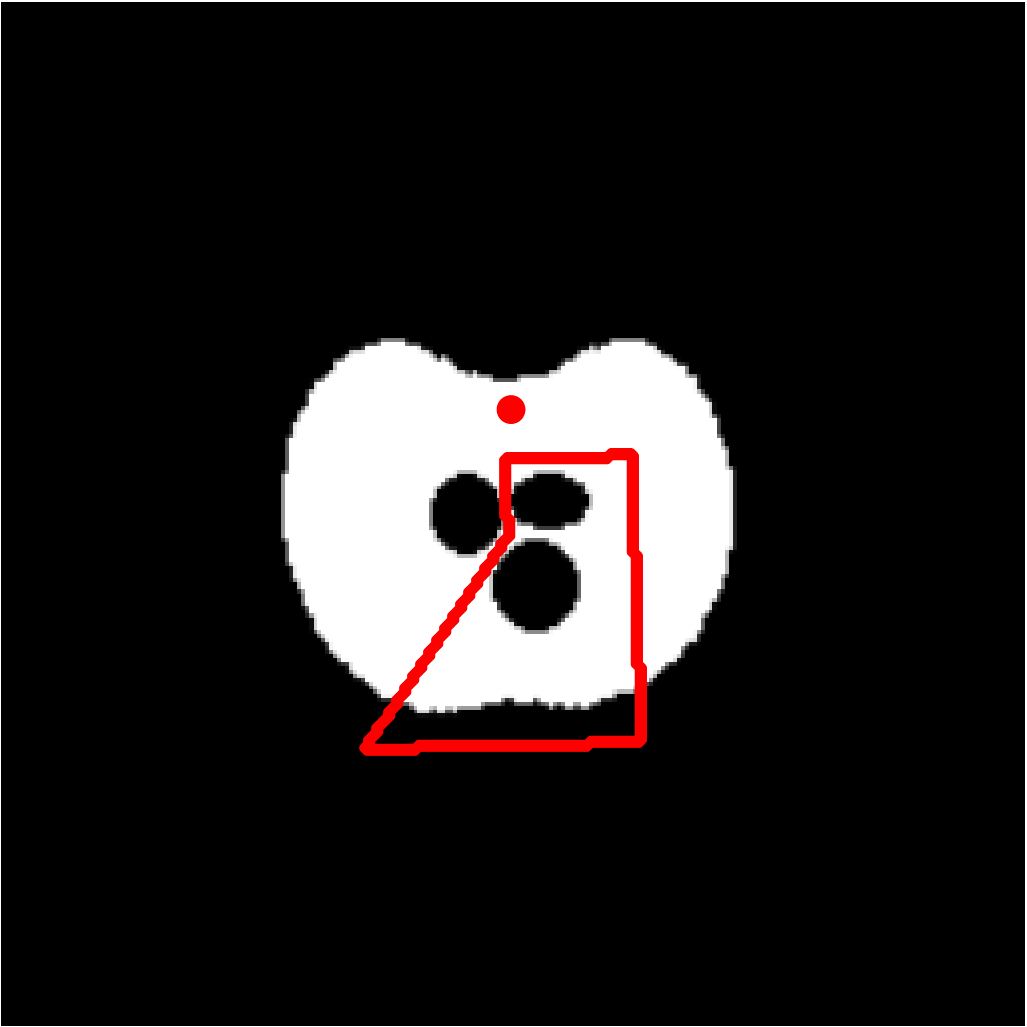}
\includegraphics[width=1.13in,height=1.13in]{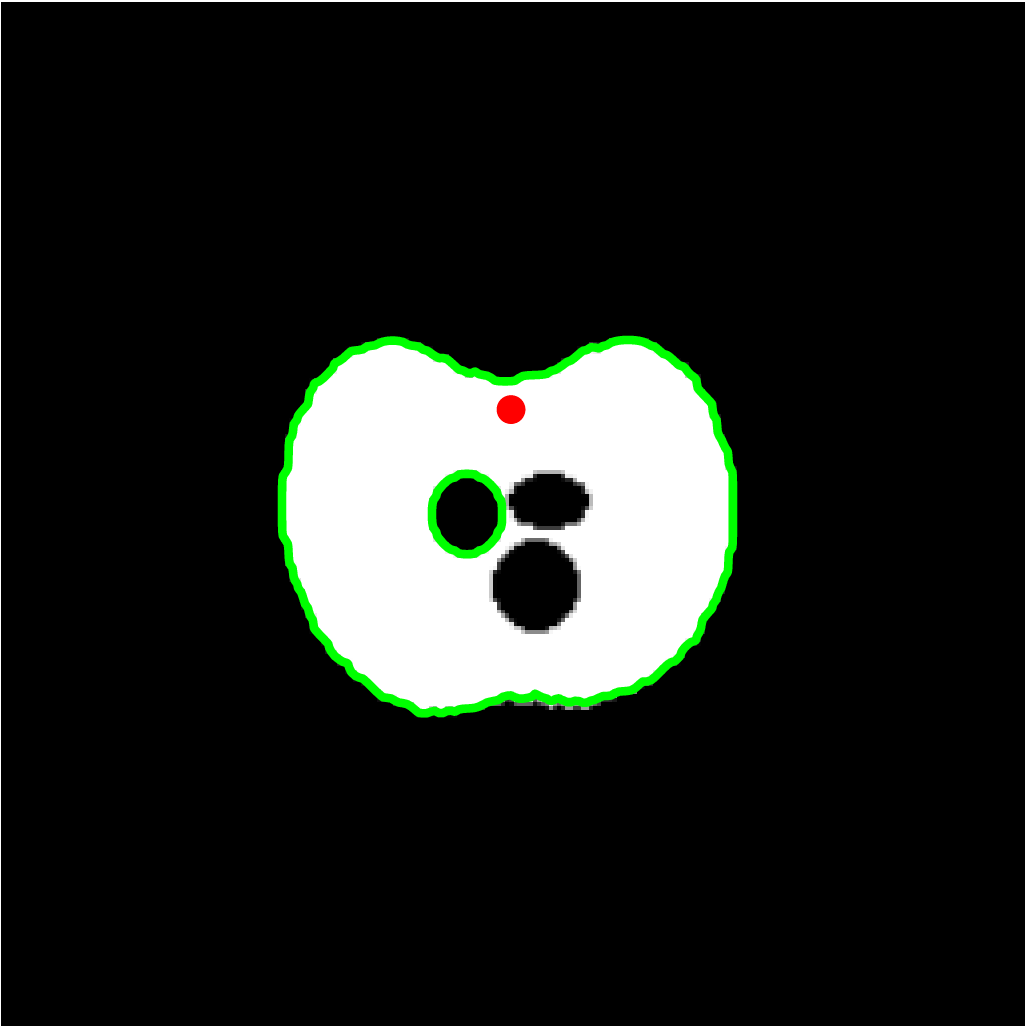}}
\hspace{0.5cm}
\subfigure[Center and restricted domain (left) and segmentation result (right). \textcolor{black}{Running time 3.8 seconds.}]{
\includegraphics[width=1.13in,height=1.13in]{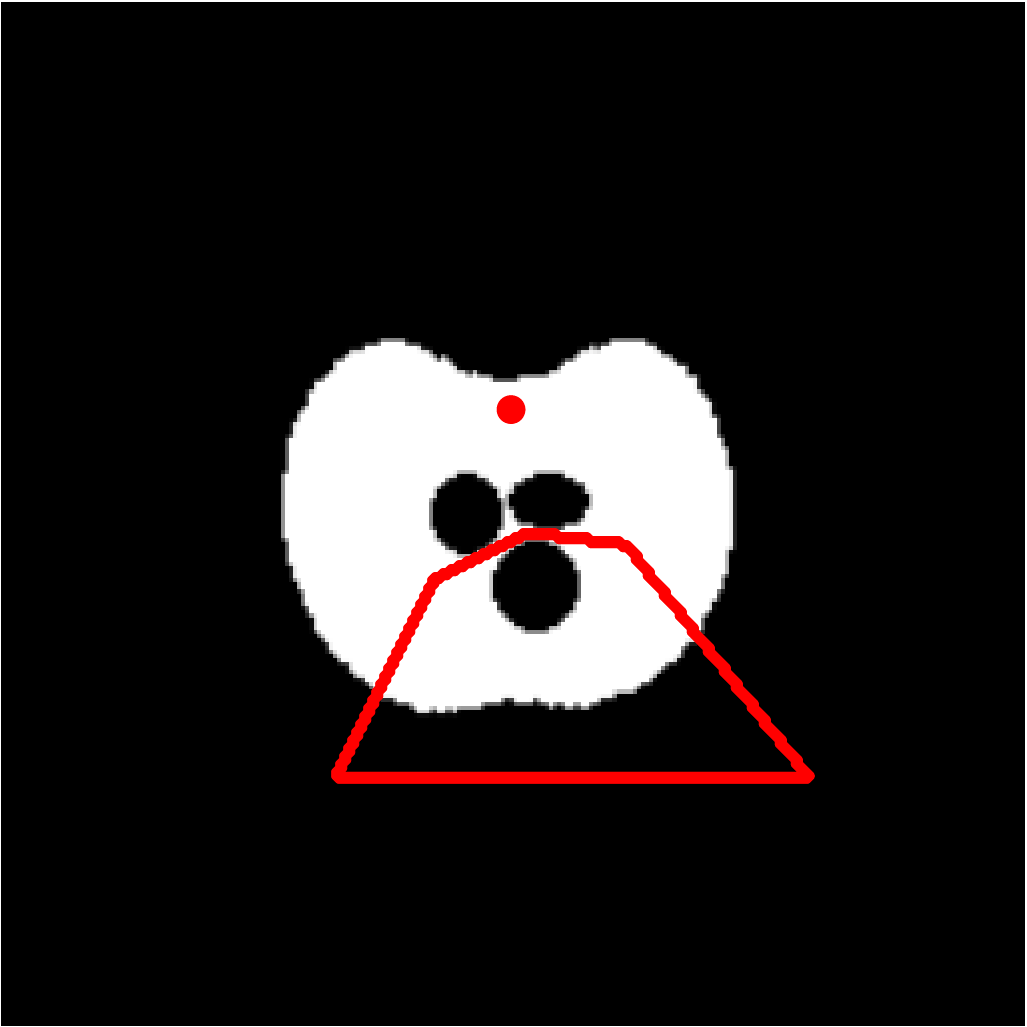}
\includegraphics[width=1.13in,height=1.13in]{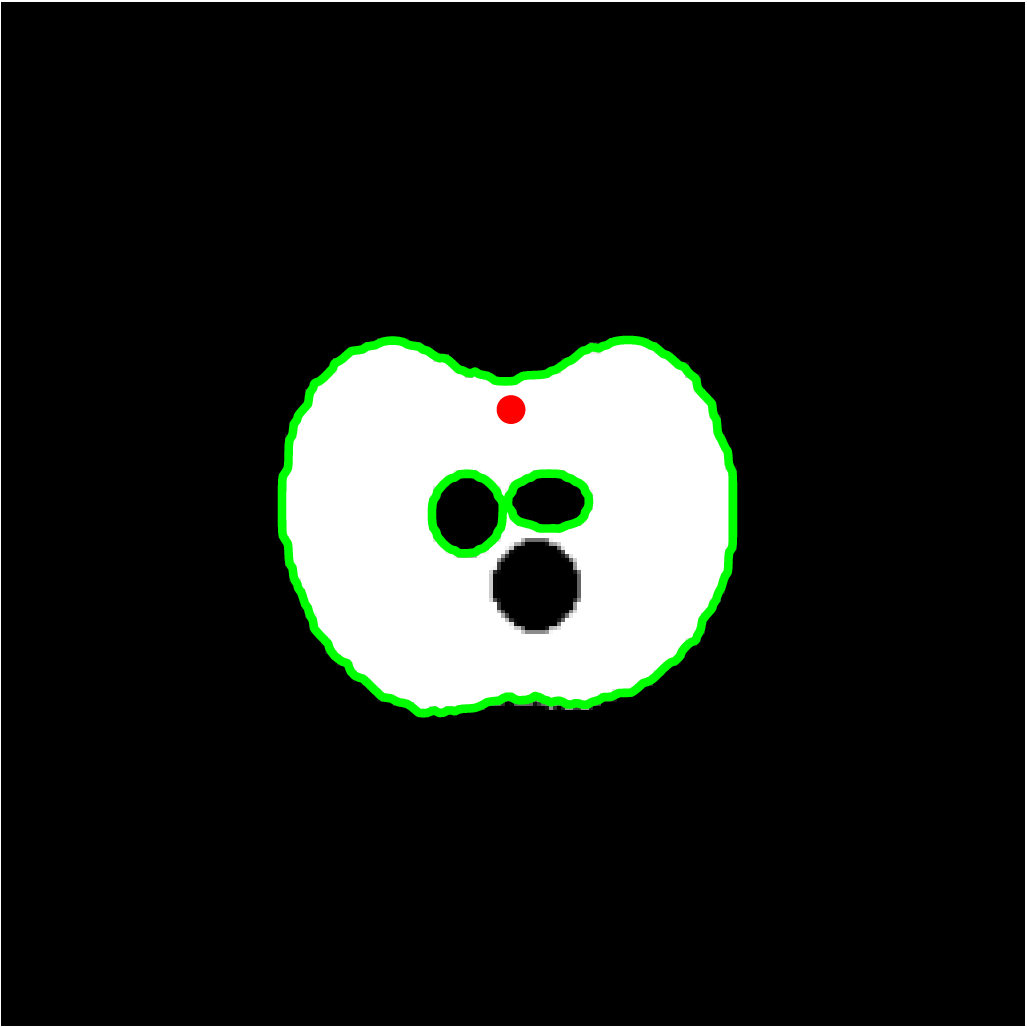}}
\caption{Test on the proposed model \eqref{proposed_model2}. The red point is the center and the domain inside the red curve is the corresponding constrained region with respect to the center.}\label{Exp1_fig2_1}
\end{figure}

\begin{figure}[htbp!]
\centering
\subfigure[Target image with the initial contour]{
\includegraphics[width=1.13in,height=1.13in]{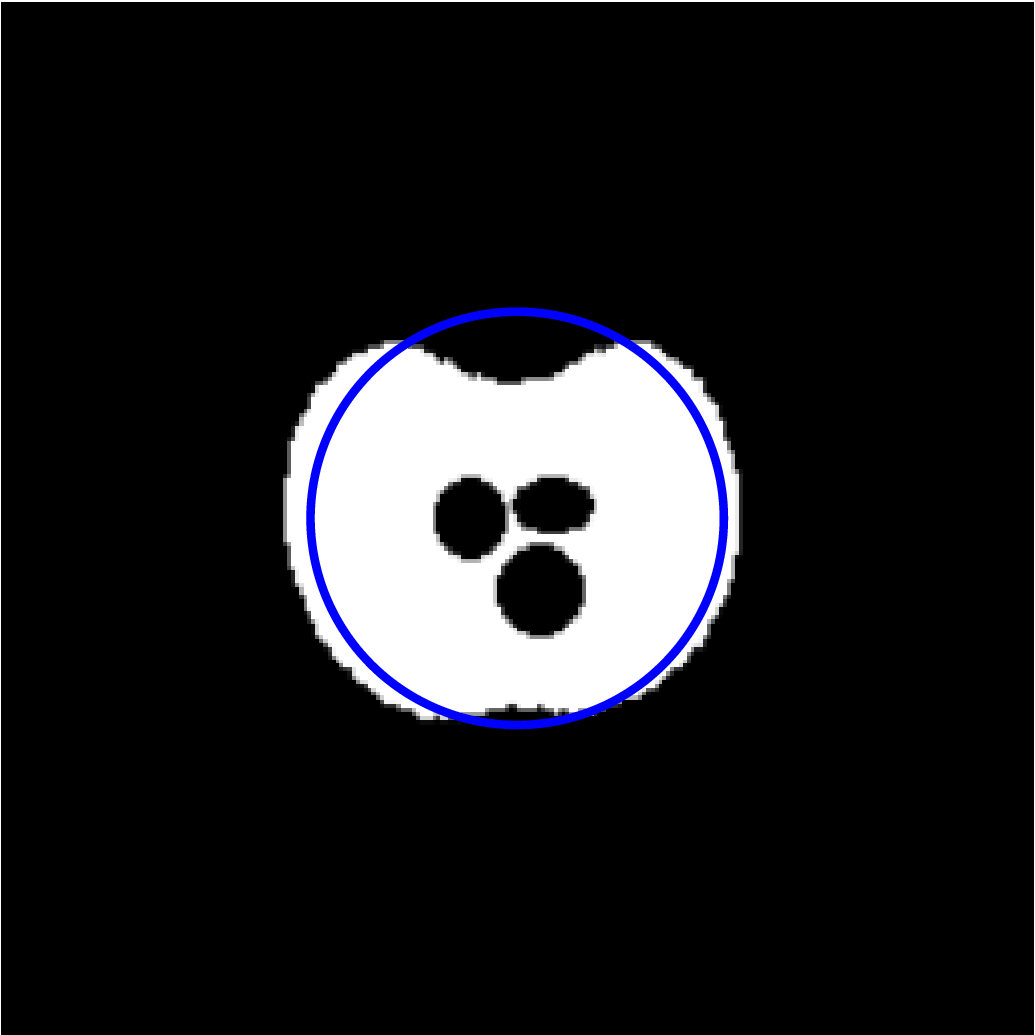}}
\hspace{0.5cm}
\subfigure[Centers and restricted domains (left) and segmentation result (right). \textcolor{black}{Running time 6.5 seconds.}]{
\includegraphics[width=1.13in,height=1.13in]{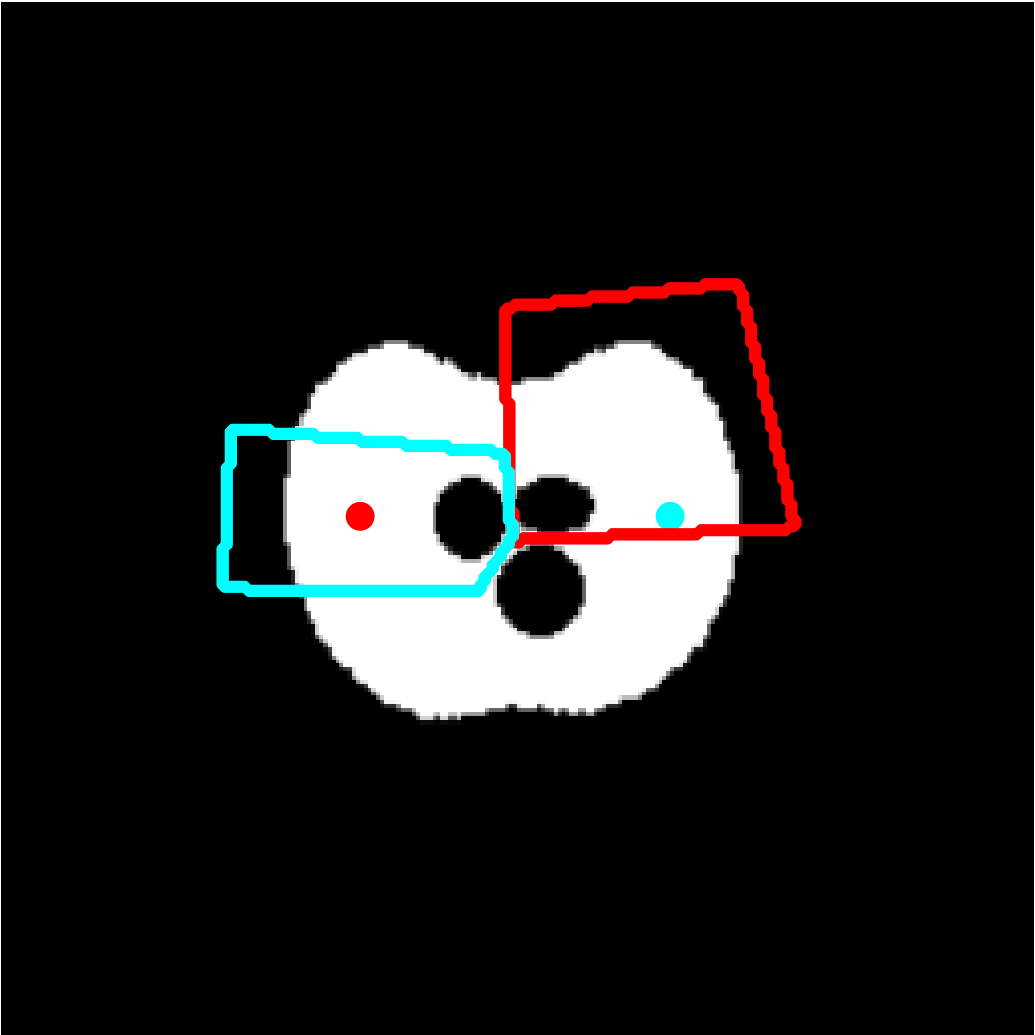}
\includegraphics[width=1.13in,height=1.13in]{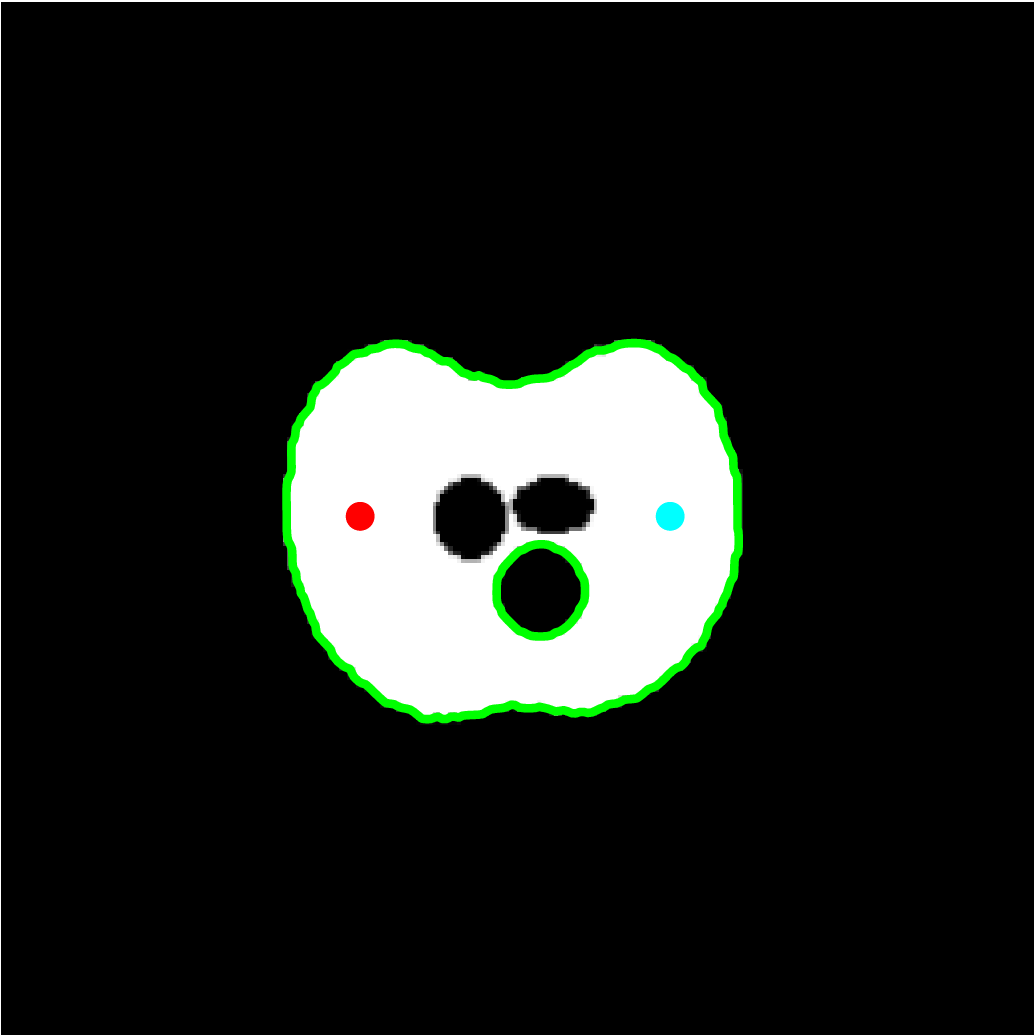}}
\hspace{0.5cm}
\subfigure[Centers and restricted domains (left) and segmentation result (right). \textcolor{black}{Running time 7.4 seconds.}]{
\includegraphics[width=1.13in,height=1.13in]{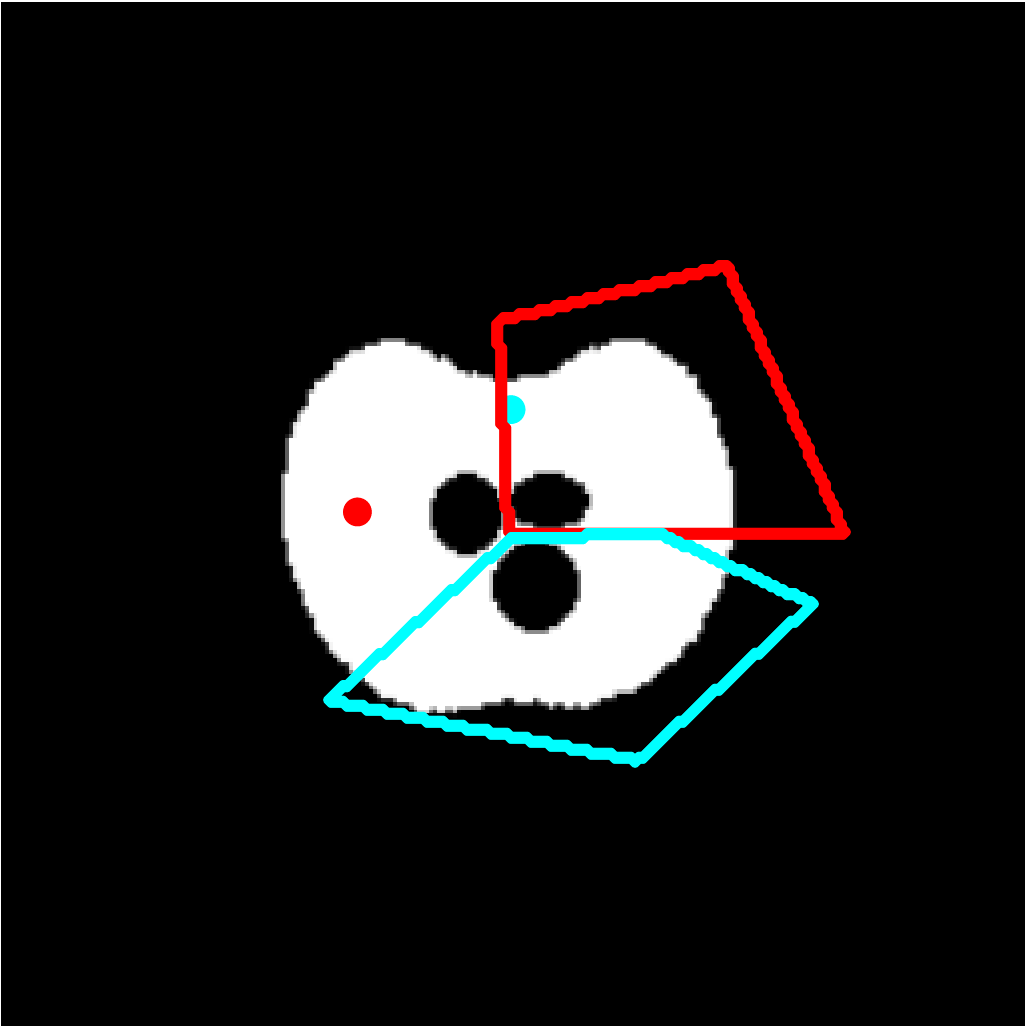}
\includegraphics[width=1.13in,height=1.13in]{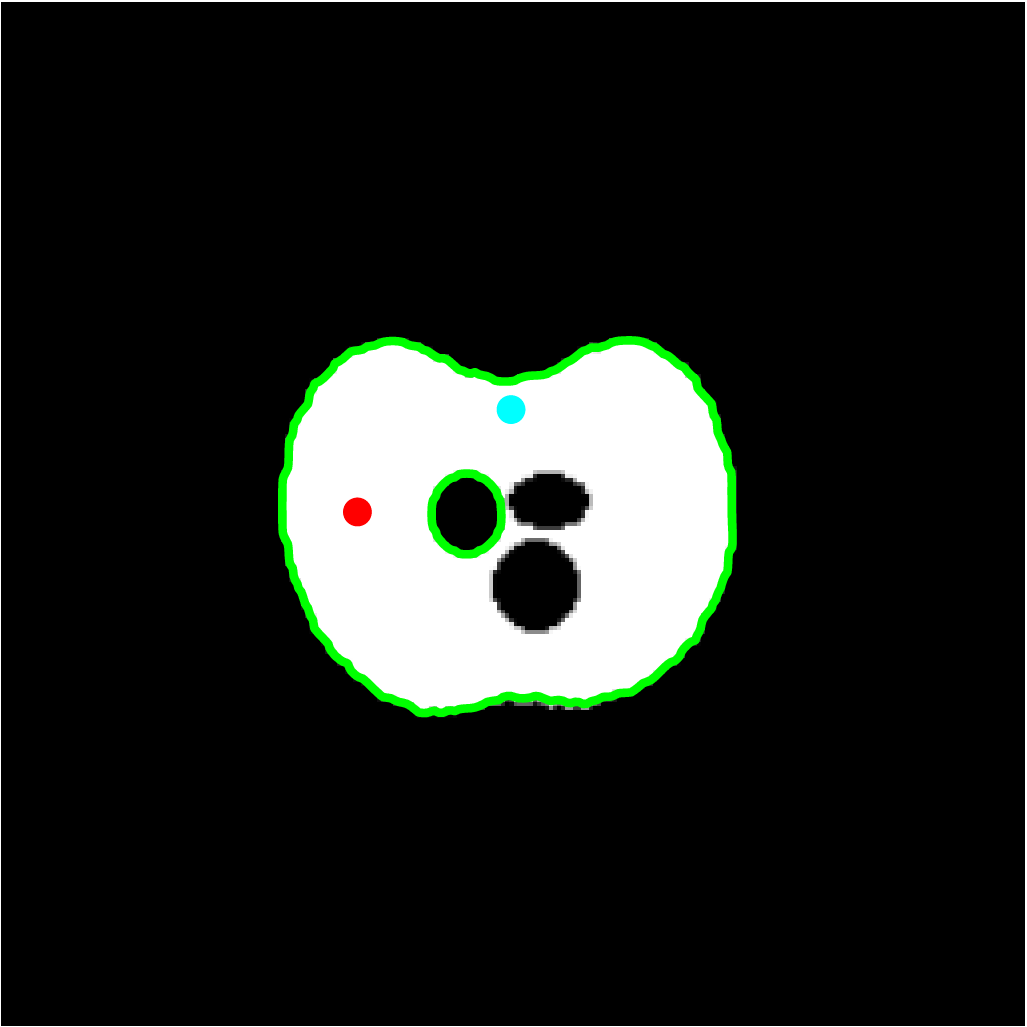}}
\caption{Test on the proposed model \eqref{proposed_model3}. 
Each center and its associated domain share the same color and form a pair.}\label{Exp1_fig2_2}
\end{figure}

The constrained region in the proposed model \eqref{proposed_model1} encompasses the entire domain $\Omega$. However, the constrained region in the proposed model \eqref{proposed_model2} confines to a specific portion of the overall domain, allowing for a more flexible segmentation. In Fig. \ref{Exp1_fig2_1}(b-c), a center point and its corresponding constrained region (red point and region inside the red curve) are specified. The prescribed region effectively ensures that the segmentation within it adheres to a star-shape configuration. The distinction between the proposed models \eqref{proposed_model2} and \eqref{proposed_model3} lies in the number of centers that they incorporate. The proposed model \eqref{proposed_model2} features a single center, producing a partially star-shape segmentation, wherein any radial line from the designated center intersects the segmentation boundary within the constrained region only once. On the other hand, the proposed model \eqref{proposed_model3} incorporates multiple centers, resulting in a segmentation where any radial line from any center intersects the segmentation boundary within the associated region only once. This segmentation can be conceptualized as the intersection of several partially star-shape segmentations. In Fig. \ref{Exp1_fig2_2}, we present examples with two centers and corresponding constrained regions. The segmentation results indeed exhibit a generalized partially star-shape domain concerning all given centers. \textcolor{black}{Clearly, the convexity-preserving segmentation model \cite{zhang2021topology2} lacks the flexibility to produce such diverse segmentation results.}

\begin{figure}[htbp!]
\centering
\subfigure[Target image with the initial contour]{
\includegraphics[width=1.13in,height=1.13in]{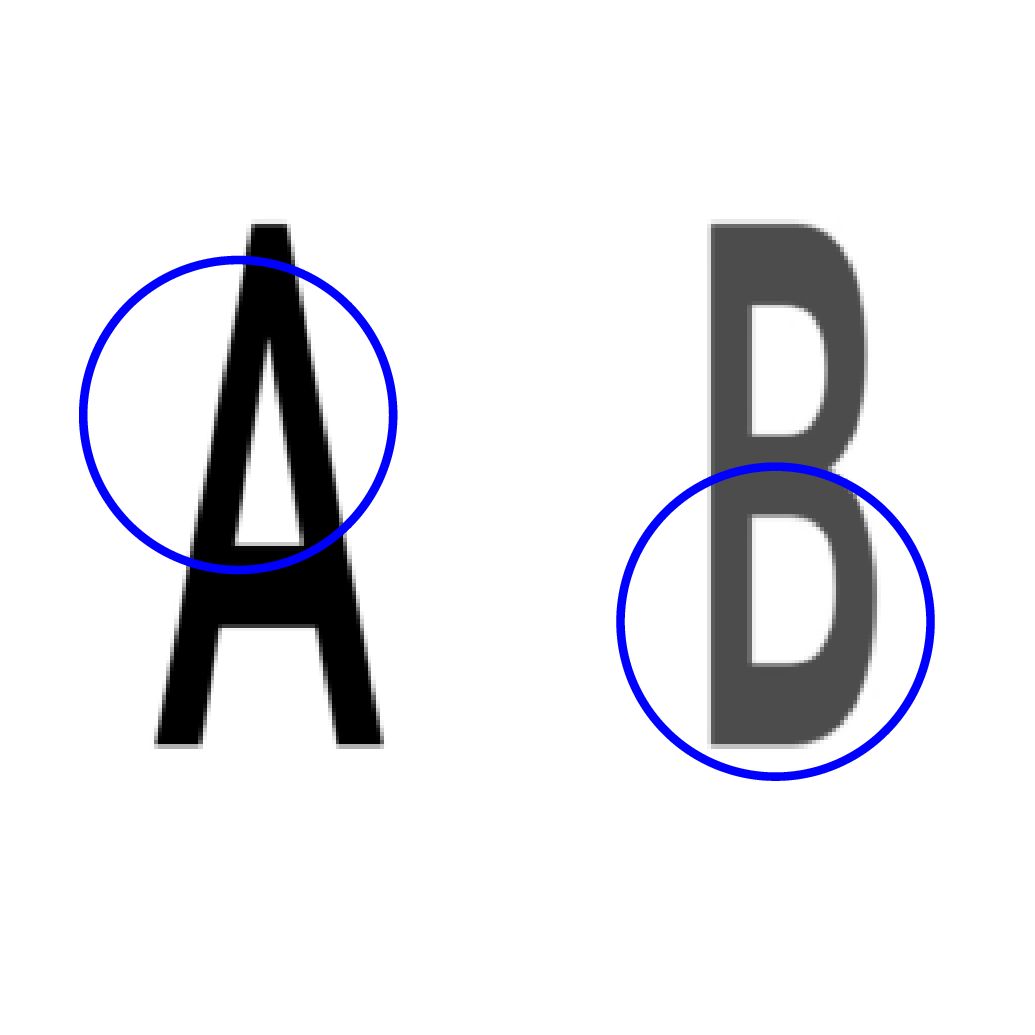}}
\hspace{0.5cm}
\subfigure[Segmentation by CV. \textcolor{black}{Running time 3.6 seconds.}]{
\includegraphics[width=1.13in,height=1.13in]{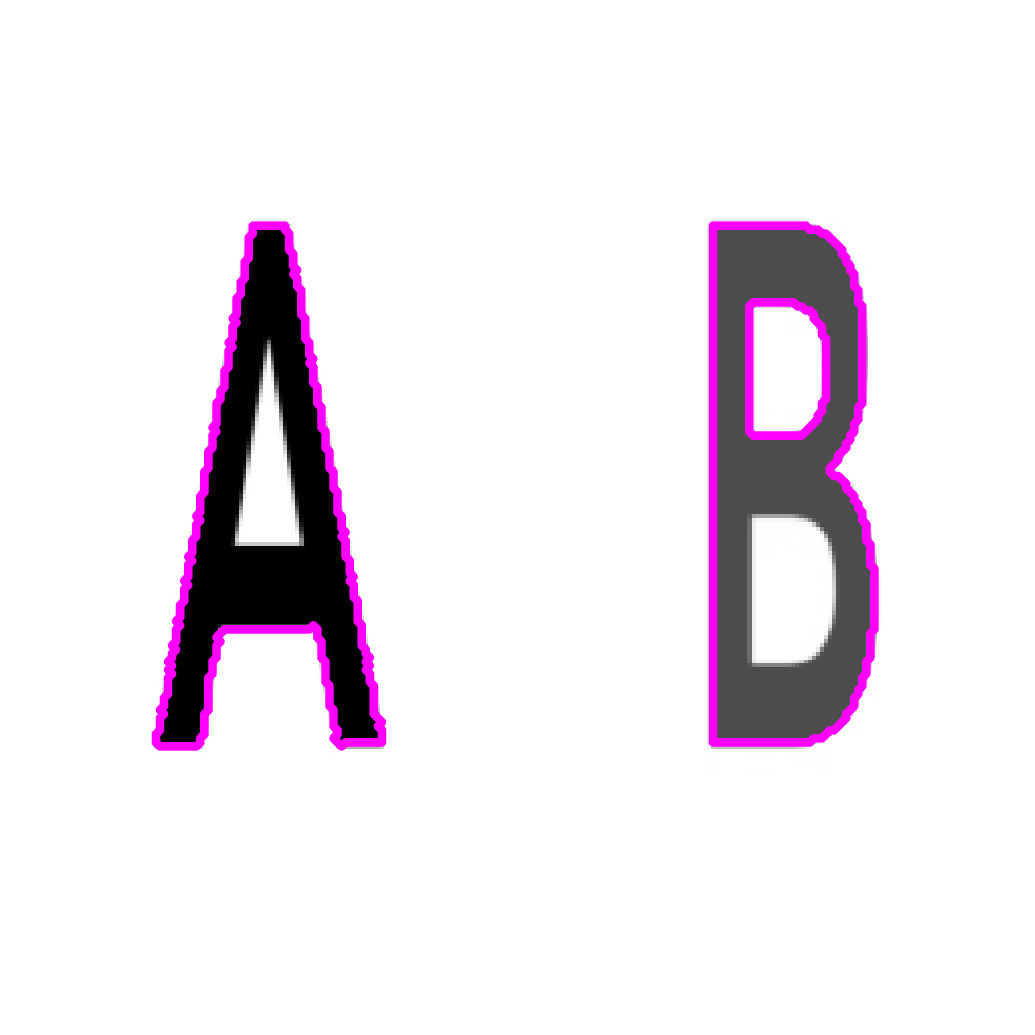}}
\hspace{0.5cm}
\subfigure[\textcolor{black}{Segmentation by the convexity-preserving model. Running time 48.9 seconds.}]{
\includegraphics[width=1.13in,height=1.13in]{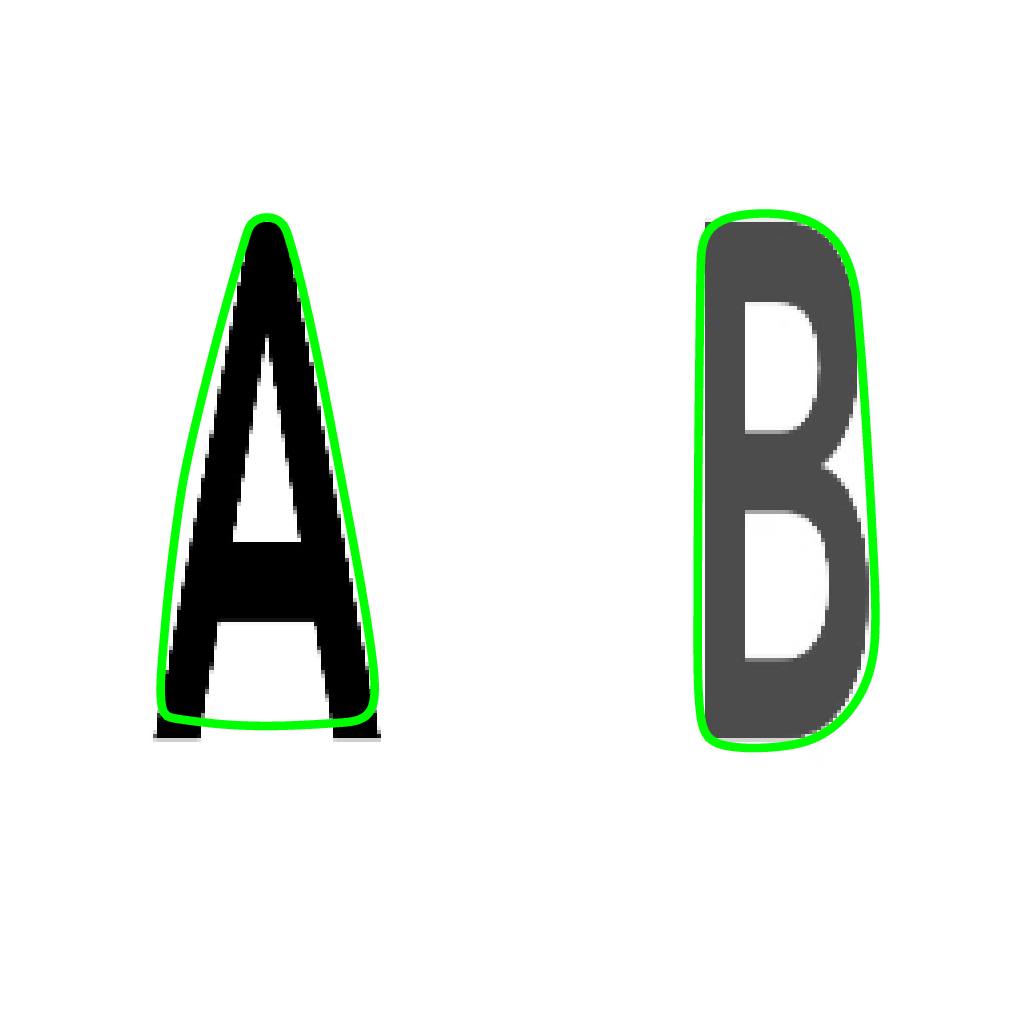}} \\
\subfigure[Segmentation by \eqref{proposed_model4} with one center. \textcolor{black}{Running time 4.1 seconds.} ]{
\includegraphics[width=1.13in,height=1.13in]{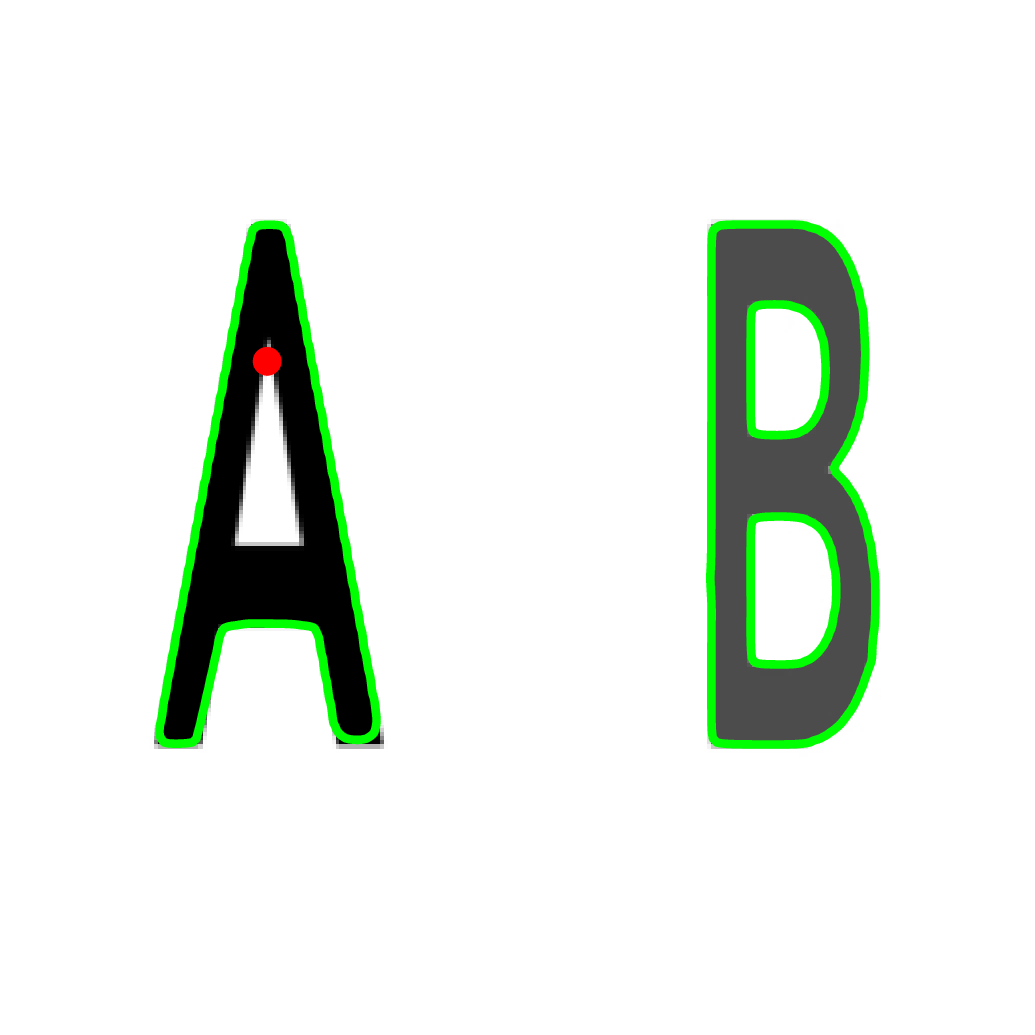}}
\hspace{0.5cm}
\subfigure[Segmentation by \eqref{proposed_model4} with one center. \textcolor{black}{Running time 6.1 seconds.}]{
\includegraphics[width=1.13in,height=1.13in]{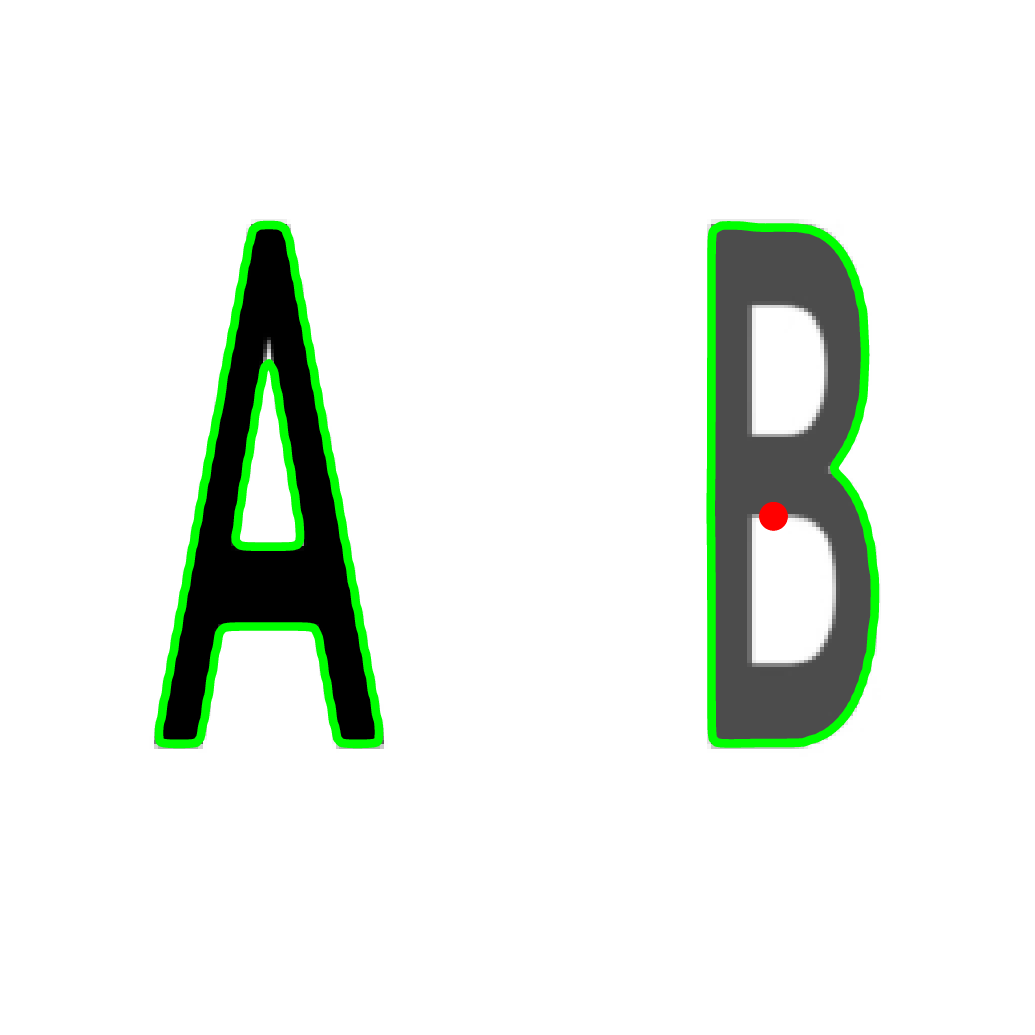}}
\hspace{0.5cm}
\subfigure[Segmentation by \eqref{proposed_model4} with two centers. \textcolor{black}{Running time 7.1 seconds.}]{
\includegraphics[width=1.13in,height=1.13in]{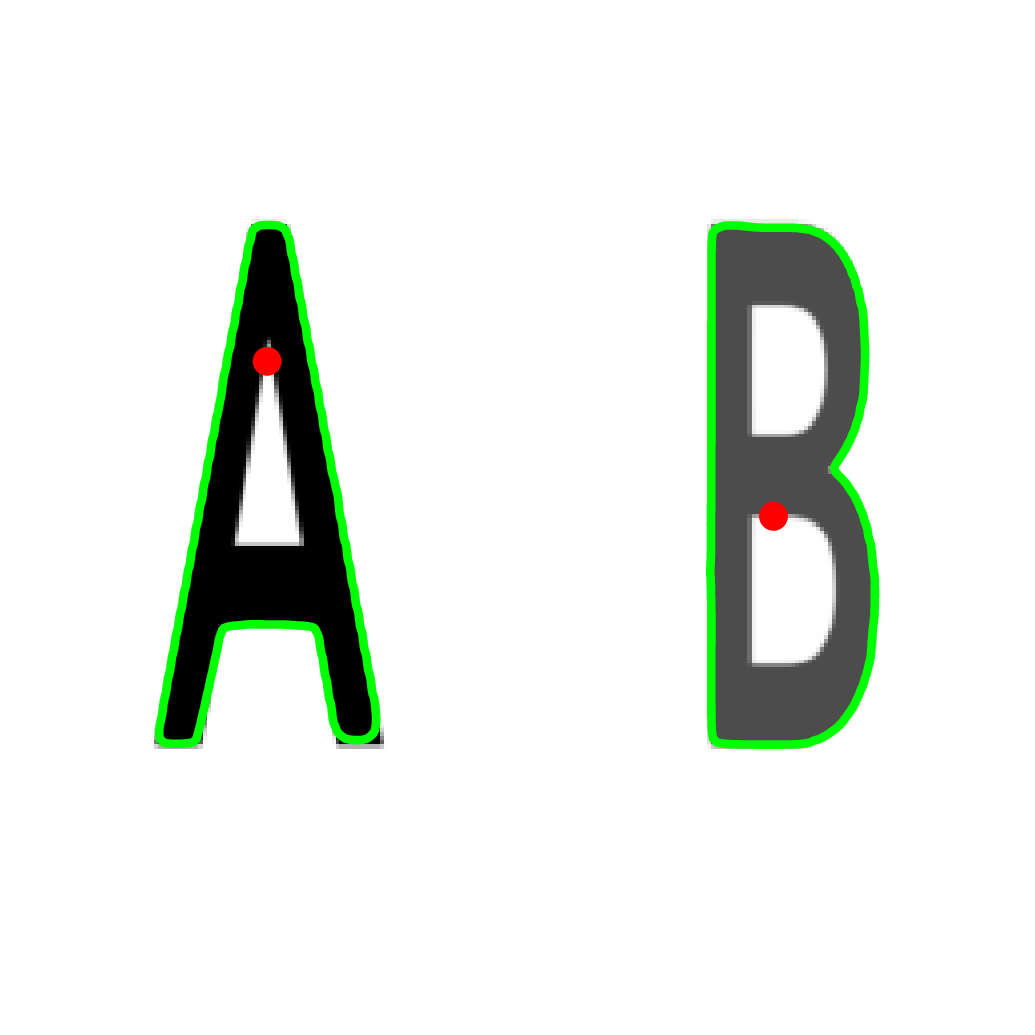}}
\caption{Test on the proposed model \eqref{proposed_model4}. (a) Target image with the initial contour. (b) Segmentation result by the CV model. \textcolor{black}{(c) Segmentation result by the convexity-preserving model \cite{zhang2021topology2}.} (d-f) Segmentation results by the proposed model \eqref{proposed_model4} with different centers (red point). Here, by setting the suitable centers, we can let the segmentation of `A', `B' or both `A' and `B' be star-shape, respectively.}
\label{Exp1_fig3}
\end{figure}

The proposed models \eqref{proposed_model1}-\eqref{proposed_model3} focus on two phase segmentation. To deal with the multiphase segmentation, the model \eqref{proposed_model4} is proposed and can segment two regions, where one is star-shape and the other one is not. It can also obtain two star-shape domains by simply adding one more constraint. Fig. \ref{Exp1_fig3}(a) displays an example with the initial contours. We observe that the proposed model \eqref{proposed_model4} effectively segment two regions, where one is star-shape in Fig. \ref{Exp1_fig3}\textcolor{black}{(d-e)} with one center and both two are star-shape in Fig. \ref{Exp1_fig3}\textcolor{black}{(f)} with two centers. However, the CV model depends on the initial contour and does not freely extract the star-shape domain by the user's requirement. \textcolor{black}{The convexity-preserving model \cite{zhang2021topology2} only leads to the convex hull of `A' and `B' in Fig. \ref{Exp1_fig3}(c).} This example especially shows the advantage of our method when we need to identify the outer boundaries of one or multiple star-shape objects and do not care about the interior structures of the objects.  Such kind of applications happen often in practice.  

\begin{figure}[htbp!]
\centering
\subfigure[\textcolor{black}{Target image with the initial contour}]{
\includegraphics[width=1.15in,height=1.15in]{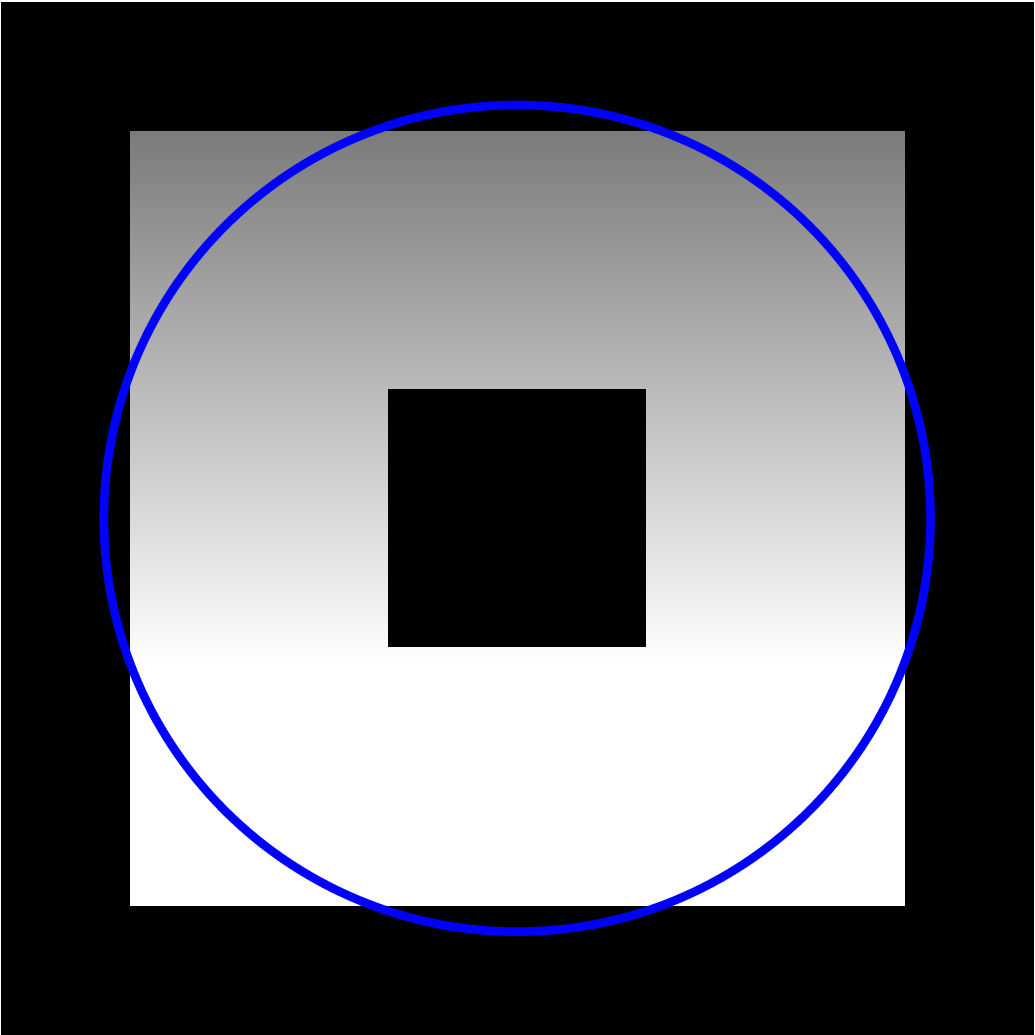}}\hspace{0.5cm}
\subfigure[\textcolor{black}{Segmentation results by the convexity-preserving model (left) and \eqref{proposed_model1} (right). Running times are 20.8 seconds and 0.8 seconds.}]{
\includegraphics[width=1.15in,height=1.15in]{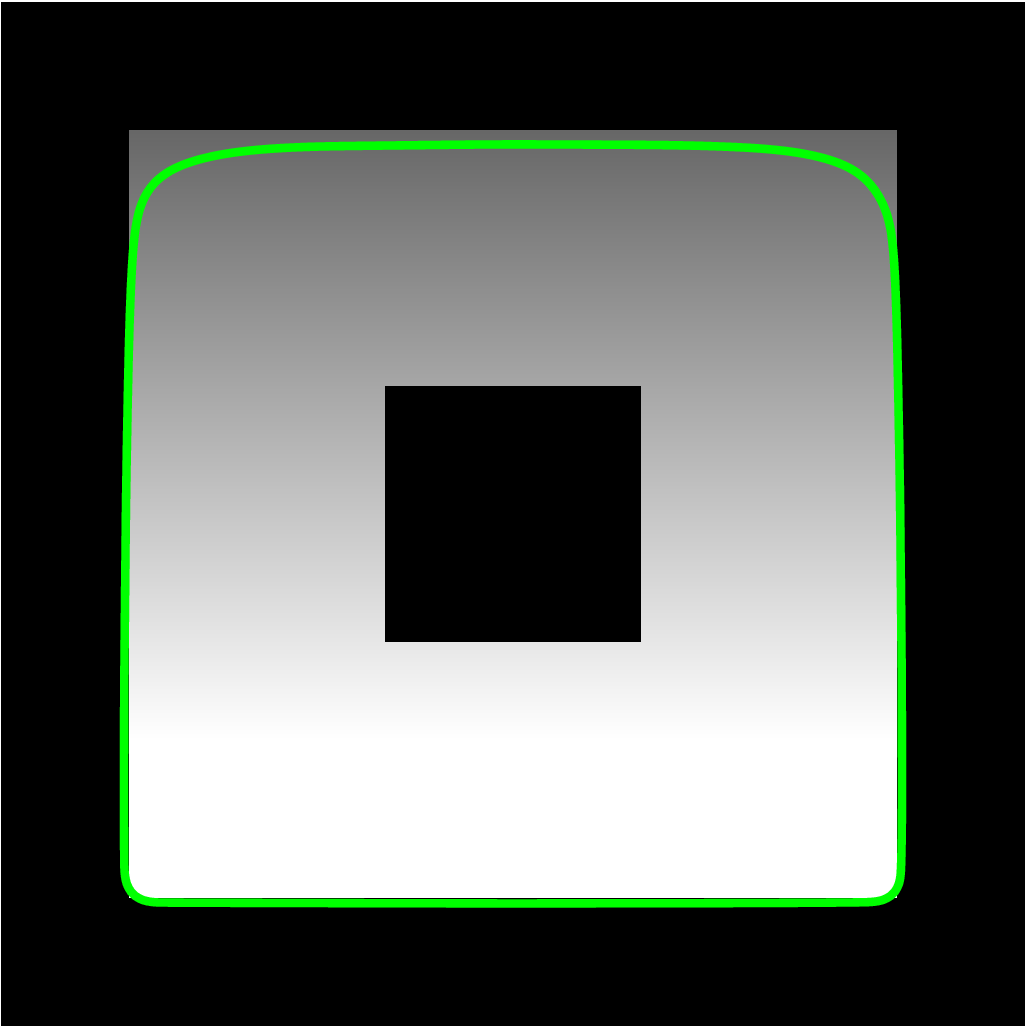}
\includegraphics[width=1.15in,height=1.15in]{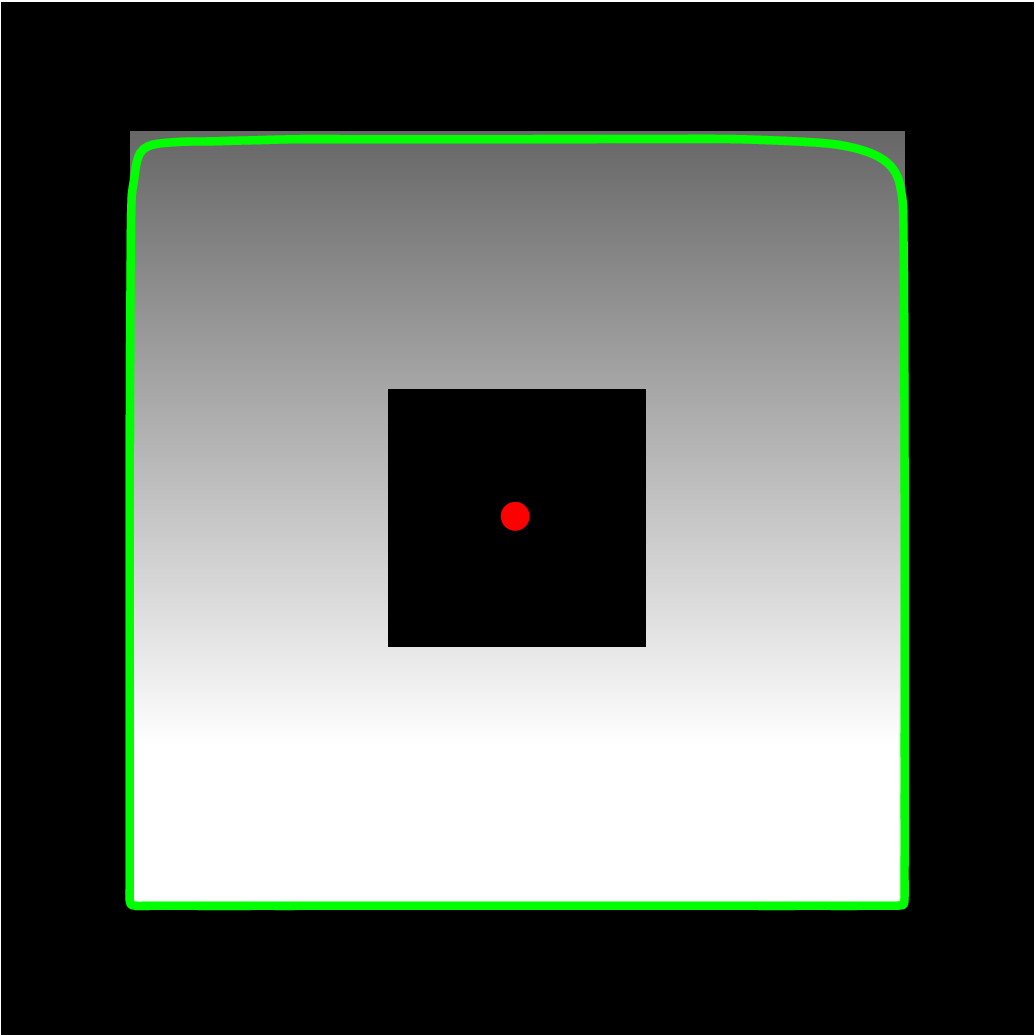}}
\hspace{0.5cm}
\subfigure[Target image with the initial contour and landmarks (left) and segmentation result by \eqref{proposed_model5} (right). \textcolor{black}{Running time 0.7 seconds.}]{
\includegraphics[width=1.15in,height=1.15in]{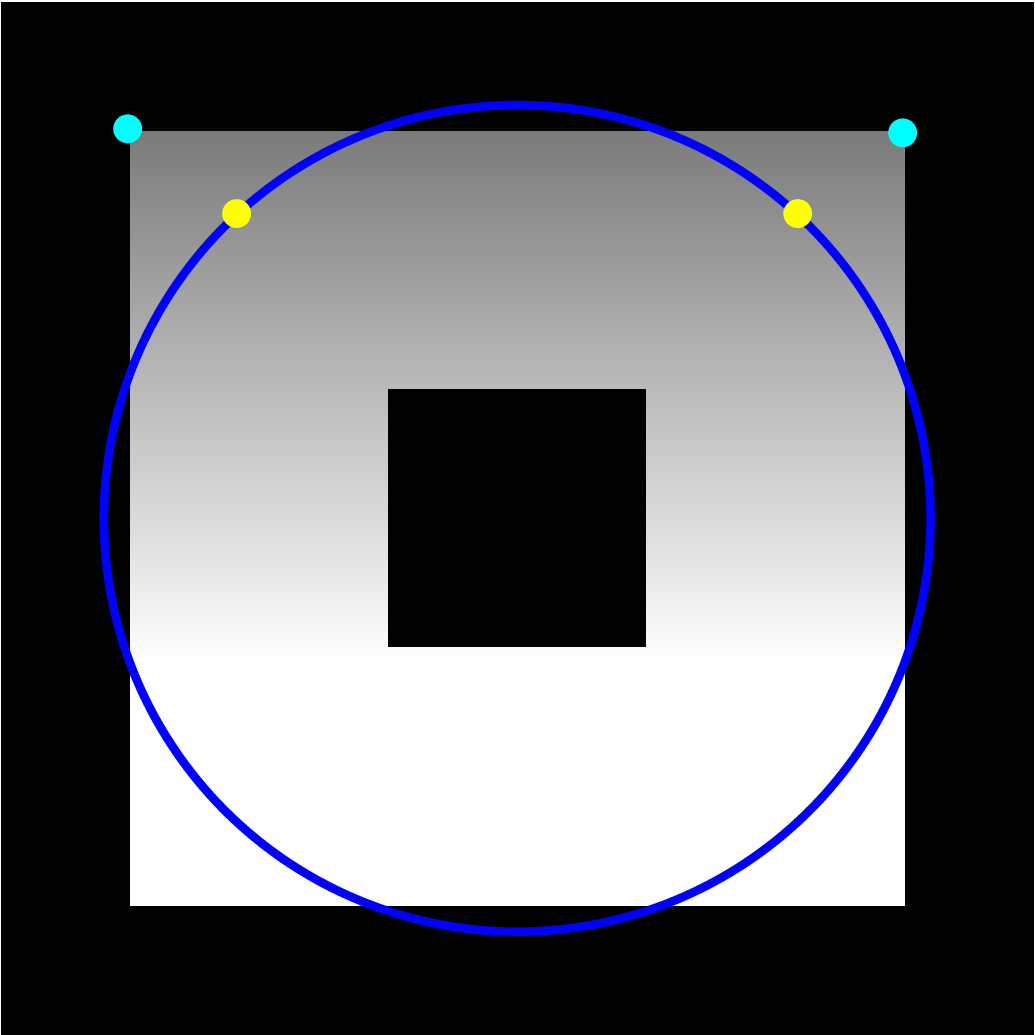}
\includegraphics[width=1.15in,height=1.15in]{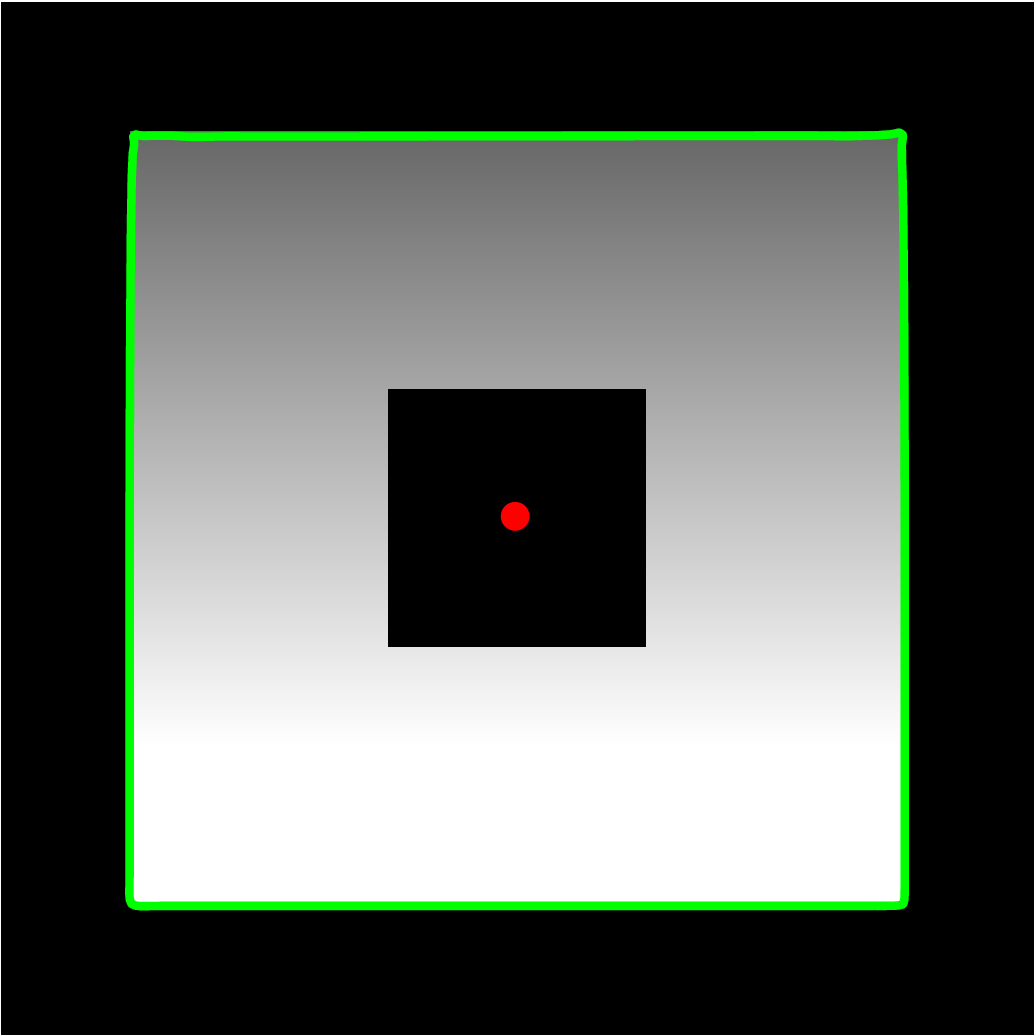}}
\caption{Test on the proposed model \eqref{proposed_model5}. \textcolor{black}{(a) Target image with the initial contour. (b) Segmentation results by the convexity-preserving model \cite{zhang2021topology2} (left) and \eqref{proposed_model1} (right).} The red point is the center. (c) Target image with the initial contour and landmarks (left) and segmentation result by \eqref{proposed_model5} (right). The red point is the center. The yellow and cyan points form two pairs of landmarks. }\label{Exp1_fig4}
\end{figure}

\textcolor{black}{Due to the inhomogeneous intensity distribution, the convexity-preserving model \cite{zhang2021topology2} fails to achieve an accurate segmentation result, as illustrated in Fig. \ref{Exp1_fig4}(b). Although the proposed model \eqref{proposed_model1} demonstrates some improvement through its fitting term, the segmentation performance remains unsatisfactory.}
But for this example, we clearly know that the final contour should pass through the corners. Thus, we add the landmark constraints and employ the proposed model \eqref{proposed_model5} to get an accurate segmentation, displayed in Fig. \ref{Exp1_fig4}(c). Since the proposed models in this paper employ the registration framework, adding the landmark constraints will be very straightforward. This also reflects the flexibility and advantage of the registration-based segmentation method.

\subsection{The performance of the proposed models \eqref{proposed_model1}-\eqref{proposed_model5} on real images}

In this part, we use some real images to assess the effectiveness of the proposed models \eqref{proposed_model1}-\eqref{proposed_model5}.

\begin{figure}[htbp!]
\centering
\subfigure[Target images and initial contours]{
\includegraphics[width=1.15in,height=1.15in]{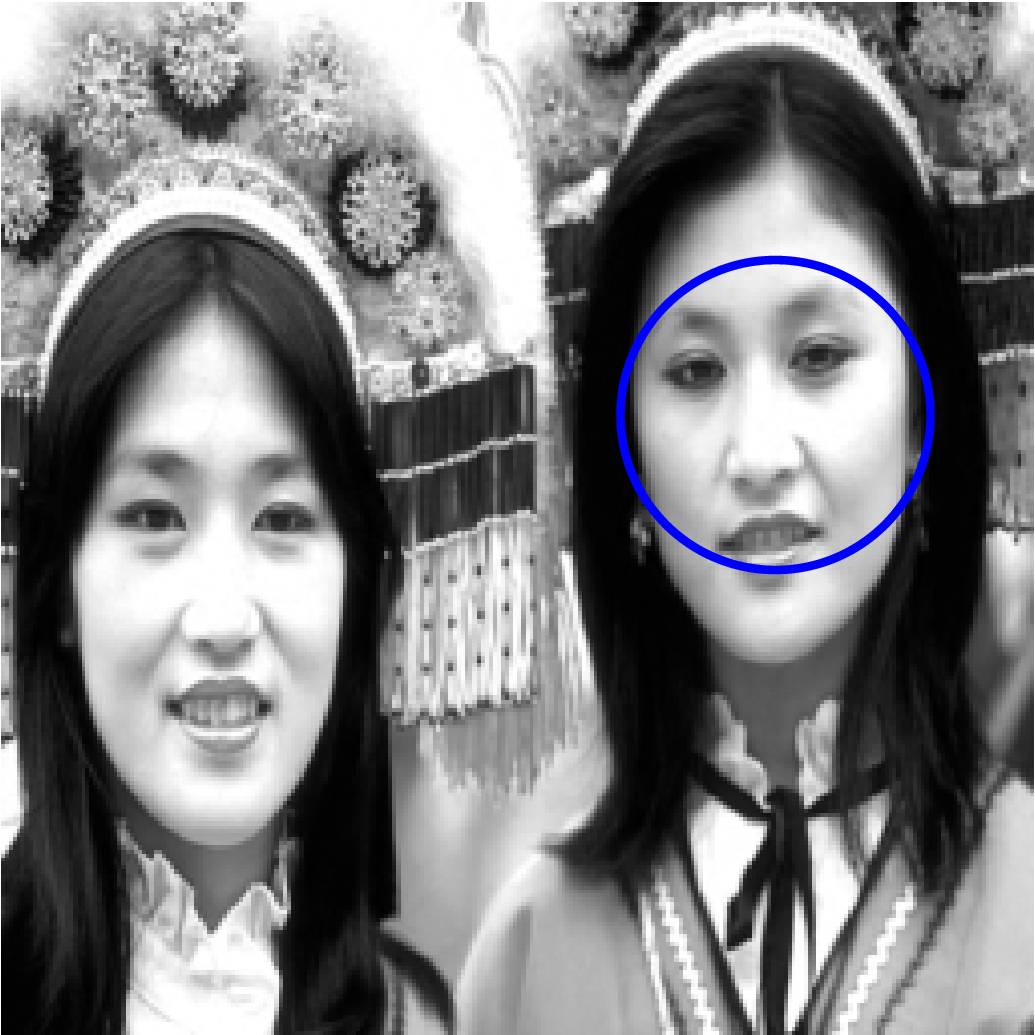}
\includegraphics[width=1.15in,height=1.15in]{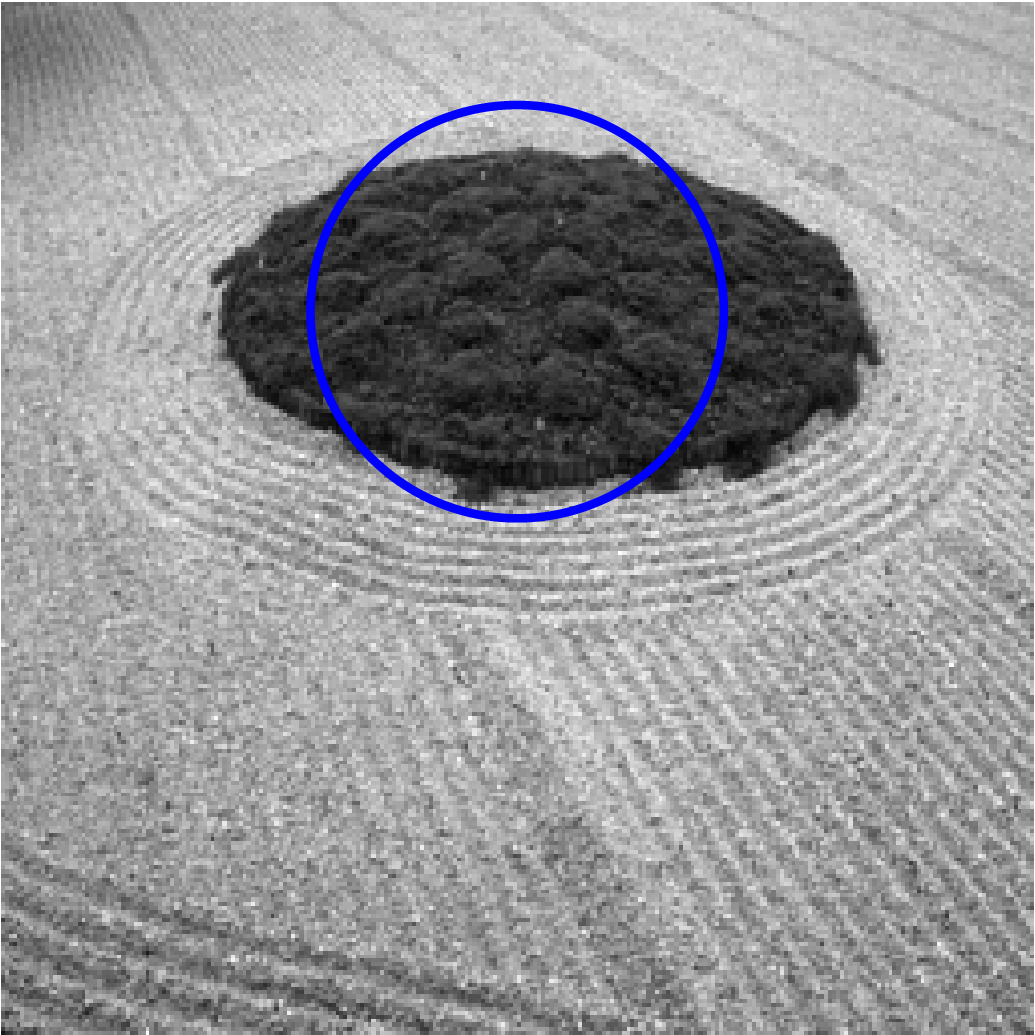}
\includegraphics[width=1.15in,height=1.15in]{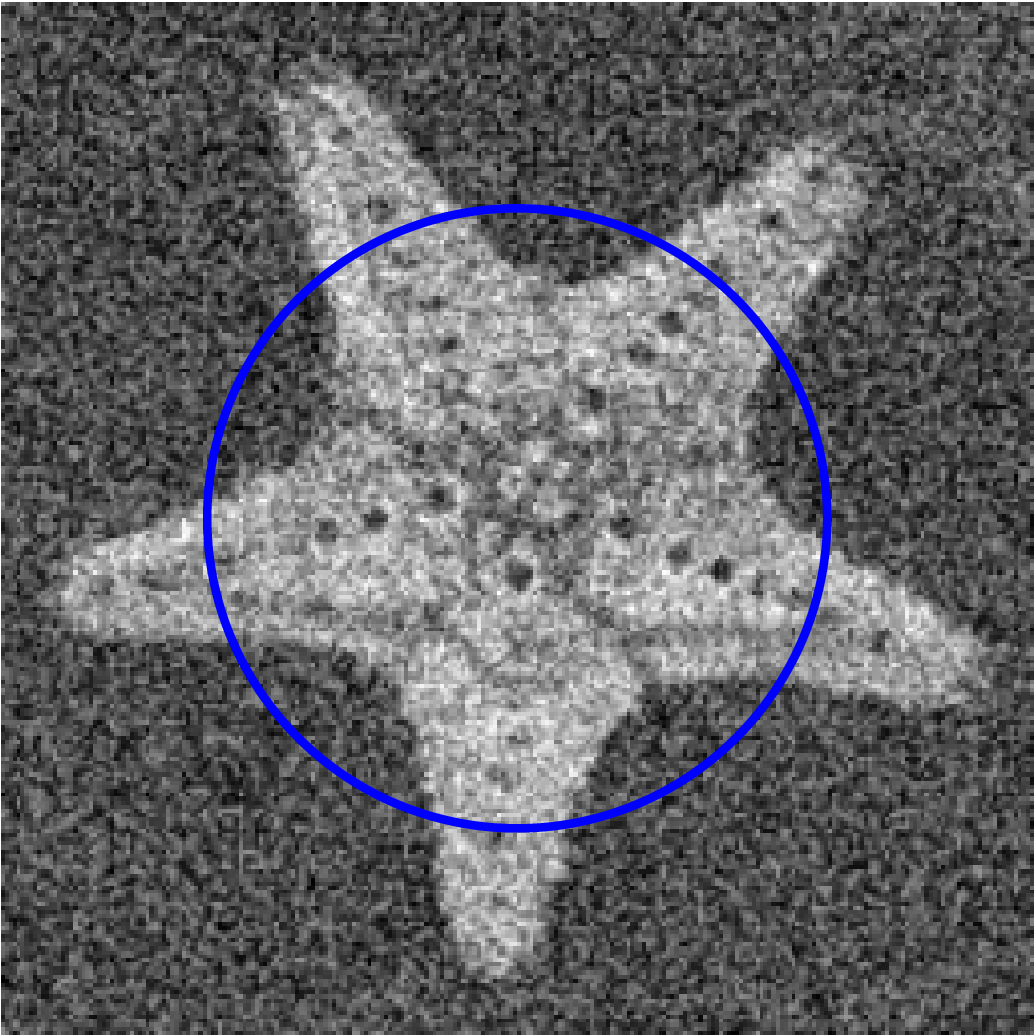}
\includegraphics[width=1.15in,height=1.15in]{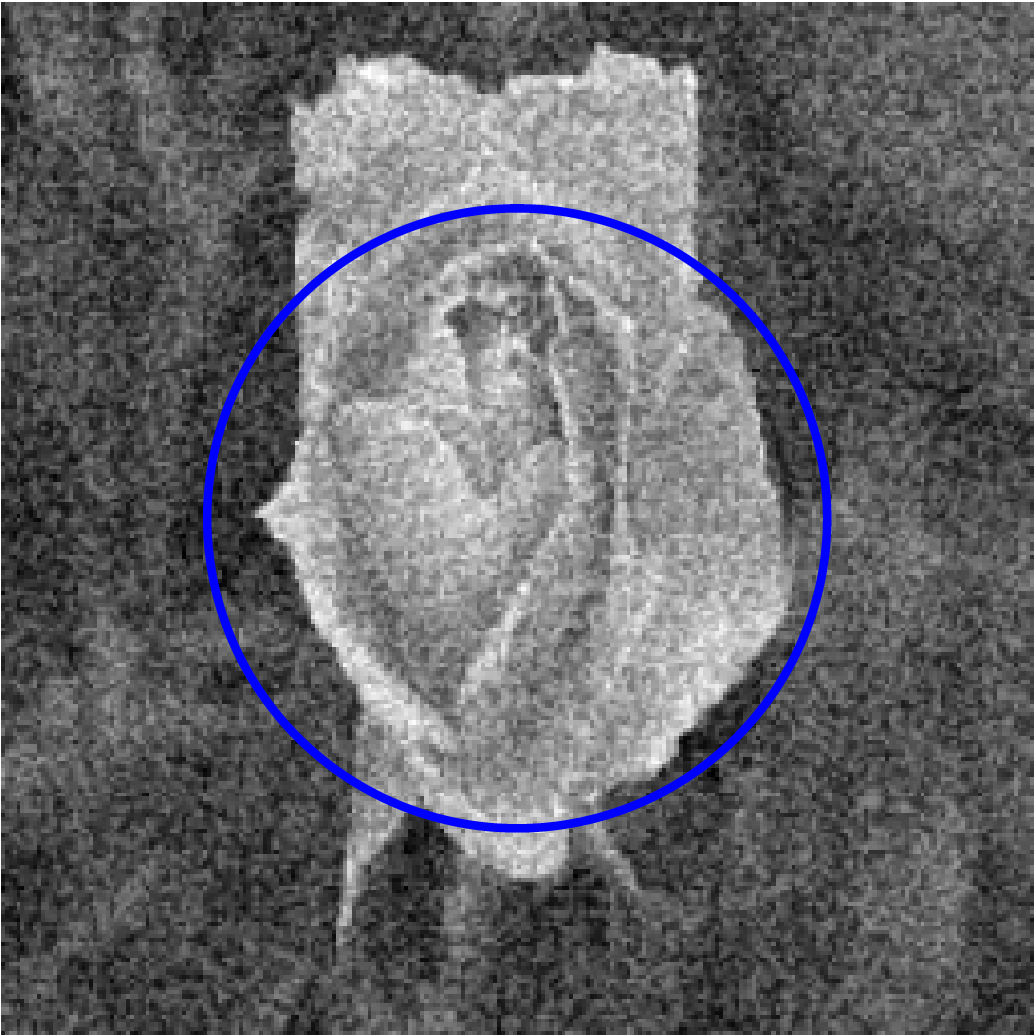}} 
\subfigure[Segmentation results by CV. \textcolor{black}{The running times are 3.0 s, 3.0 s, 4.6 s, and 3.6 s.}]{
\includegraphics[width=1.15in,height=1.15in]{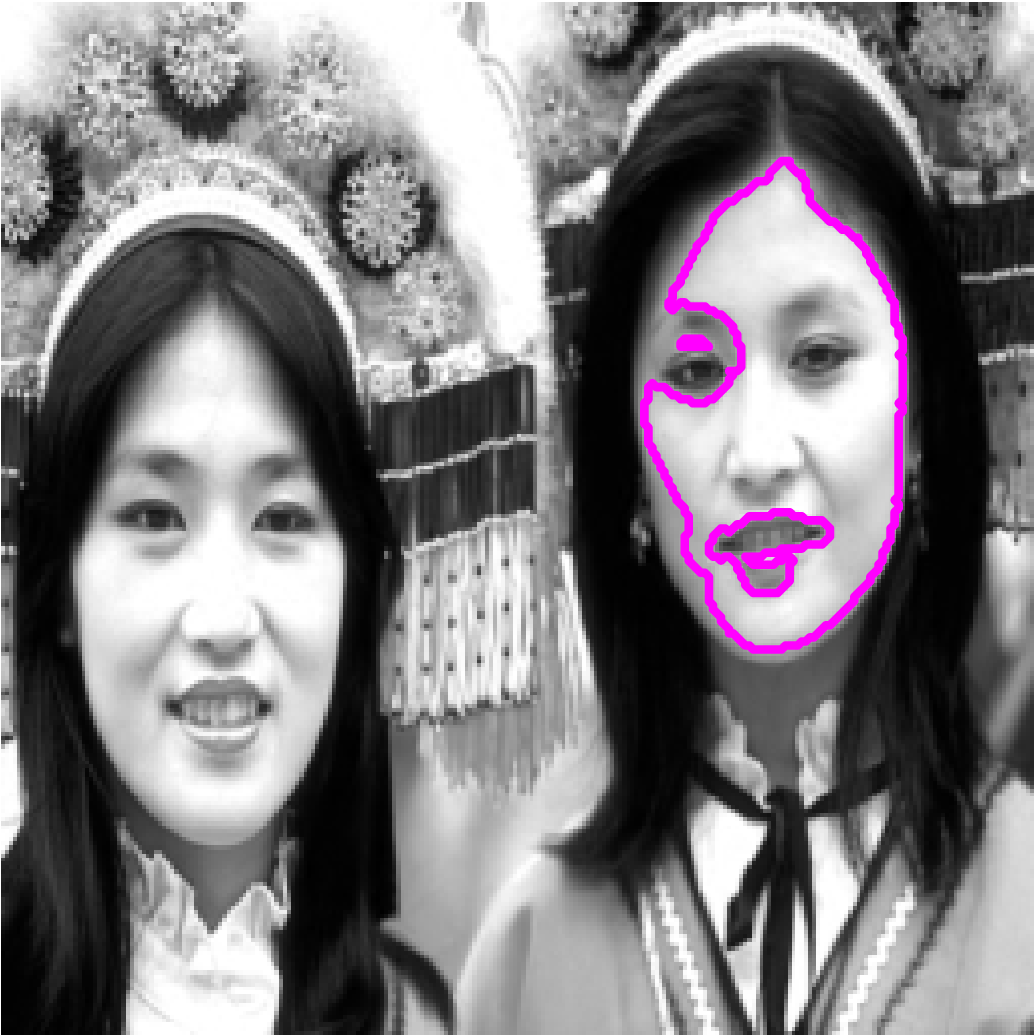}
\includegraphics[width=1.15in,height=1.15in]{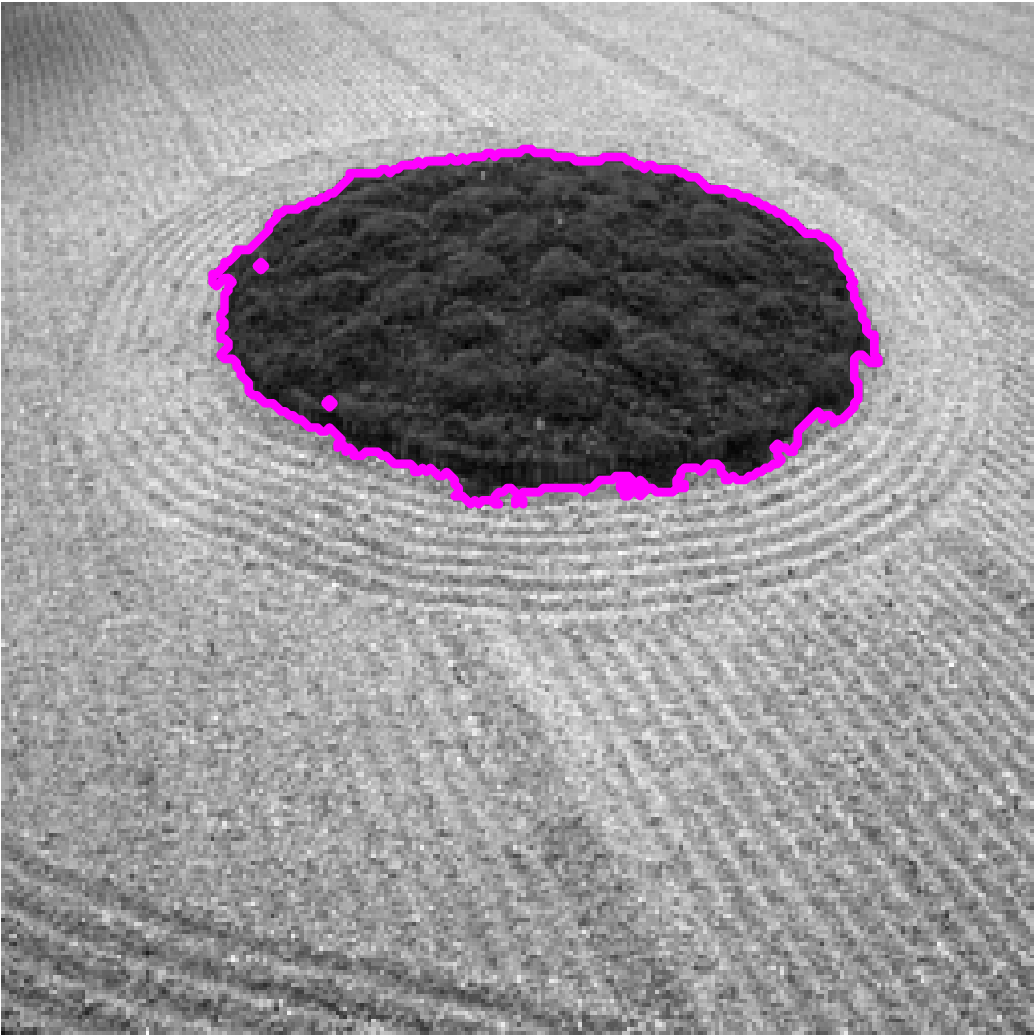}
\includegraphics[width=1.15in,height=1.15in]{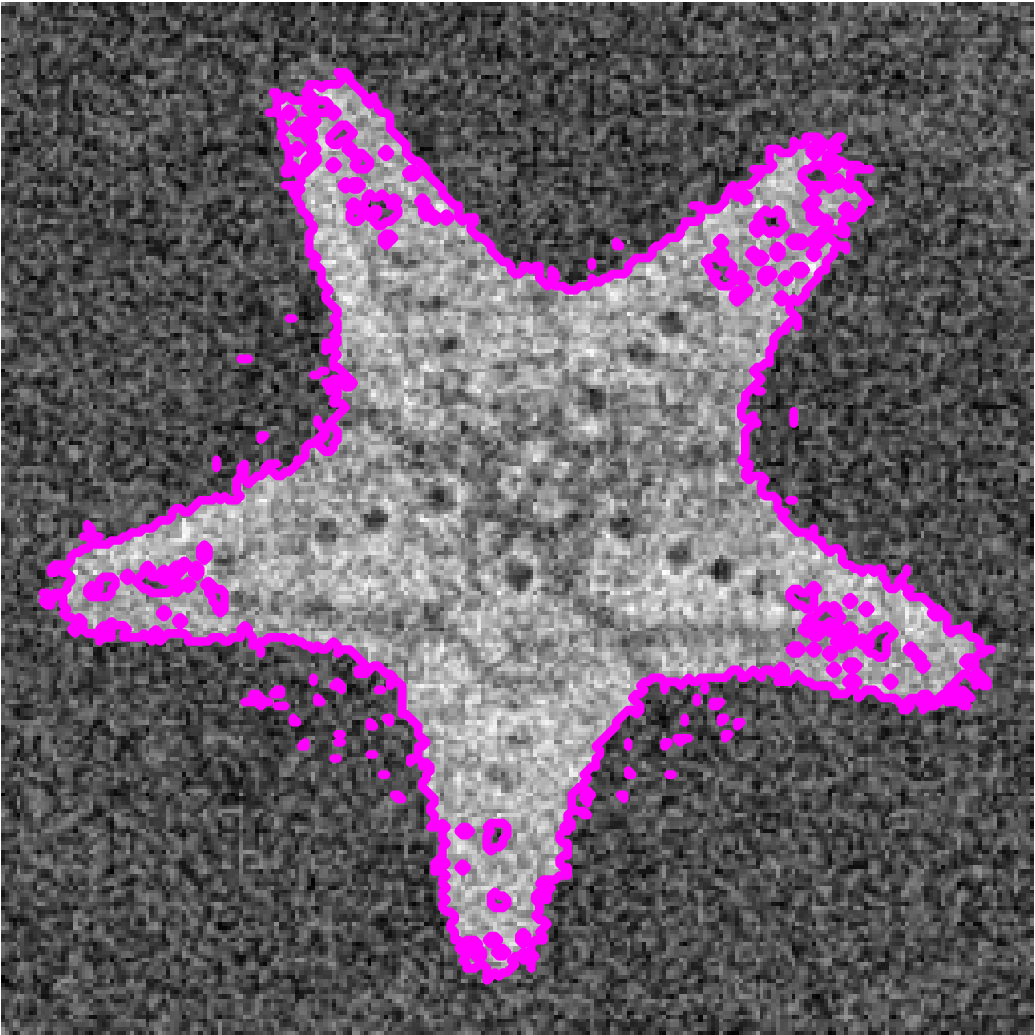}
\includegraphics[width=1.15in,height=1.15in]{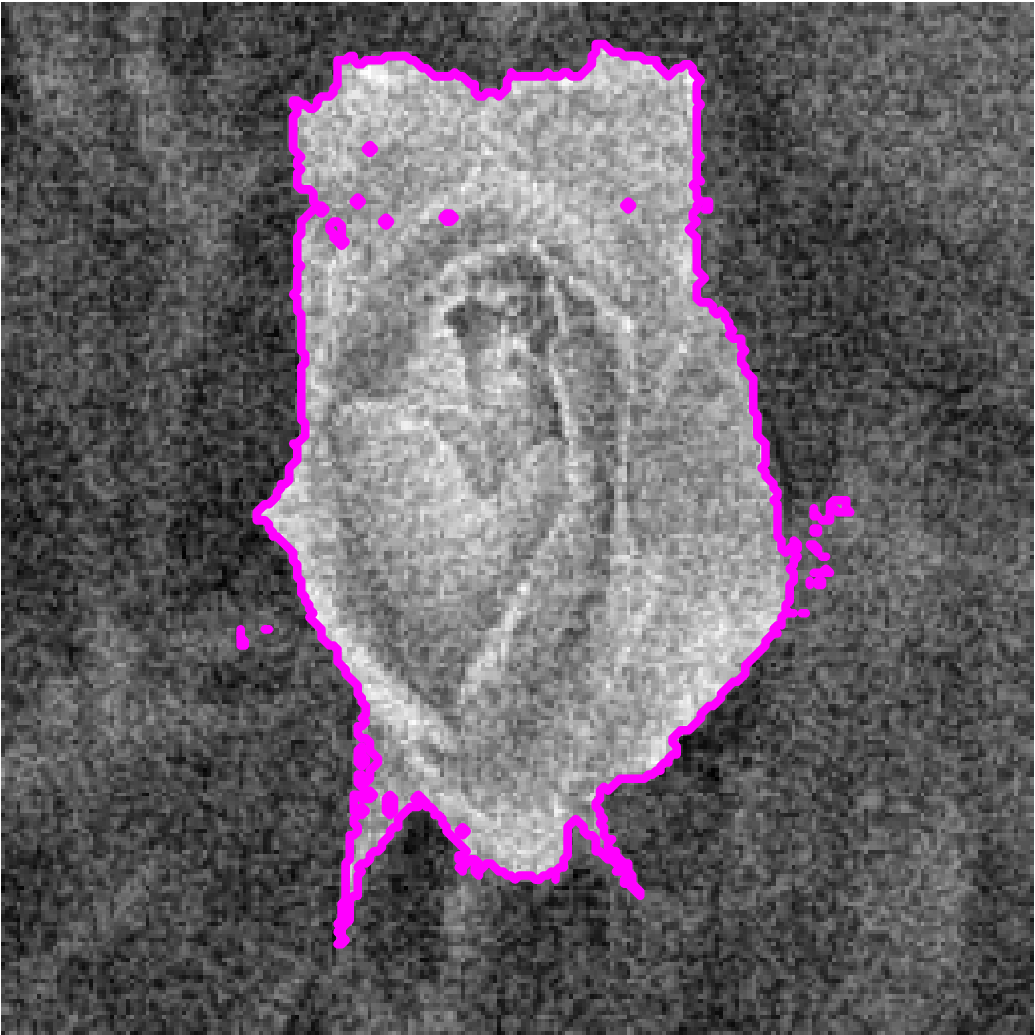}} 
\subfigure[\textcolor{black}{Segmentation results by convexity-preserving model. The running times are 19.9 s, 1.7 s, 9.2 s, and 9.4 s.}]{
\includegraphics[width=1.15in,height=1.15in]{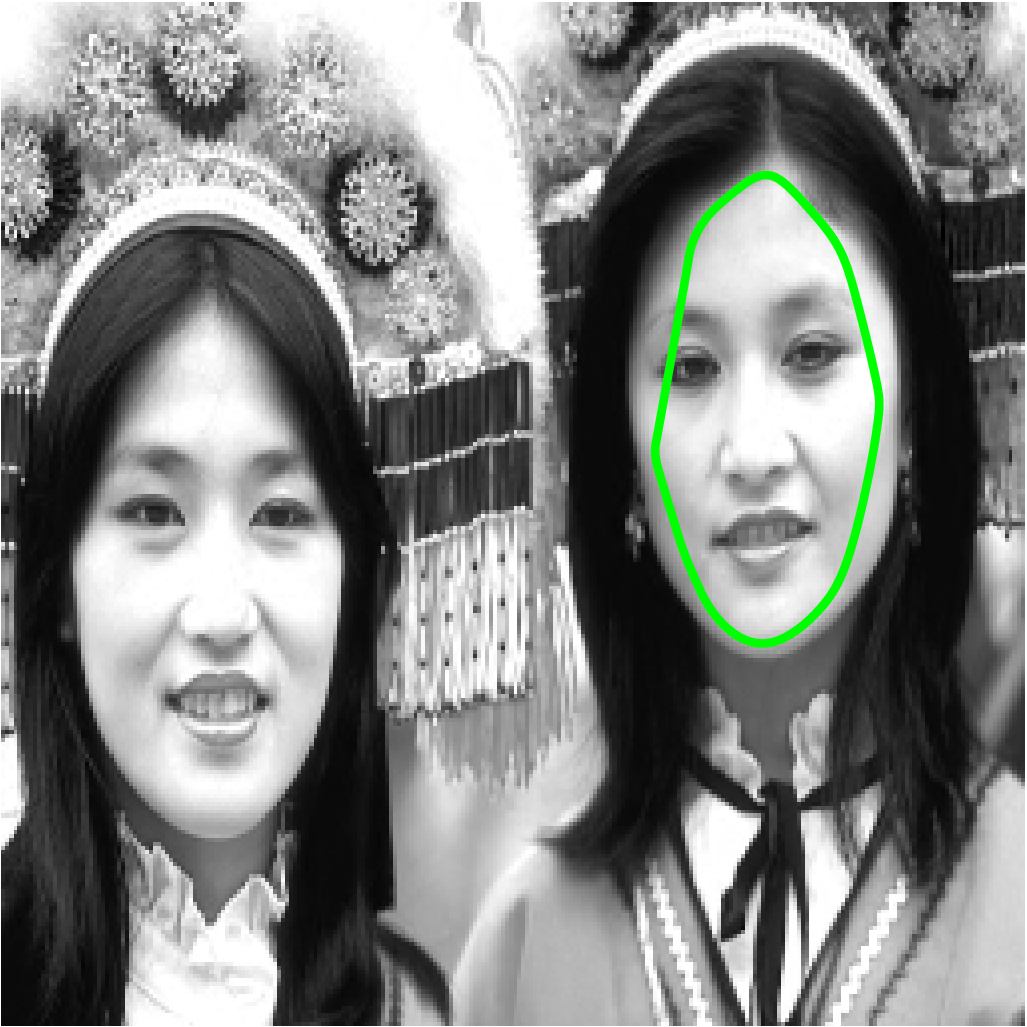}
\includegraphics[width=1.15in,height=1.15in]{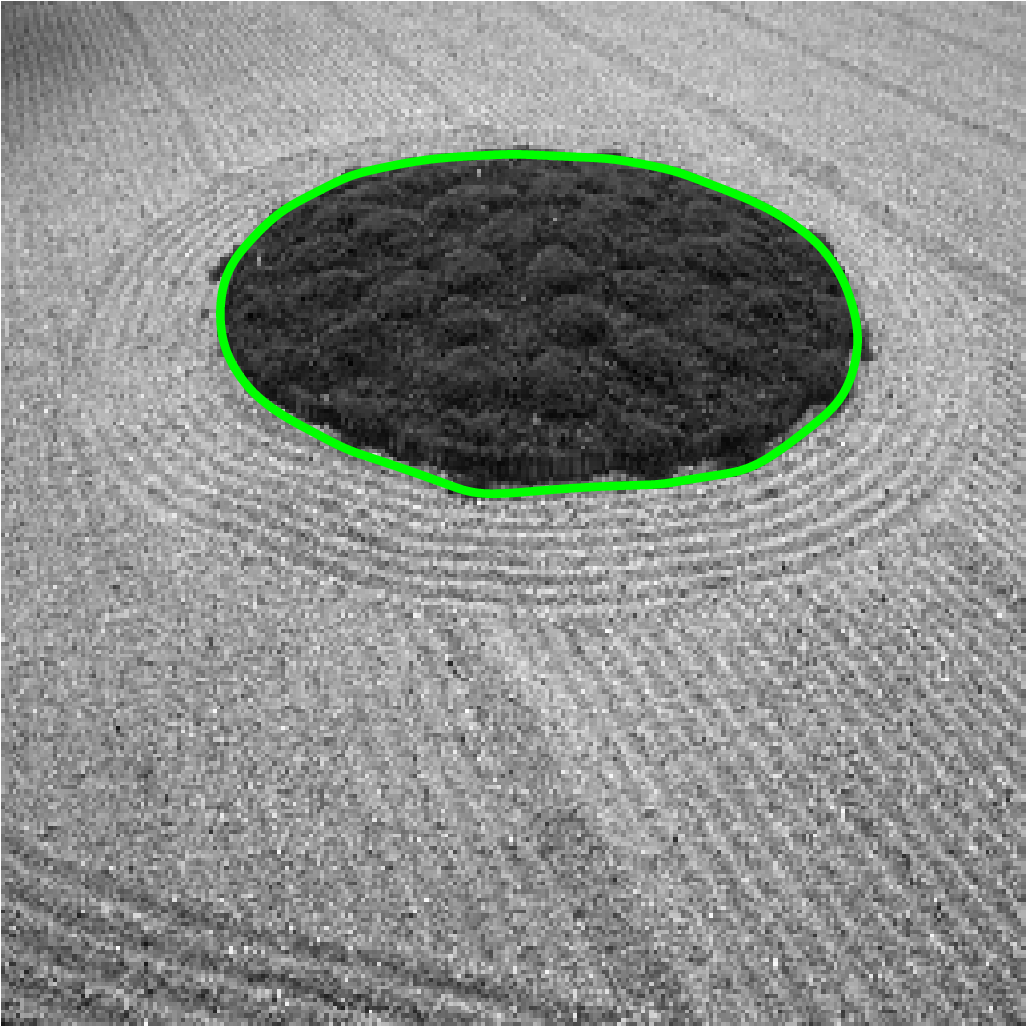}
\includegraphics[width=1.15in,height=1.15in]{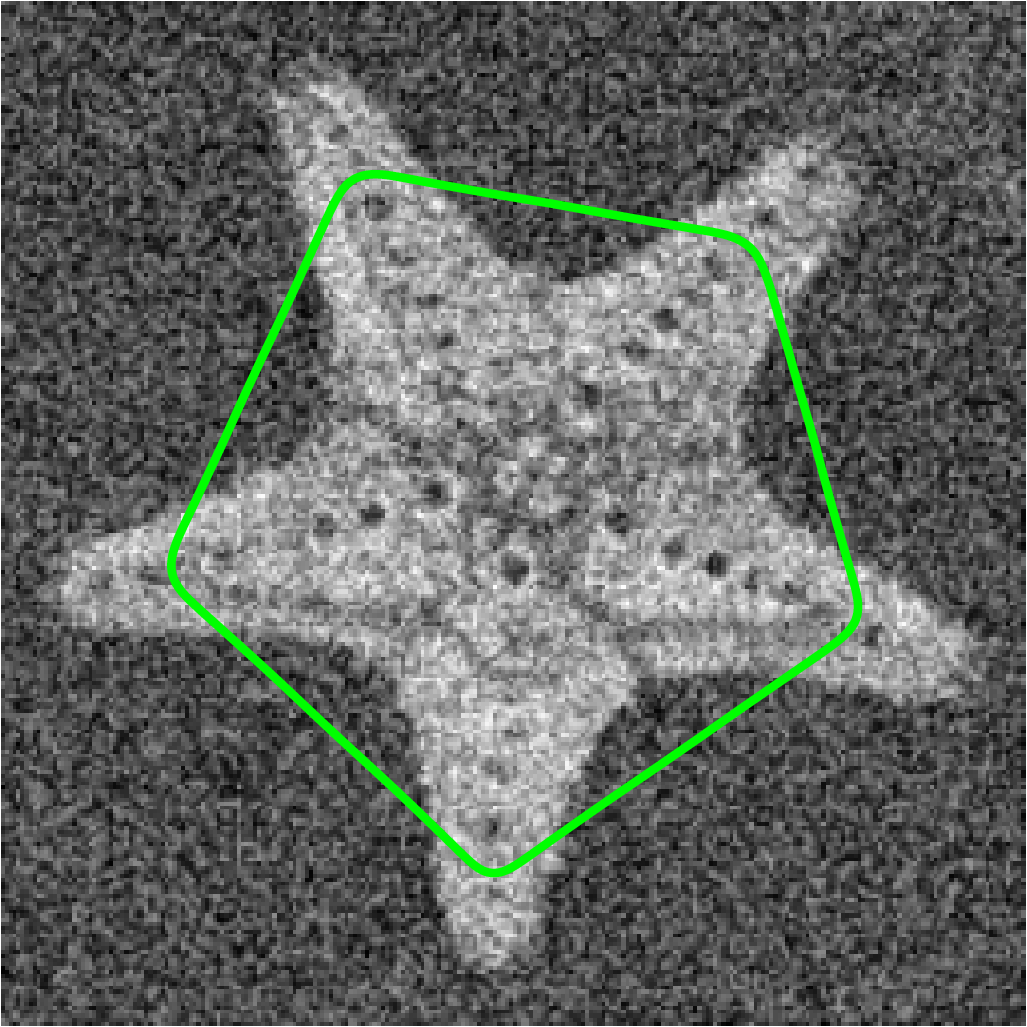}
\includegraphics[width=1.15in,height=1.15in]{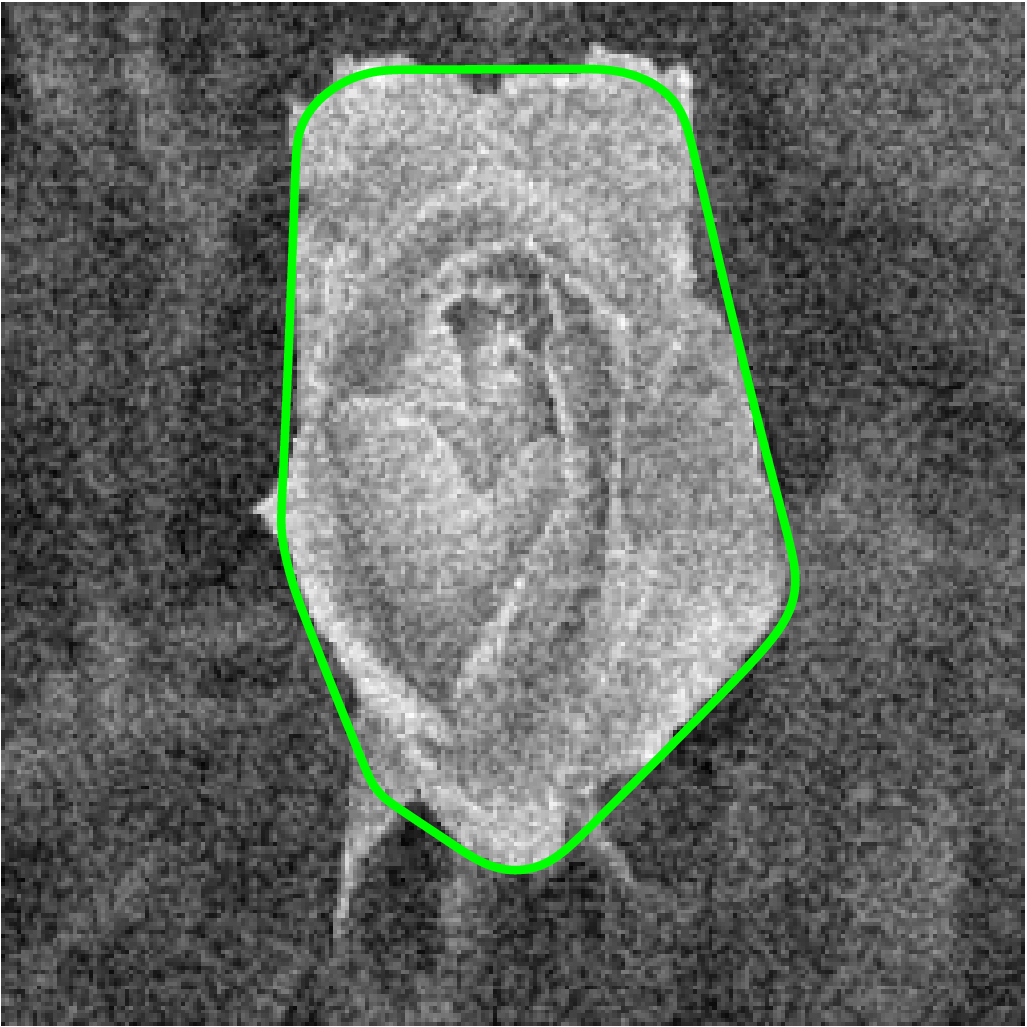}} 
\subfigure[Segmentation results by the proposed model \eqref{proposed_model1}. \textcolor{black}{The running times are 1.2 s, 0.6 s, 4.3 s, and 4.4 s.}]{
\includegraphics[width=1.15in,height=1.15in]{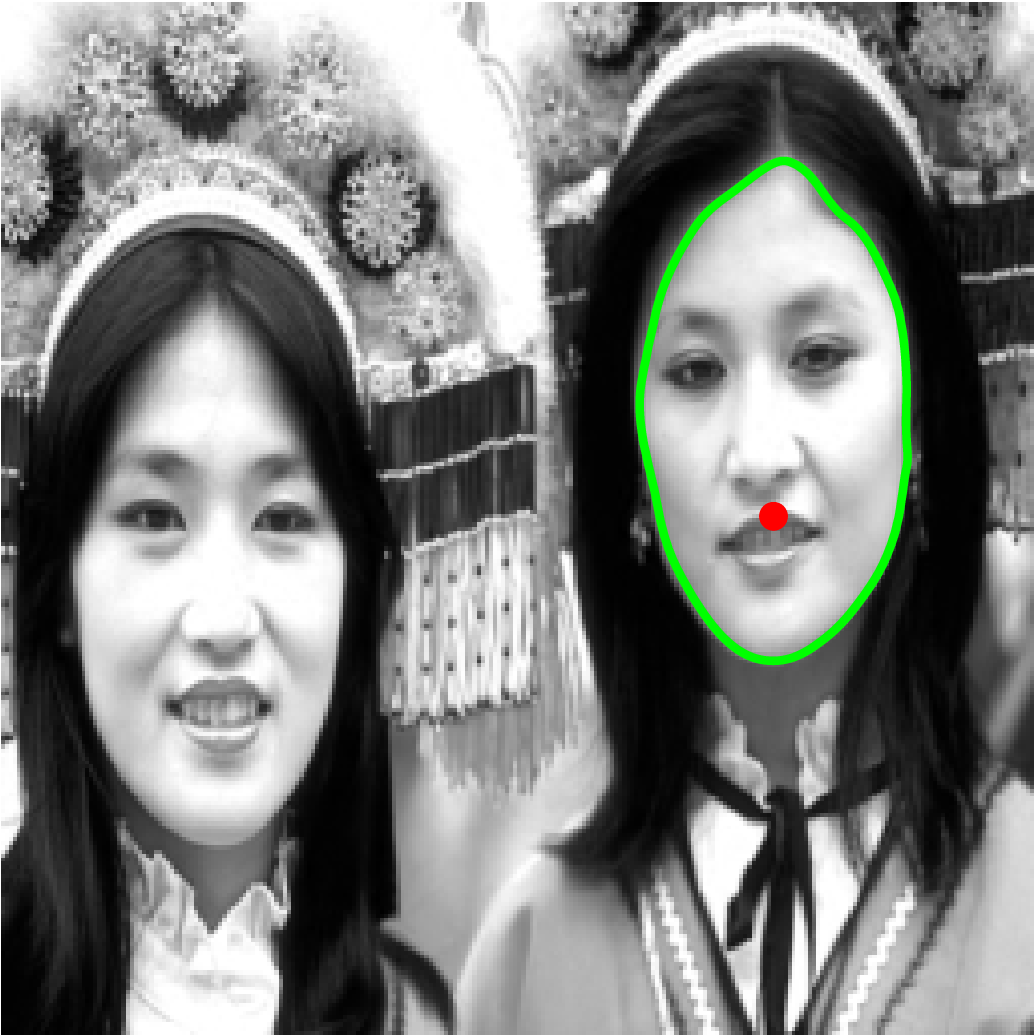}
\includegraphics[width=1.15in,height=1.15in]{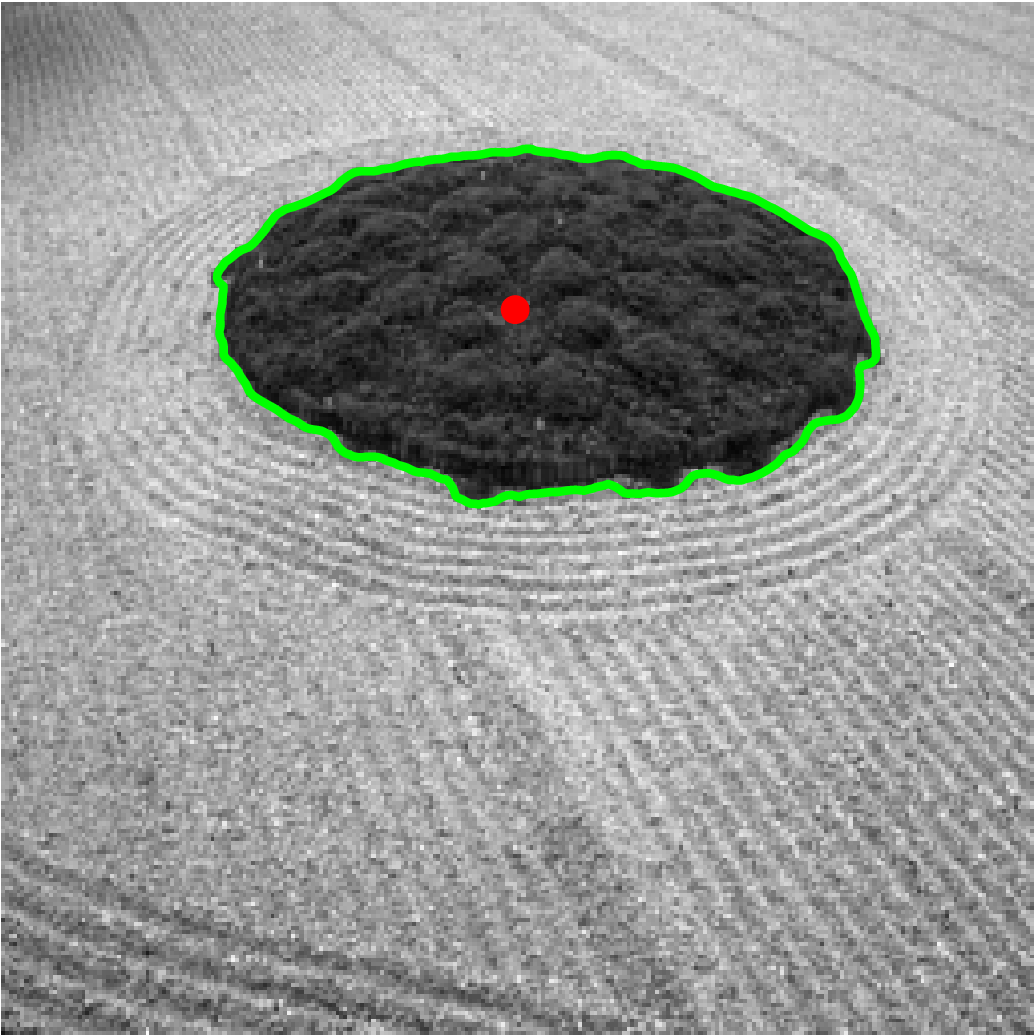}
\includegraphics[width=1.15in,height=1.15in]{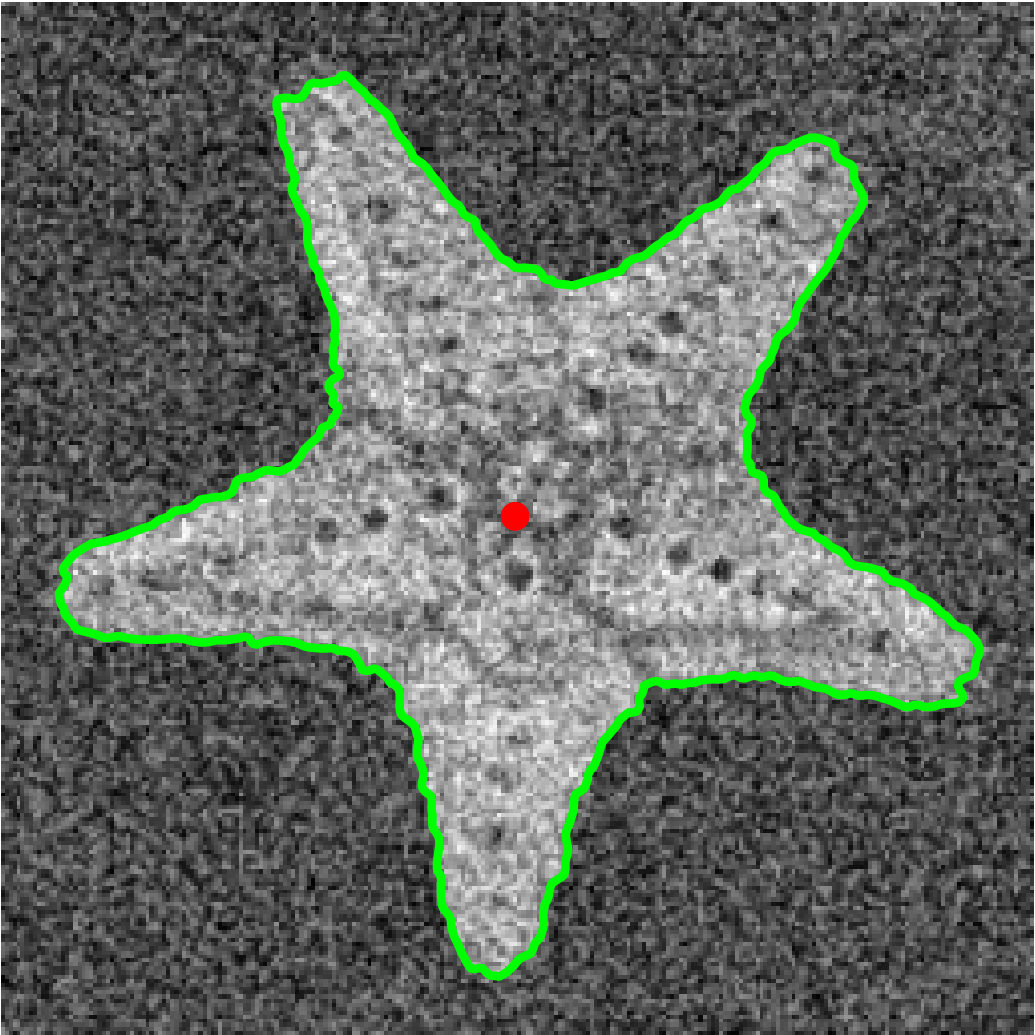}
\includegraphics[width=1.15in,height=1.15in]{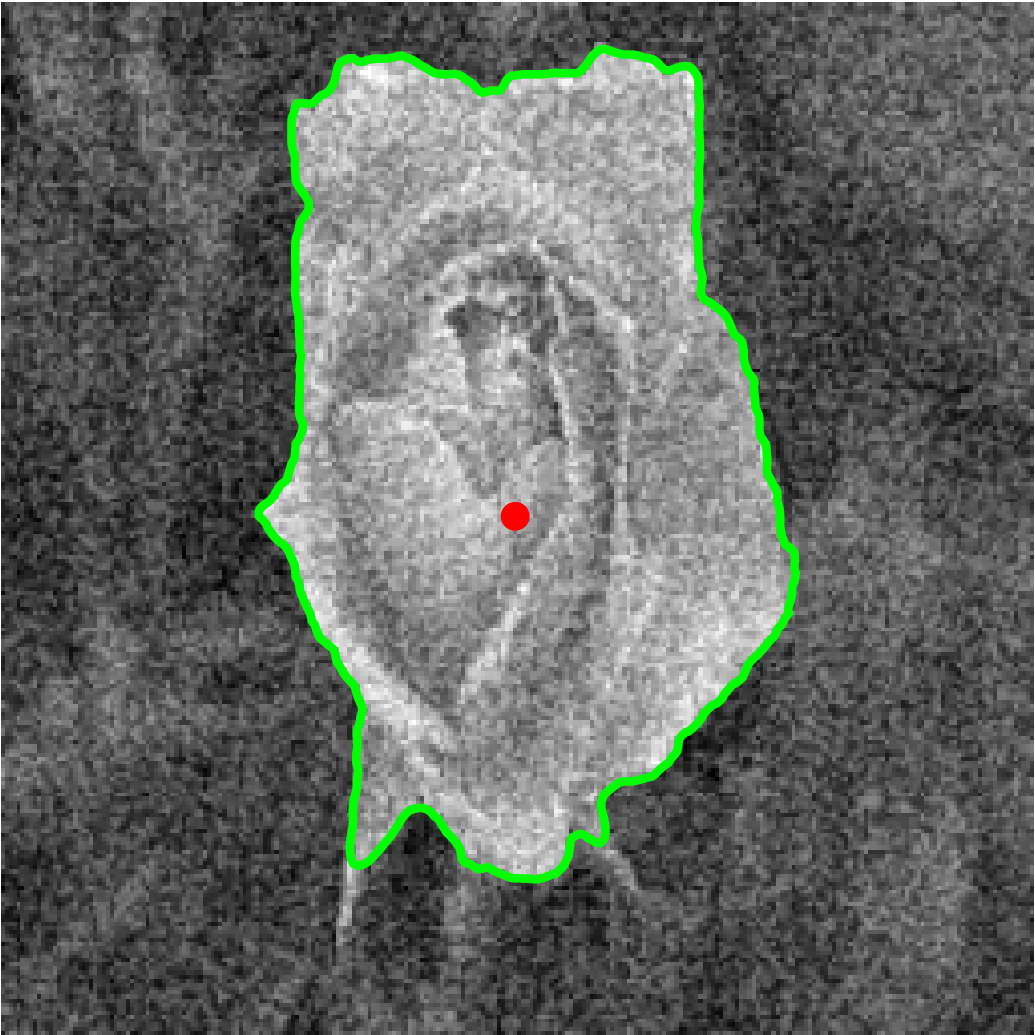}} 
\caption{Test on the proposed model \eqref{proposed_model1}. 
The first row shows the target images with different initial contours. The second row gives the segmentation results by the CV model. \textcolor{black}{The third row shows the segmentation results by the convexity-preserving model \cite{zhang2021topology2}.} The \textcolor{black}{fourth} row displays the segmentation results by the proposed model \eqref{proposed_model1}. Here, the red point is the given center. \textcolor{black}{The running time
is measured in seconds.}}\label{Exp2_fig1}
\end{figure}

\begin{figure}[htbp!]
\centering
\subfigure[Segmentation results produced by CV with different parameters. \textcolor{black}{The running times are 3.7 s, 3.3 s, 3.2 s, and 3.0 s.}]{
\includegraphics[width=1.15in,height=1.15in]{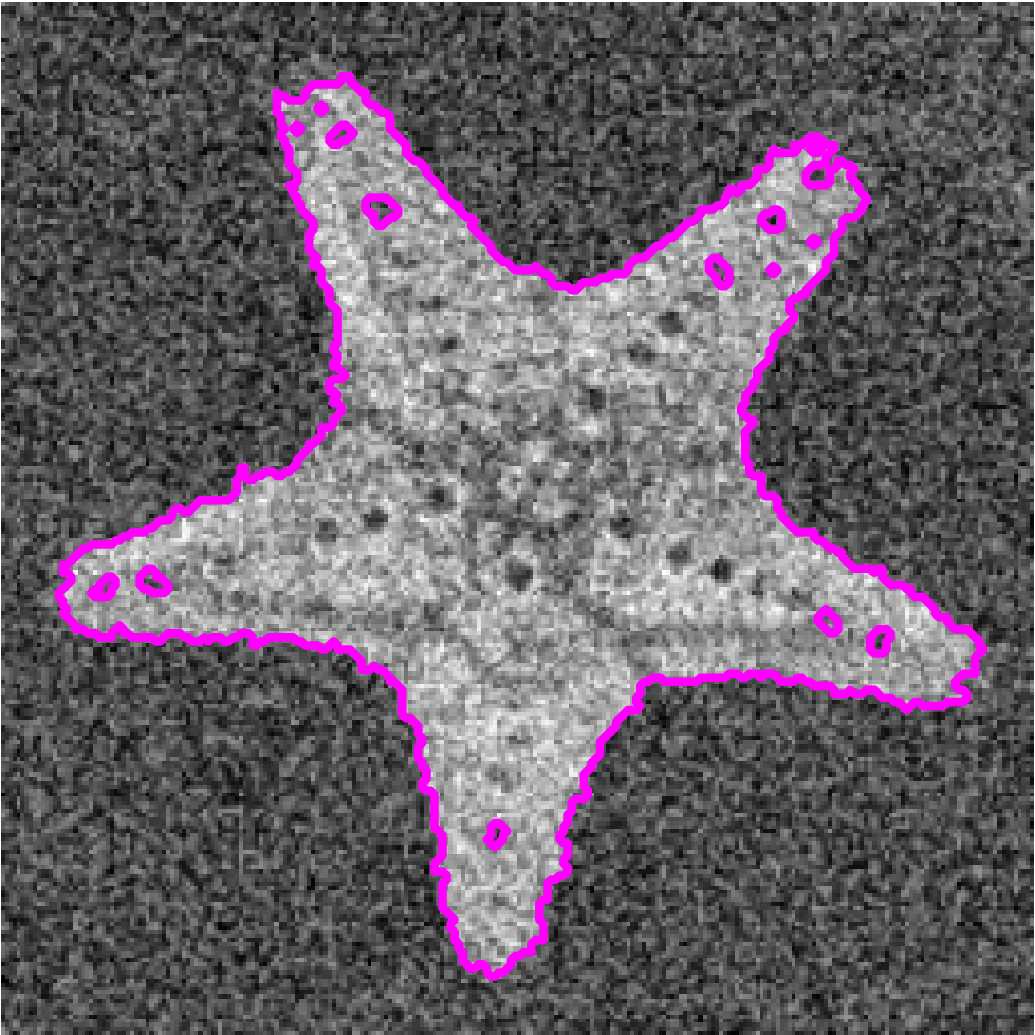}
\includegraphics[width=1.15in,height=1.15in]{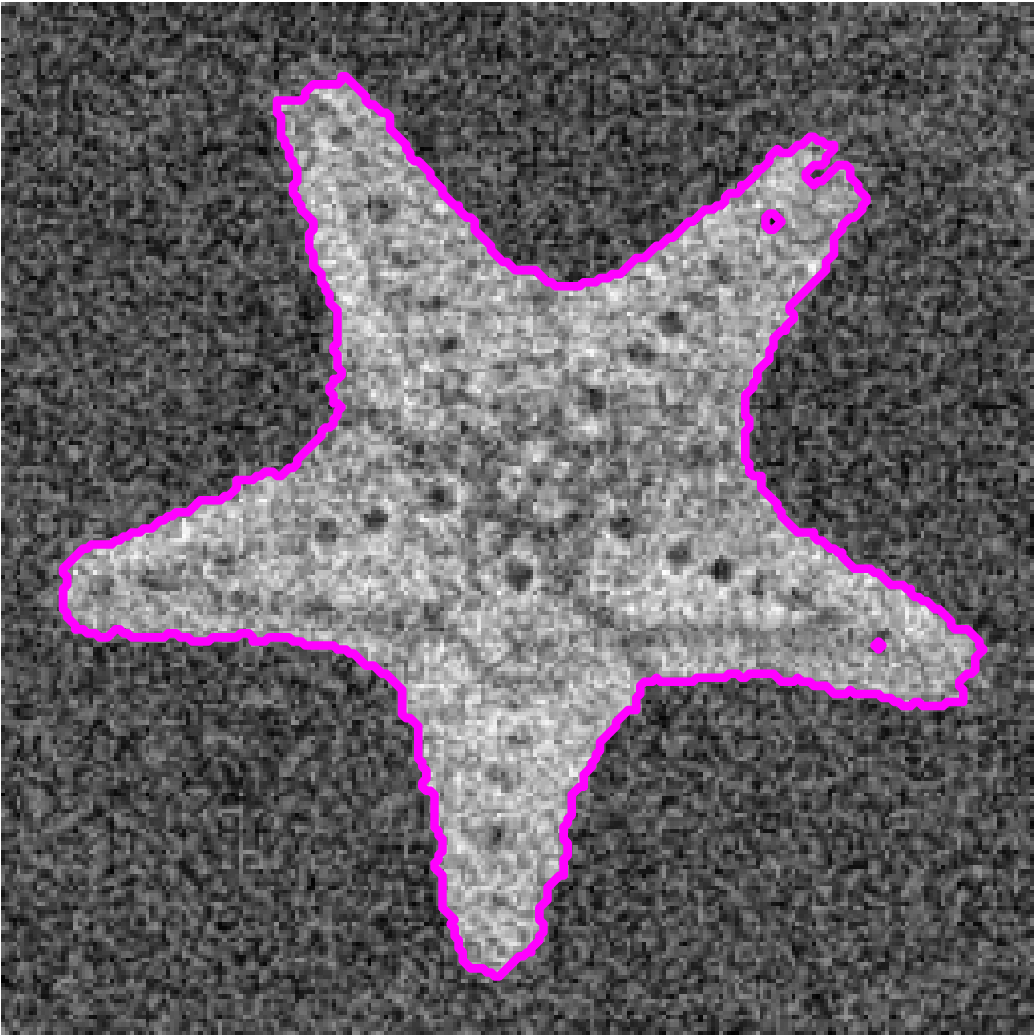}
\includegraphics[width=1.15in,height=1.15in]{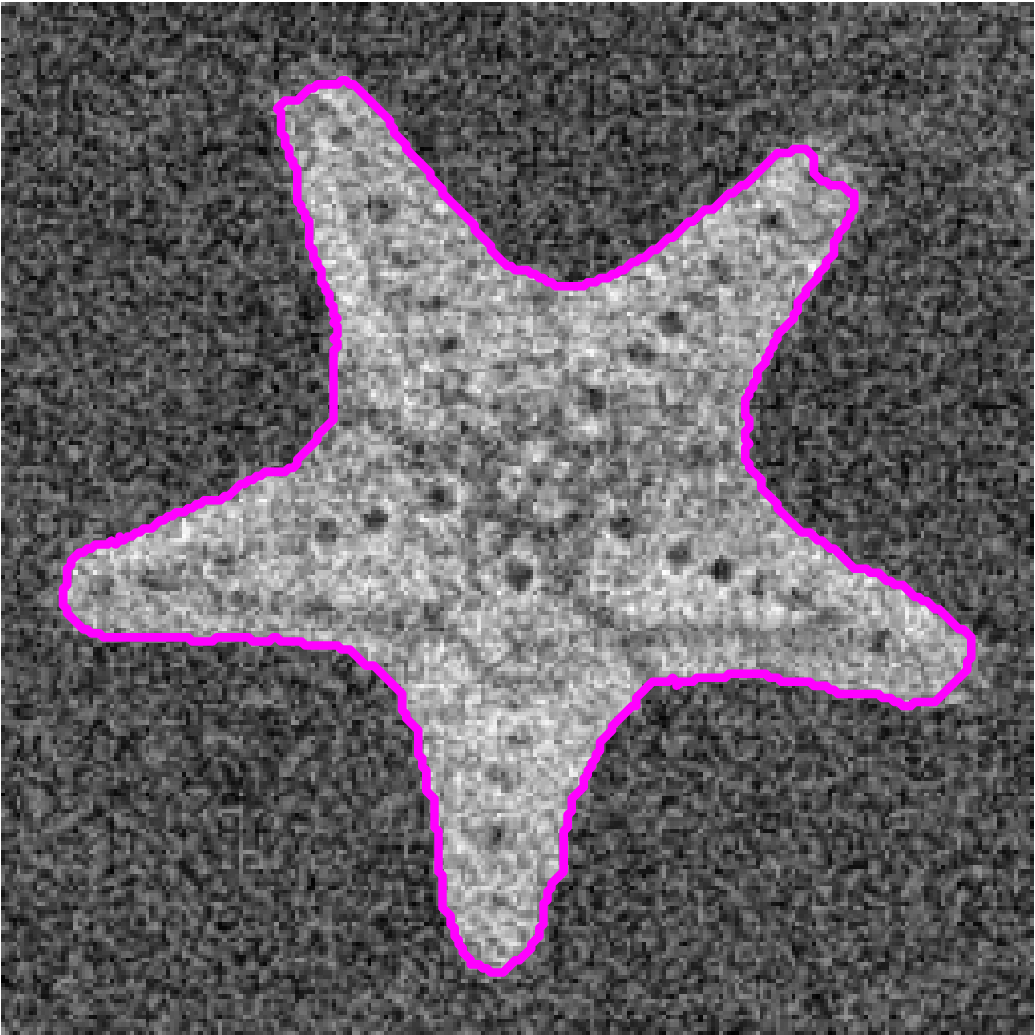}
\includegraphics[width=1.15in,height=1.15in]{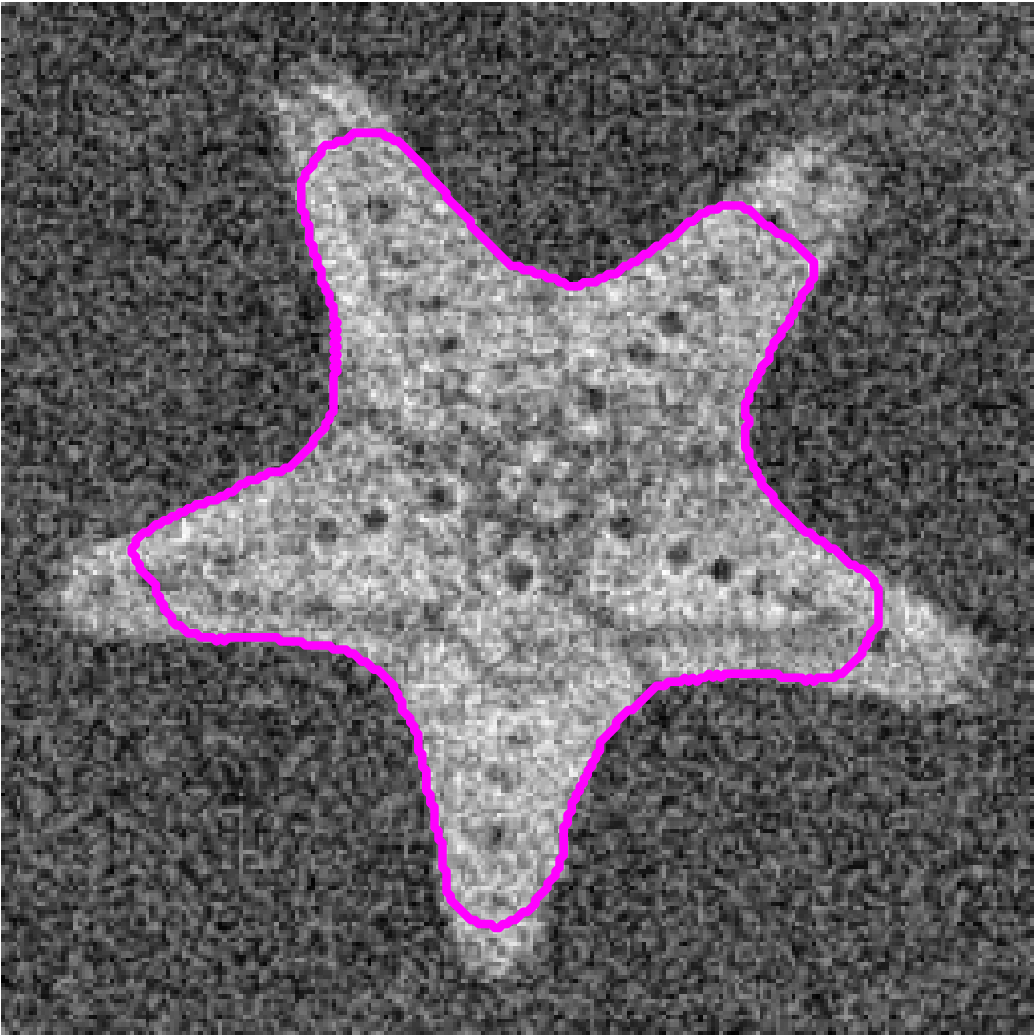}} 
\caption{Segmentation results of the noisy starfish produced by 
the CV model with different parameters. Here, we set the smooth factor in the \textit{activecontour} as 0.5, 1, 2, and 5, respectively. \textcolor{black}{The running time
is measured in seconds.}}\label{Exp2_fig2}
\end{figure}

\begin{figure}[htbp!]
\centering
\subfigure[\textcolor{black}{Initial contour (left) and segmentation (right). Running time 3.8 seconds.}]{
\includegraphics[width=1.15in,height=1.15in]{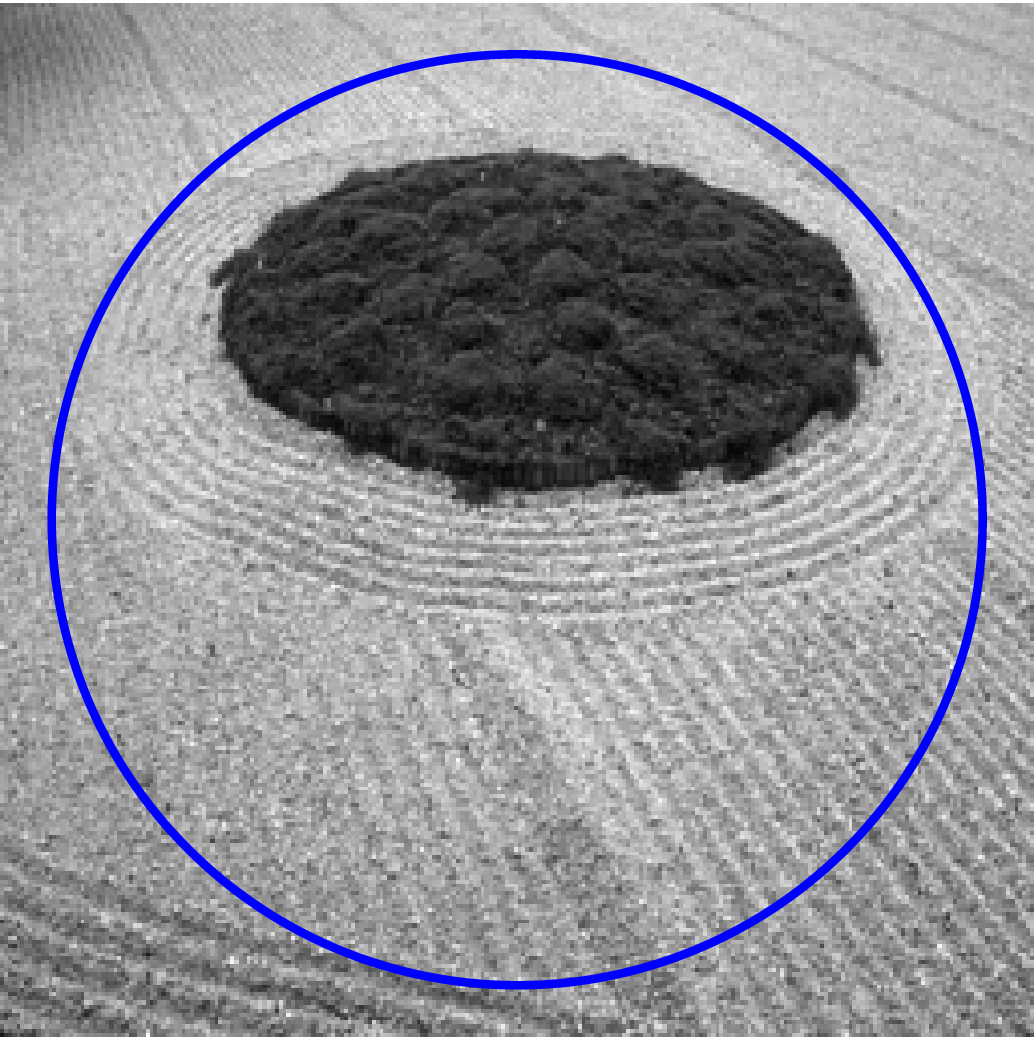}
\includegraphics[width=1.15in,height=1.15in]{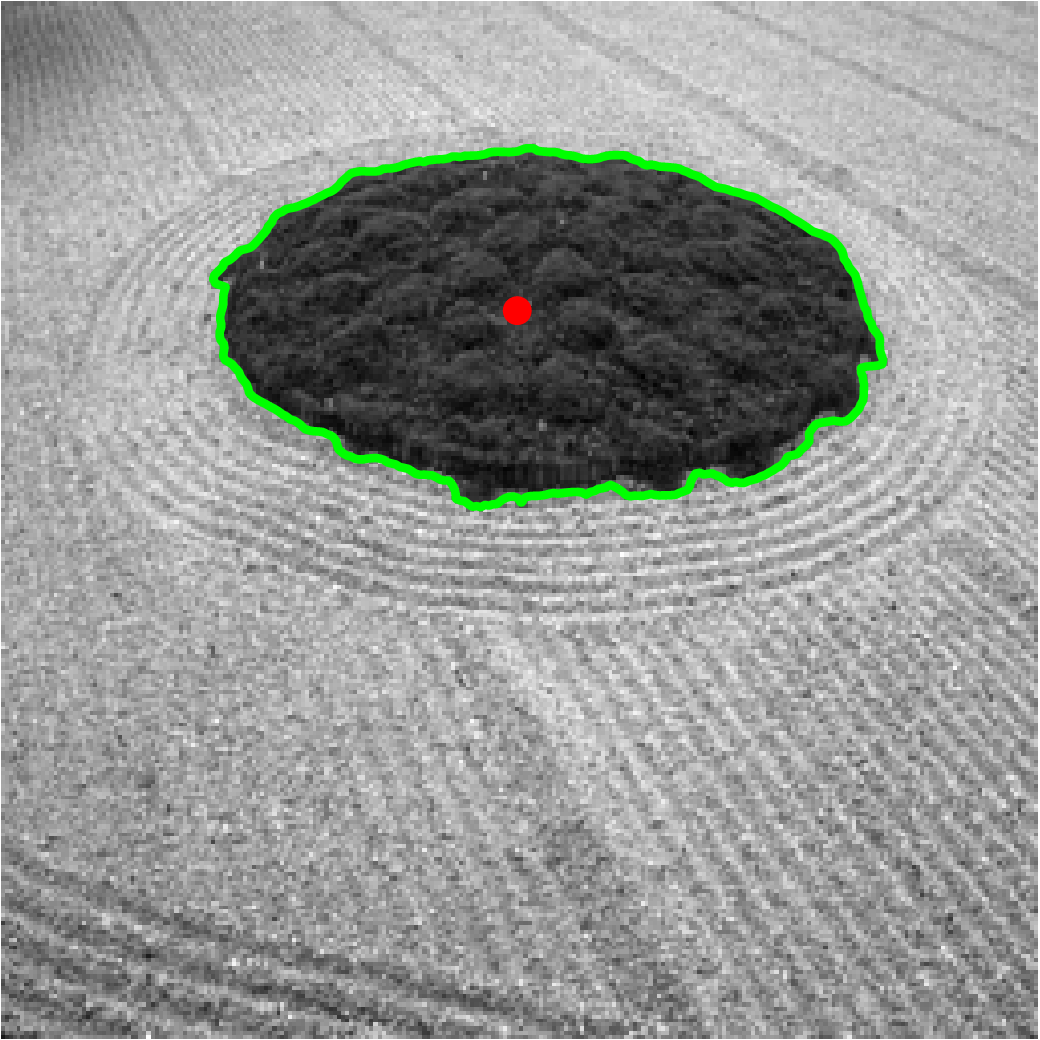}}
\subfigure[\textcolor{black}{Initial contour (right) and segmentation (left). Running time 5.0 seconds.}]{
\includegraphics[width=1.15in,height=1.15in]{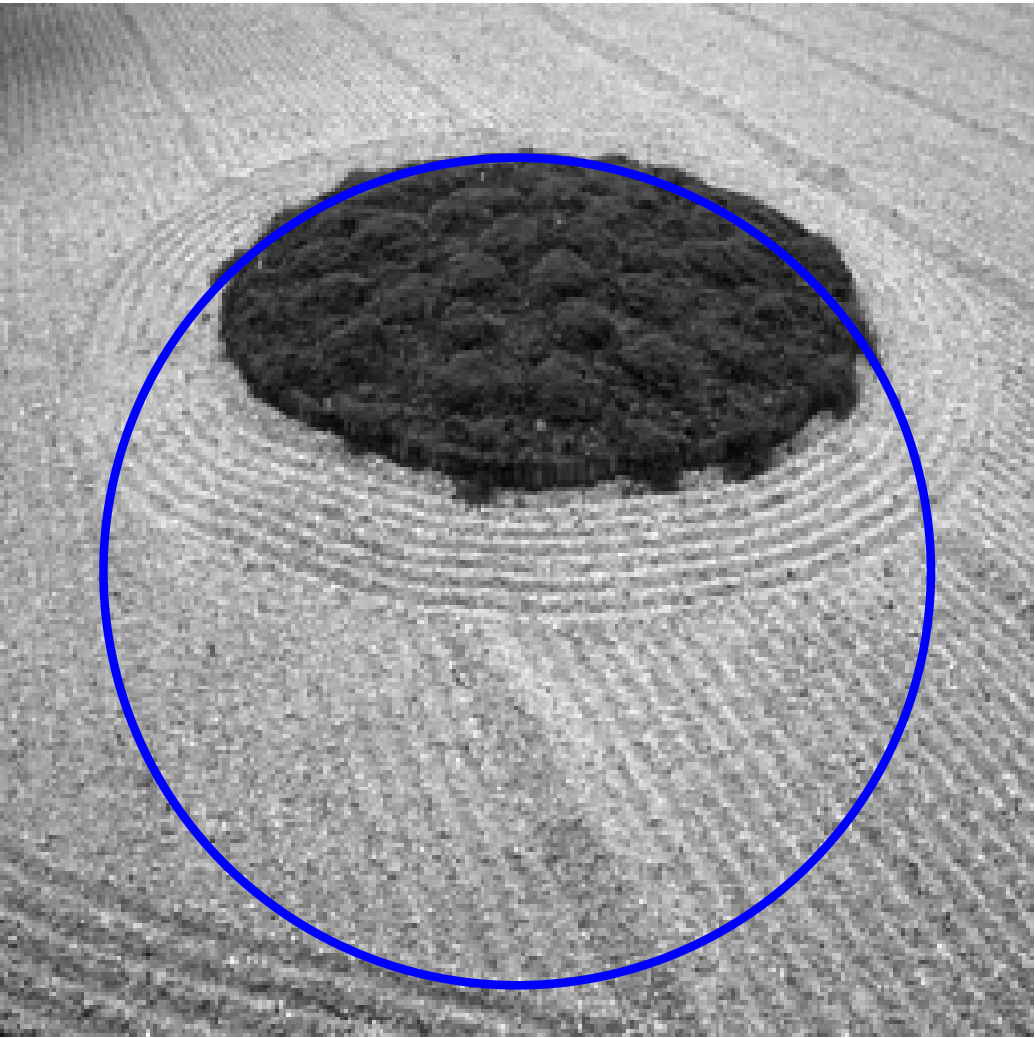}
\includegraphics[width=1.15in,height=1.15in]{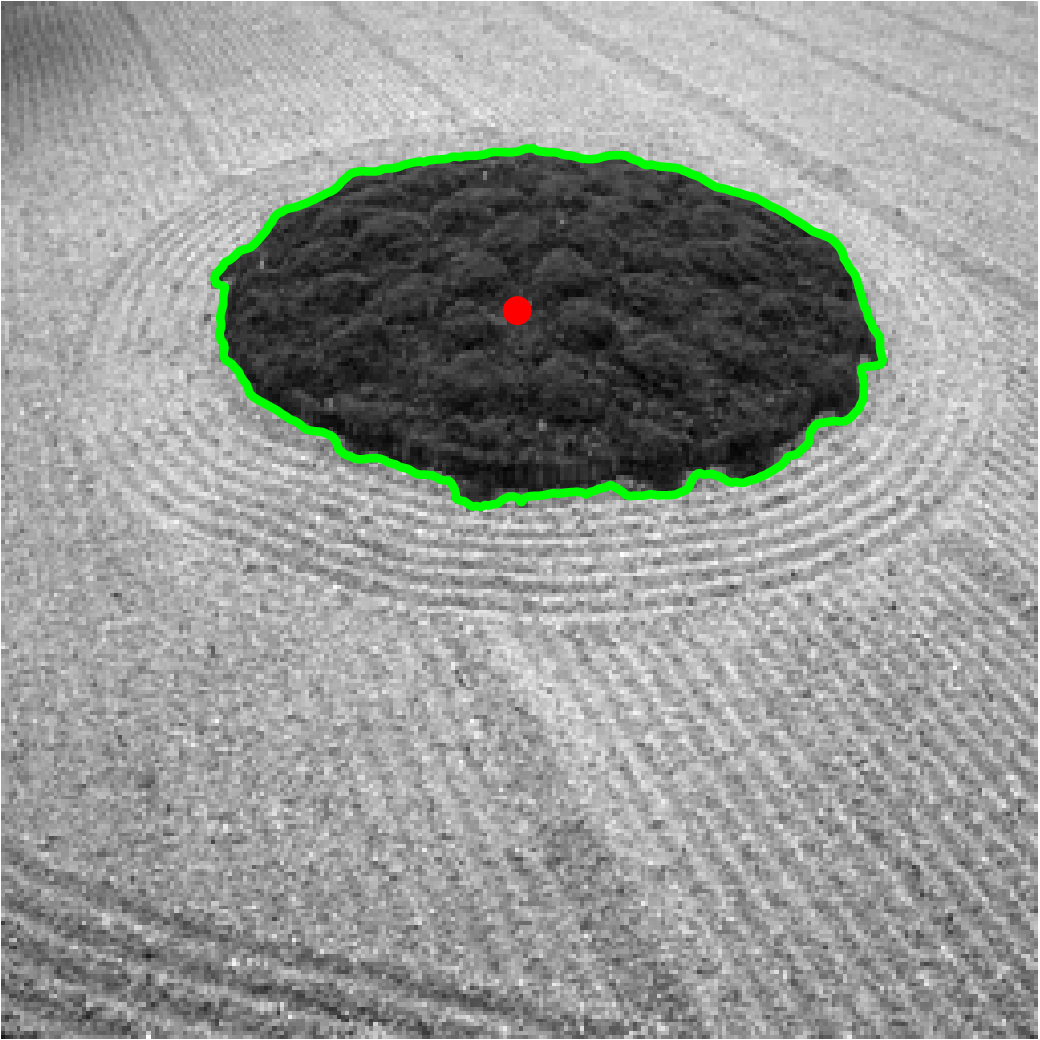}} 
\caption{\textcolor{black}{Segmentation results by the proposed model \eqref{proposed_model1} with different initial contours. Here, the red point is the given center.}}\label{Exp2_fig1_1}
\end{figure}

We first test the proposed model \eqref{proposed_model1} using four images, two clean images and two noisy images, as depicted in Fig. \eqref{Exp2_fig1}(a). For the parameter $\alpha$ in the proposed model \eqref{proposed_model1}, we set $0.01$ for the first image, and $0.001$ for the remaining three. \textcolor{black}{The segmentation results of the CV model, the convexity-preserving model \cite{zhang2021topology2}, and the proposed model \eqref{proposed_model1} are shown in Fig. \ref{Exp2_fig1}(b-d), respectively. By its theoretical property, the convexity-preserving model only captures the object outlines. Moreover, due to its use of the same average fitting term as the CV model, it remains sensitive to intensity inhomogeneity and fails to segment objects accurately, as seen in the first example of Fig. \ref{Exp2_fig1}(b).} Notably, the proposed model \eqref{proposed_model1} yields the star-shape segmentation with respect to the given centers, preserving the entire target object, unlike the CV model, which often produces outliers, especially for noisy images. This disparity arises from two factors: imposing the star-shape constraint and deforming the level set function via transformation. To ensure that the radial line from the center point intersects the star-shape domain's boundary only once, the segmentation by the proposed model \eqref{proposed_model1} must constitute a complete part. Moreover, segmentations achieved by regulating transformation smoothness typically exhibit greater robustness compared to those based on level set function length regularization. To elucidate, we apply the CV model with varying parameters to the fourth image, a noisy starfish. In this case, we set the smooth factor in \textit{activecontour} as 0.5, 1, 2, and 5, respectively, and present the corresponding segmentation results in Fig. \ref{Exp2_fig2}. It is evident that for the noisy image, the CV model struggles to produce a satisfactory segmentation result. \textcolor{black}{We also examine the influence of initial contour placement using the second image presented in Fig. \ref{Exp2_fig1_1}. Our experiments demonstrate that despite employing different initial contours, the segmentation results remain consistent. This consistency can be attributed to the multilevel strategy's ability to mitigate the impact of initial contour selection. However, it should be noted that when the initial contours are positioned farther from the optimal solution compared to that in Fig. \ref{Exp2_fig1}, the computational time increases accordingly.}

\begin{figure}[htbp!]
\centering
\subfigure[Target image and the initial contour]{
\includegraphics[width=1.13in,height=1.13in]{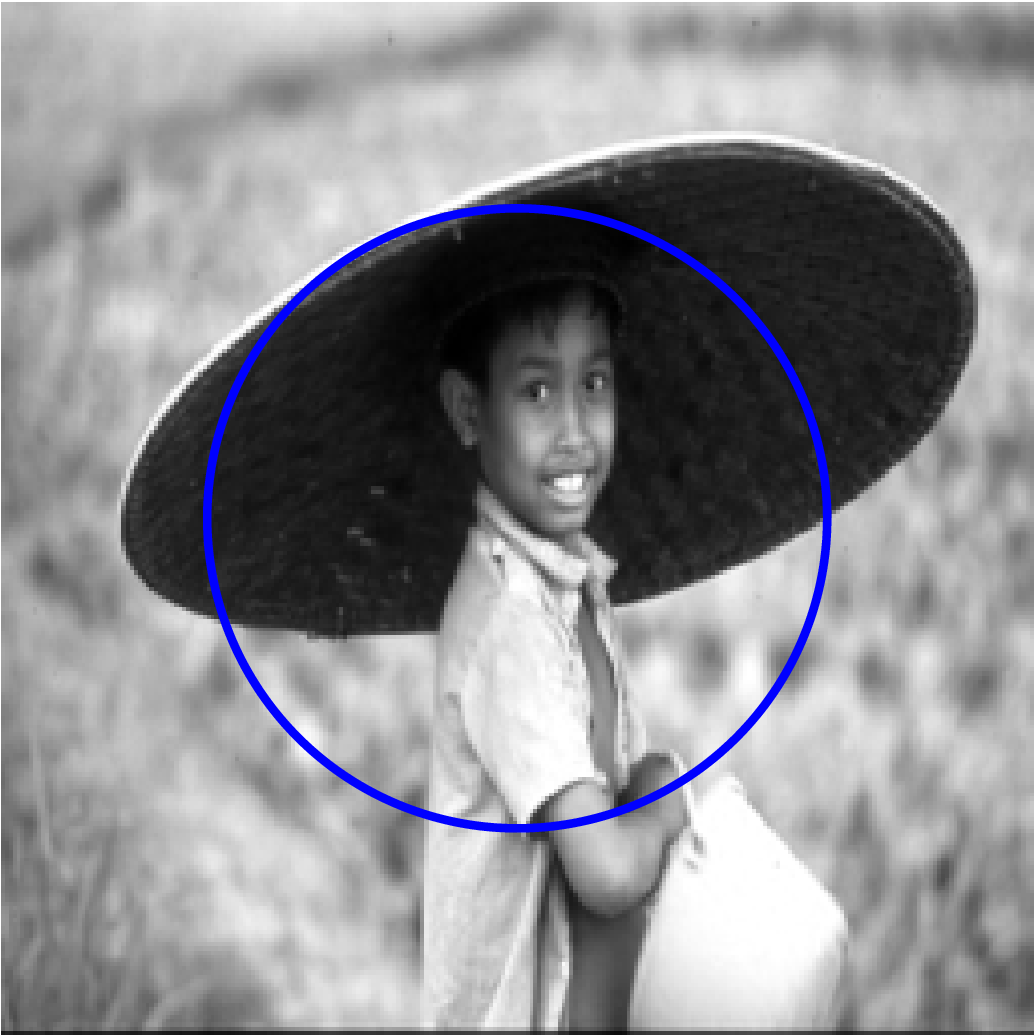}} 
\hspace{0.5cm}
\subfigure[Segmentation by CV. \textcolor{black}{Running time 4.0 seconds.}]{
\includegraphics[width=1.13in,height=1.13in]{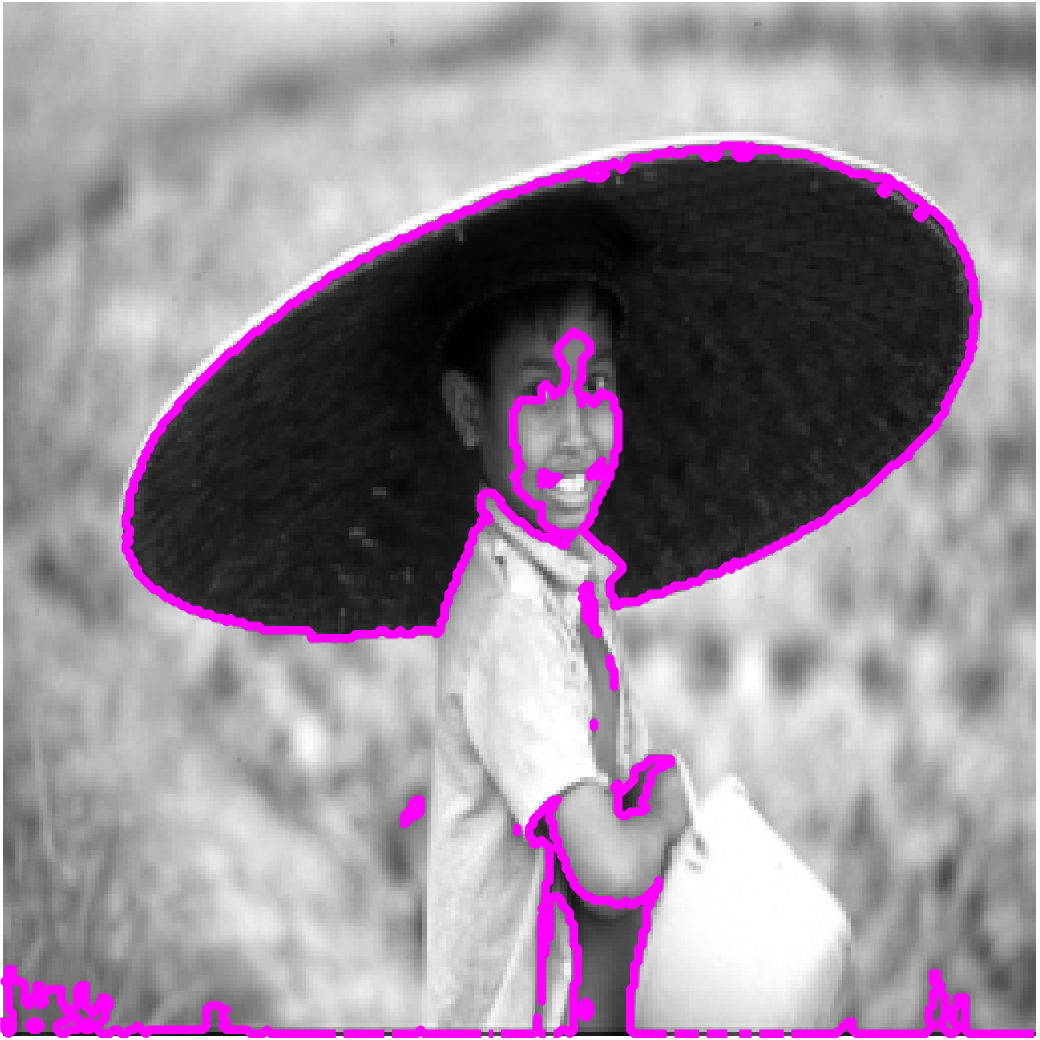}}
\hspace{0.5cm} 
\subfigure[\textcolor{black}{Segmentation by the convexity-preserving model. Running time 12.2 seconds.}]{
\includegraphics[width=1.13in,height=1.13in]{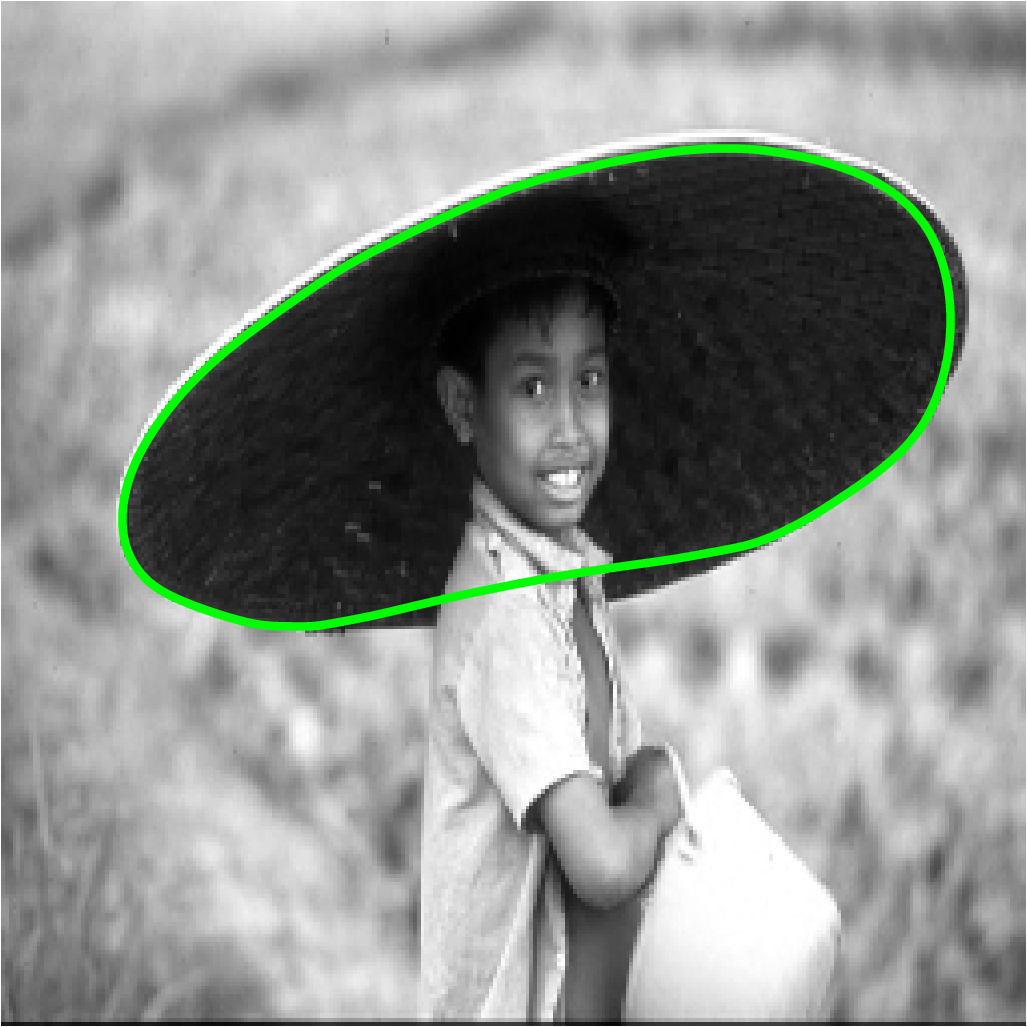}}
\\
\subfigure[Segmentation by \eqref{proposed_model1} with one center. \textcolor{black}{Running time 4.6 seconds.}]{
\includegraphics[width=1.13in,height=1.13in]{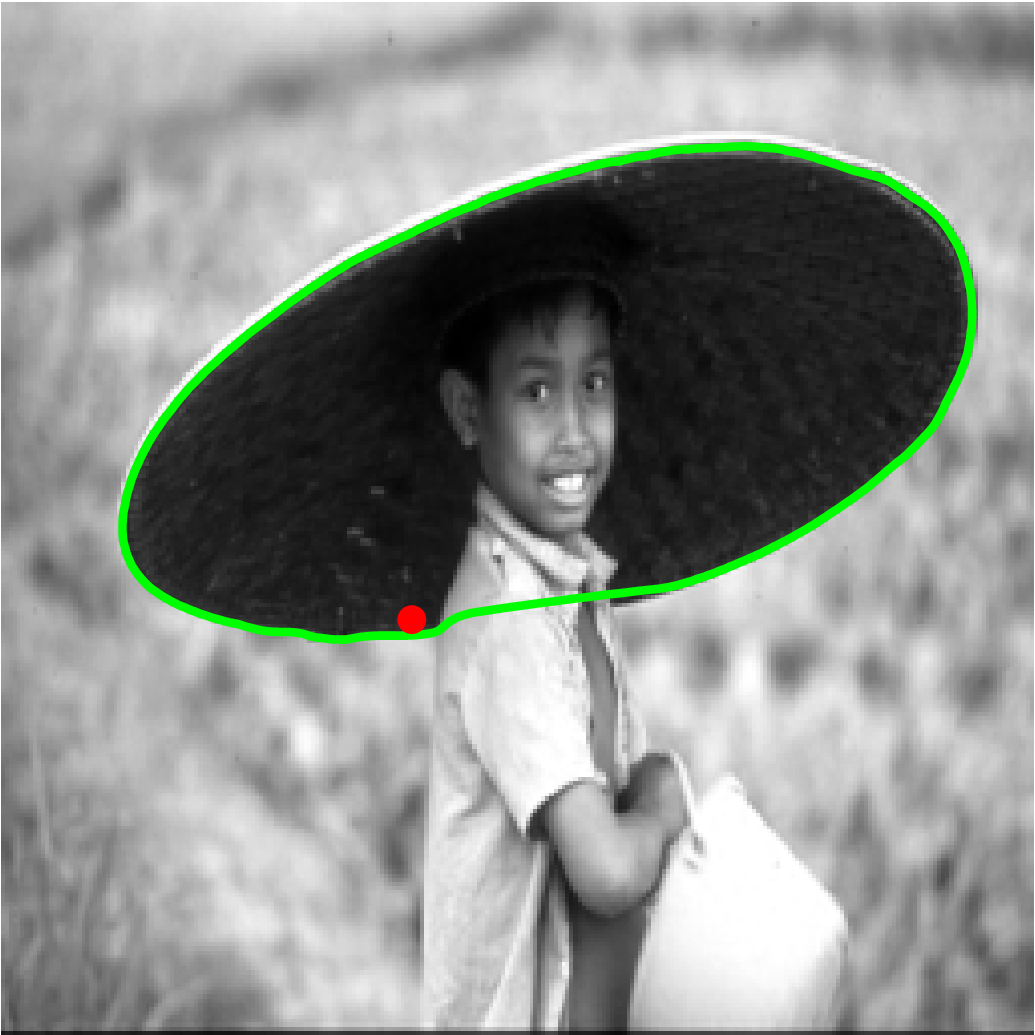}}
\hspace{0.5cm}
\subfigure[Segmentation by \eqref{proposed_model1} with one center. \textcolor{black}{Running time 3.7 seconds.}]{
\includegraphics[width=1.13in,height=1.13in]{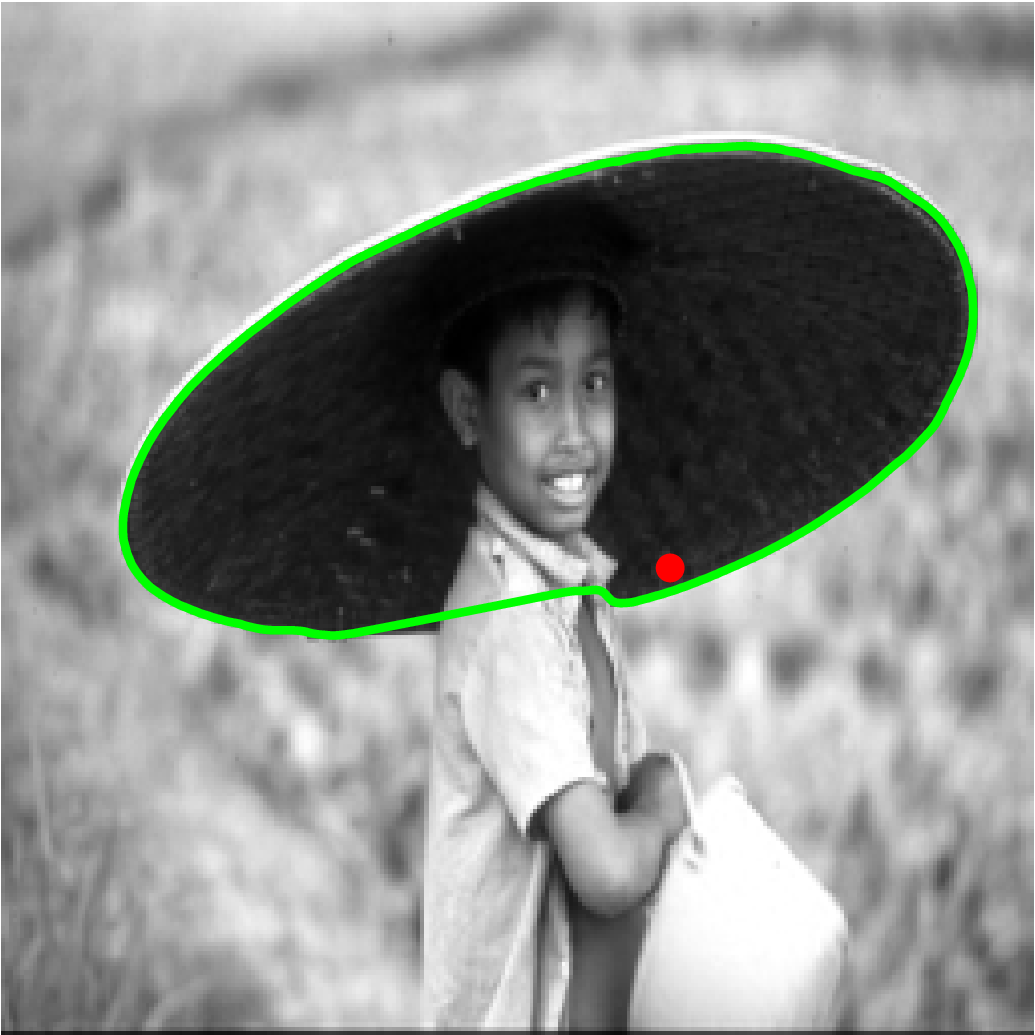}}
\hspace{0.5cm}
\subfigure[Segmentation by \eqref{proposed_model3} with two centers. \textcolor{black}{Running time 11.3 seconds.}]{
\includegraphics[width=1.13in,height=1.13in]{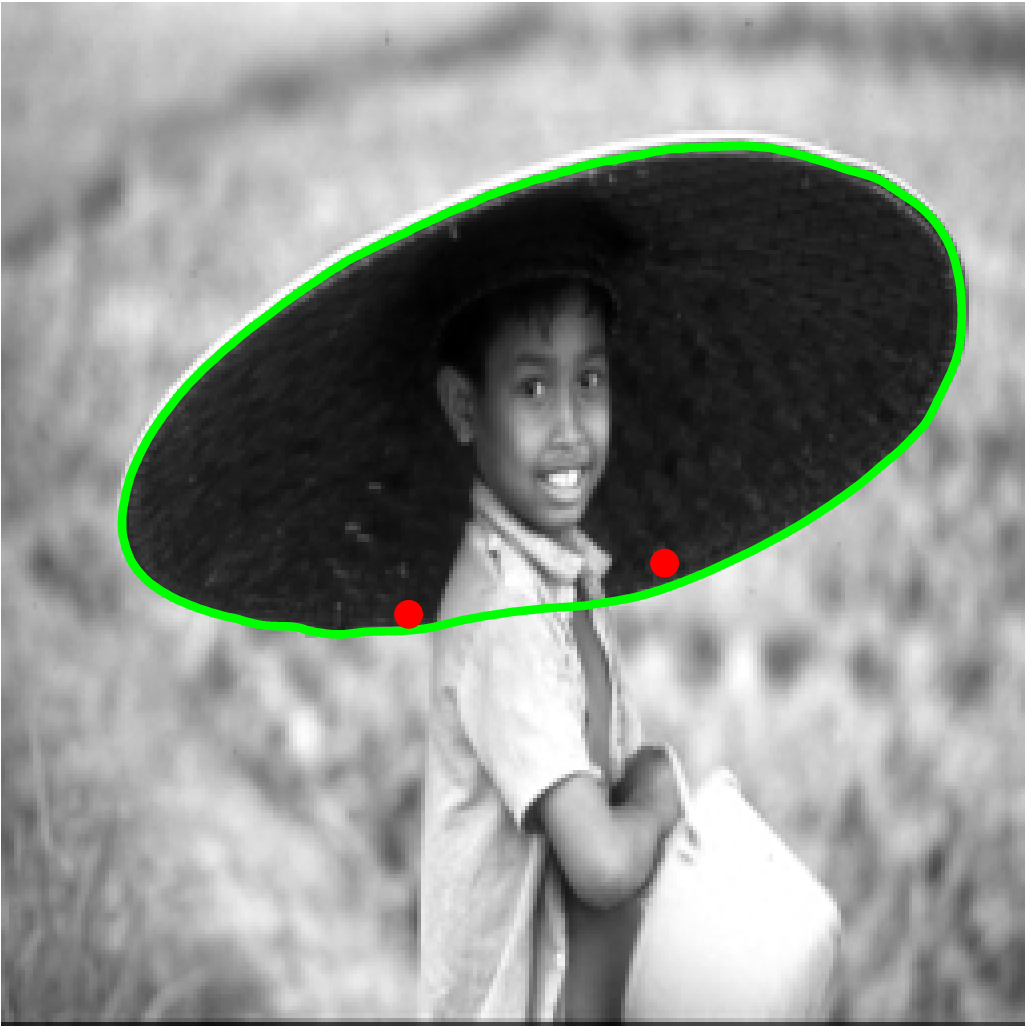}}
\caption{Test on the proposed model \eqref{proposed_model3}. (a) Target image and the initial contour. (b) Segmentation result by the CV model. \textcolor{black}{(c) Segmentation result by the convexity-preserving model \cite{zhang2021topology2}.} (d) Segmentation result by the proposed model \eqref{proposed_model1} with one center (red point). (e) Segmentation result by the proposed model \eqref{proposed_model1} with one center (red point). (f) Segmentation result by the proposed model \eqref{proposed_model3} with two centers (red points).}\label{Exp2_fig3}
\end{figure}

\begin{figure}[htbp!]
\centering
\subfigure[Target image and the initial contour]{
\includegraphics[width=1.13in,height=1.13in]{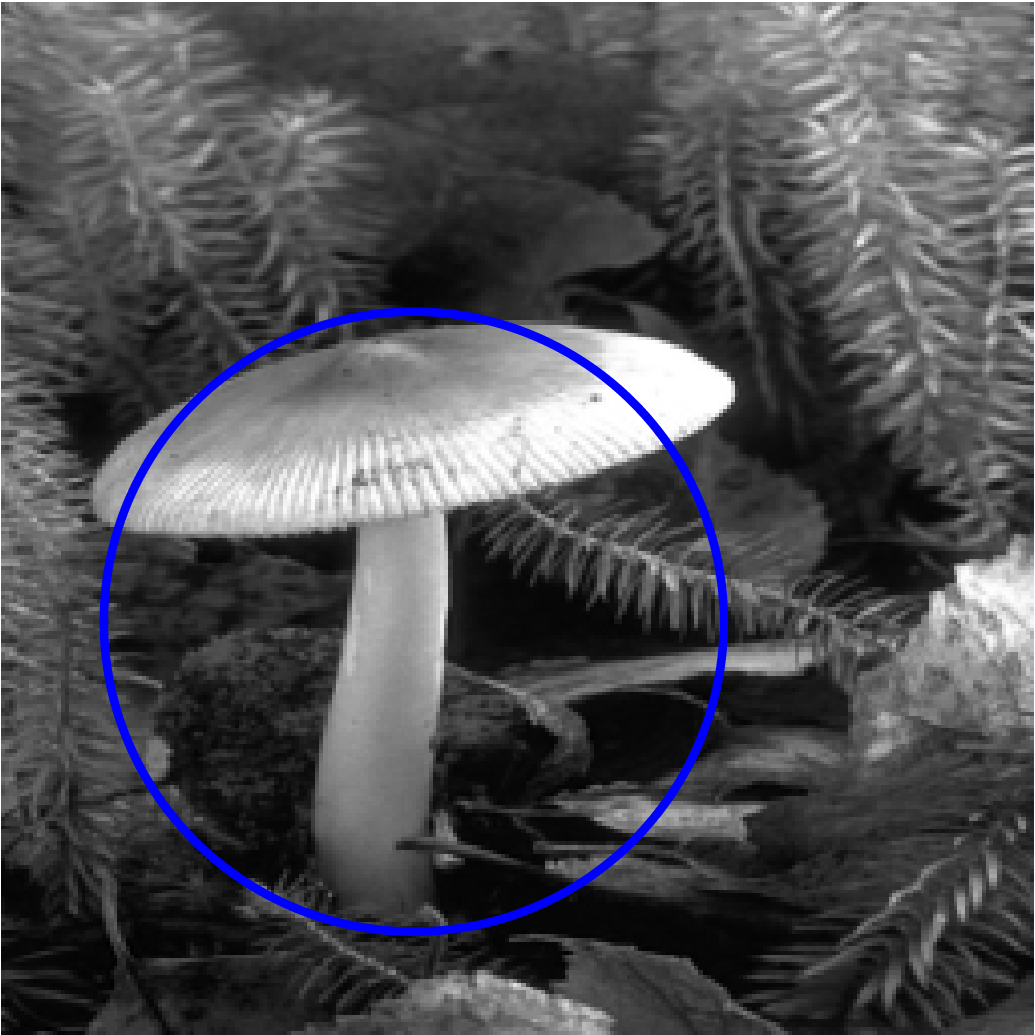}}
\hspace{0.5cm}
\subfigure[Segmentation by CV. \textcolor{black}{Running time 3.6 seconds.}]{
\includegraphics[width=1.13in,height=1.13in]{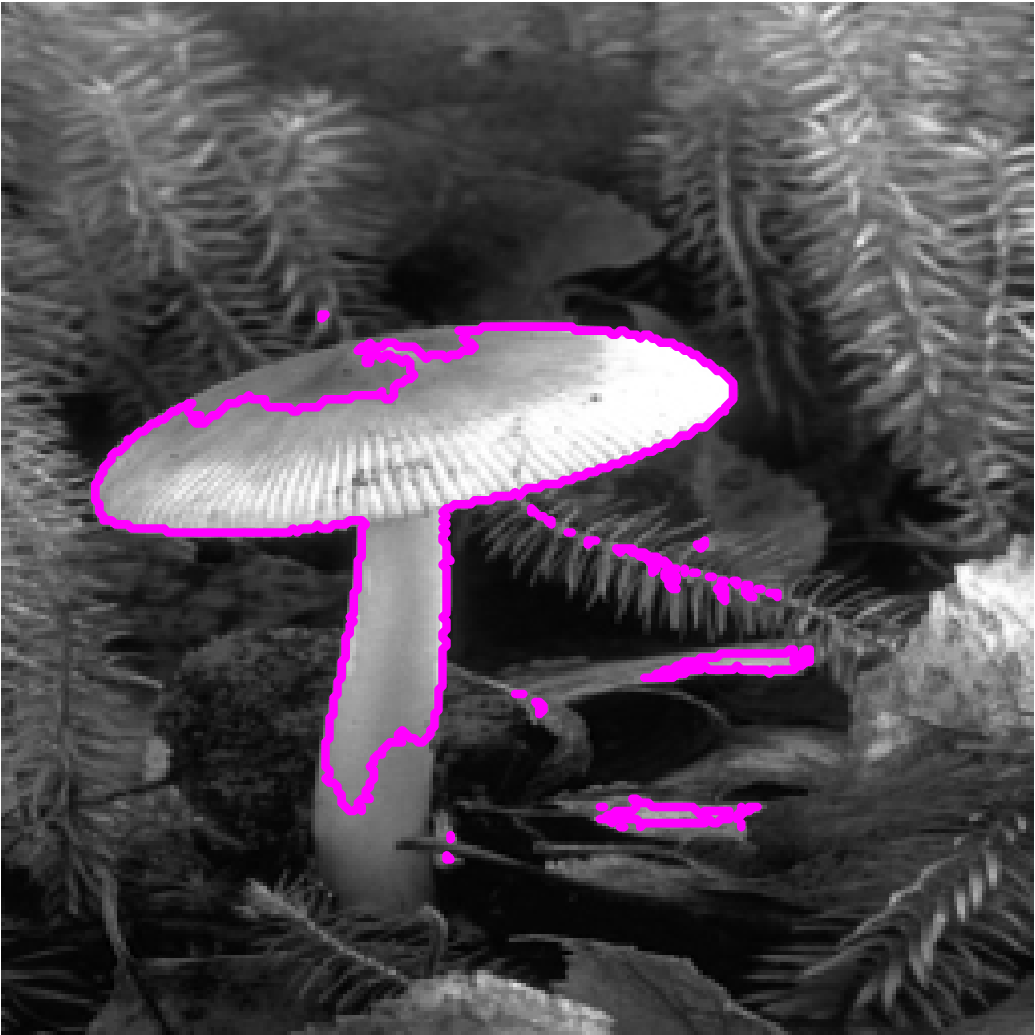}} 
\hspace{0.5cm}
\subfigure[\textcolor{black}{Segmentation by the convexity-preserving model. Running time 27.4 seconds.}]{
\includegraphics[width=1.13in,height=1.13in]{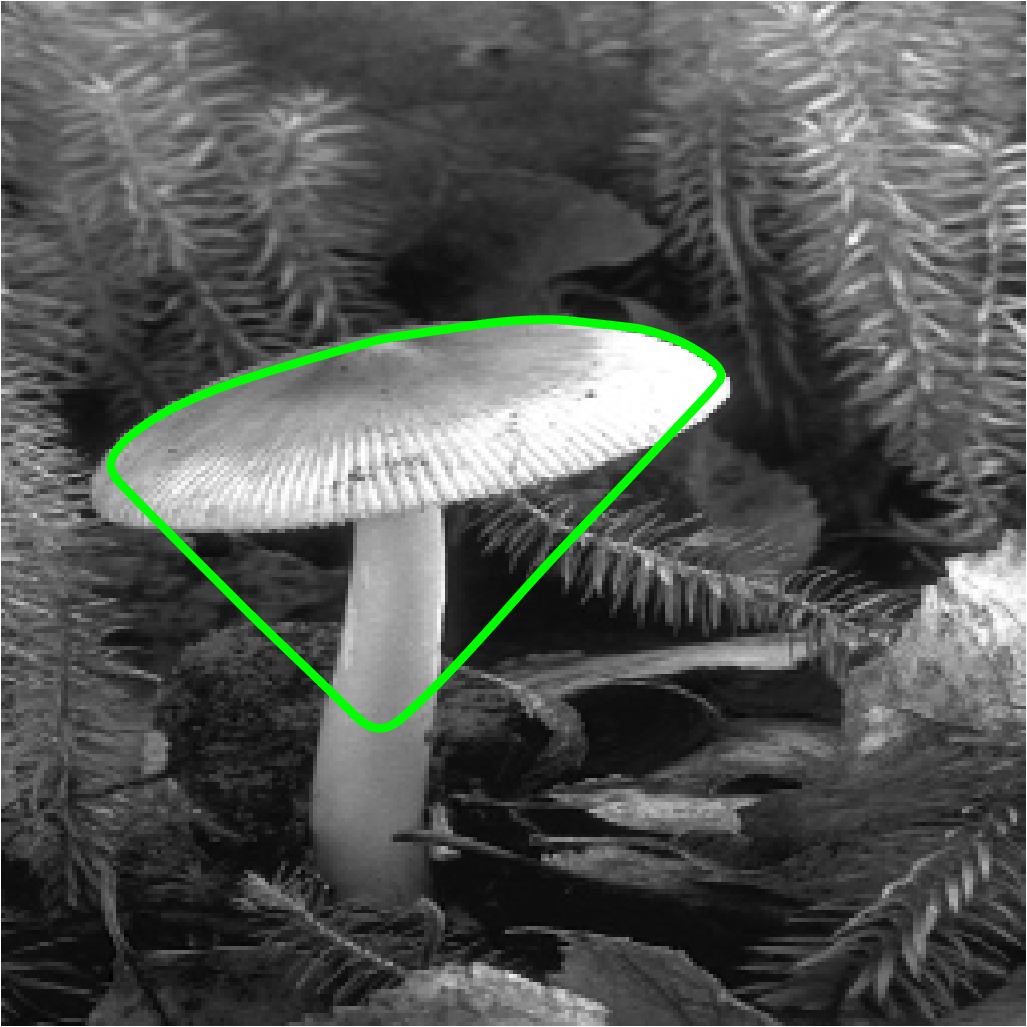}} 
\hspace{0.5cm} \\
\subfigure[Segmentation by \eqref{proposed_model1} with one center. \textcolor{black}{Running time 6.9 seconds.}]{
\includegraphics[width=1.13in,height=1.13in]{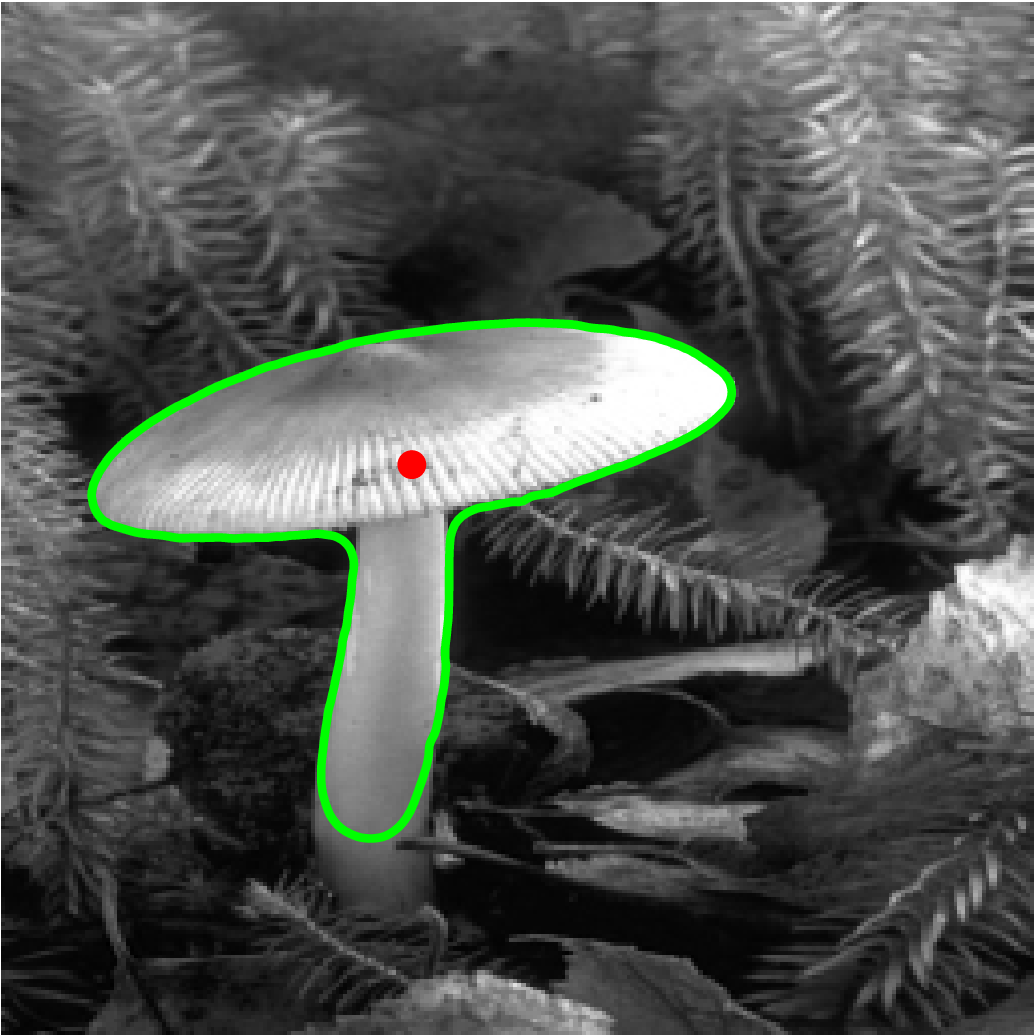}}
\hspace{0.5cm} 
\subfigure[The locally restricted region and centers for \eqref{proposed_model3}]{
\includegraphics[width=1.13in,height=1.13in]{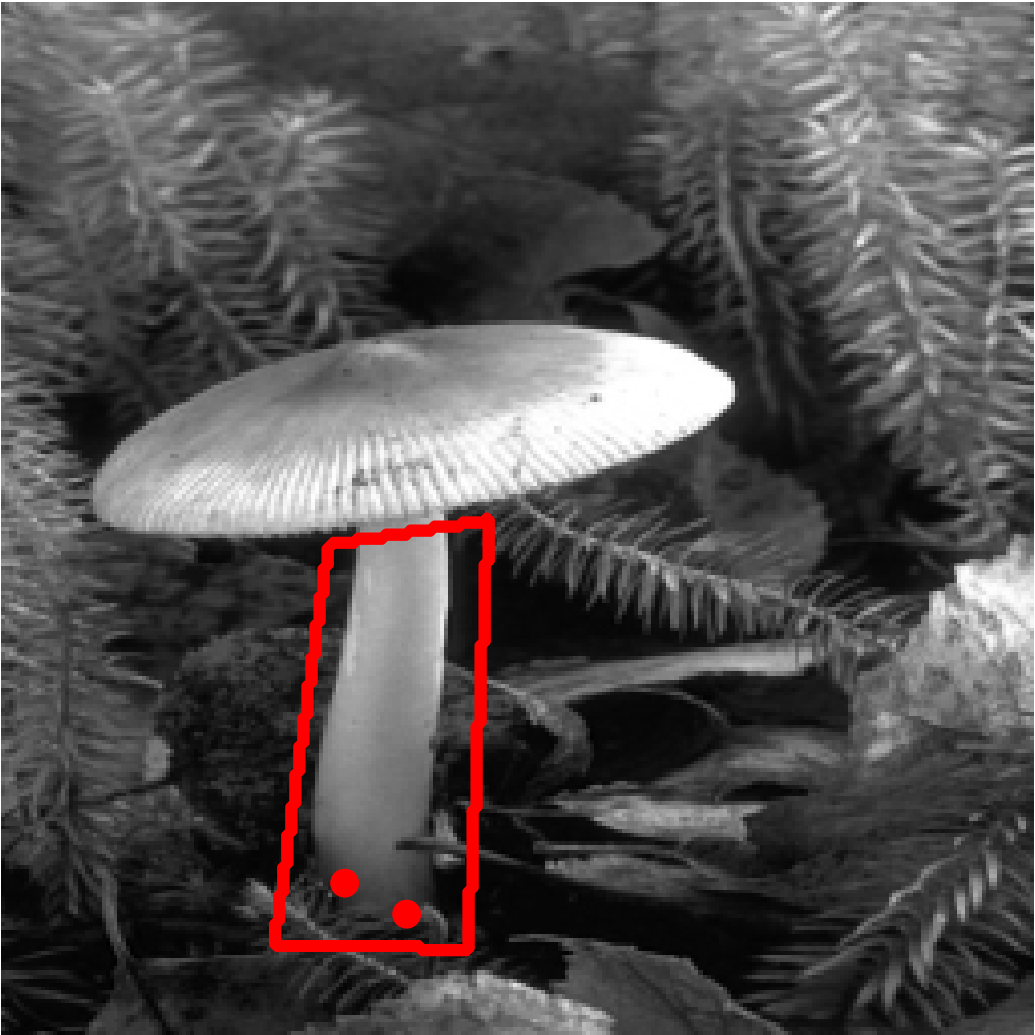}}
\hspace{0.5cm}
\subfigure[Segmentation by \eqref{proposed_model3} with two centers for the locally restricted region. \textcolor{black}{Running time 4.7 seconds.}]{
\includegraphics[width=1.13in,height=1.13in]{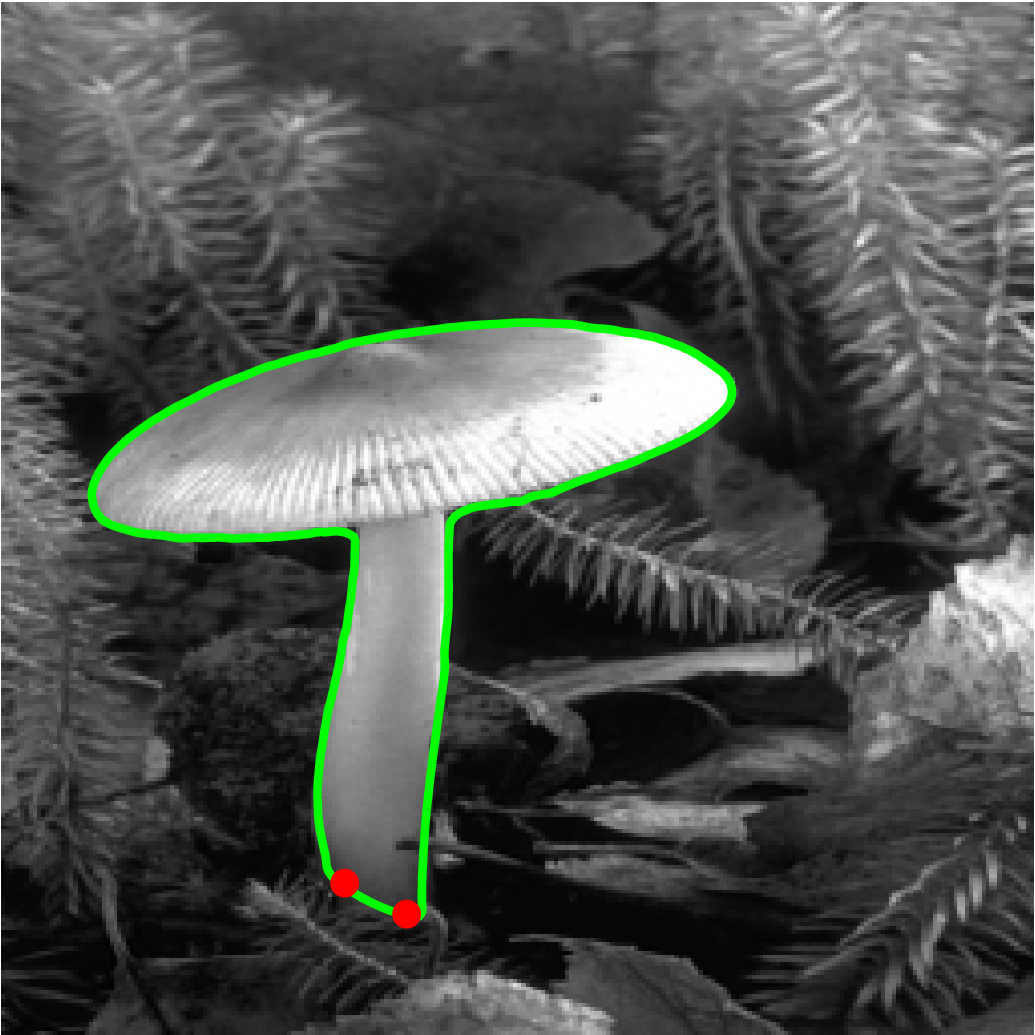}}
\caption{Test on the proposed model \eqref{proposed_model3}. (a) Target image and the initial contour. (b) Segmentation result by the CV model. \textcolor{black}{(c) Segmentation result by the convexity-preserving model \cite{zhang2021topology2}.} (d) Segmentation result by the proposed model \eqref{proposed_model1} with one center (red point). (e) The locally restricted region (red region) and centers (red points) for the proposed model \eqref{proposed_model3}. (f) Segmentation result by the proposed model \eqref{proposed_model3} with two centers for the locally restricted region.}\label{Exp2_fig4}
\end{figure}

\begin{figure}[htbp!]
\centering
\subfigure[\textcolor{black}{Target image and different initial contours}]{
\includegraphics[width=1.13in,height=1.13in]{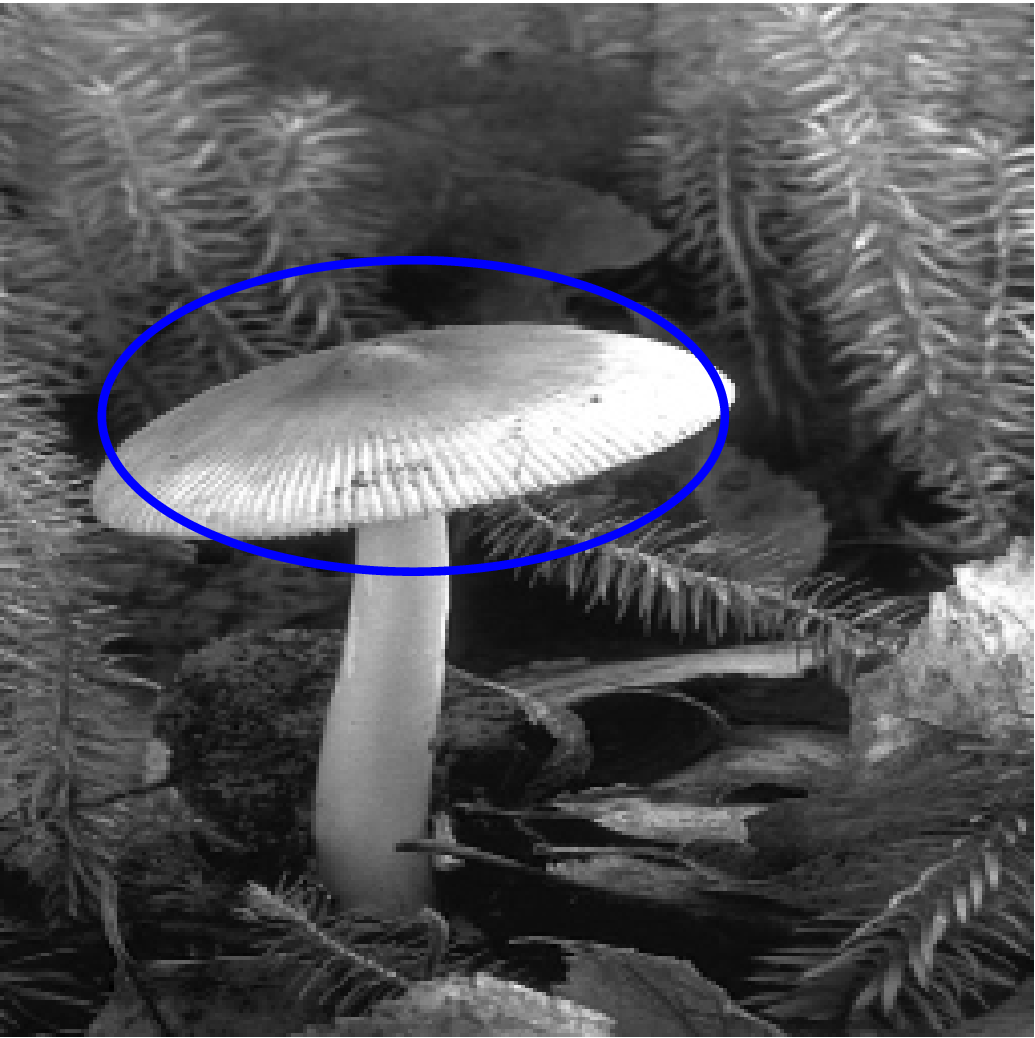}
\includegraphics[width=1.13in,height=1.13in]{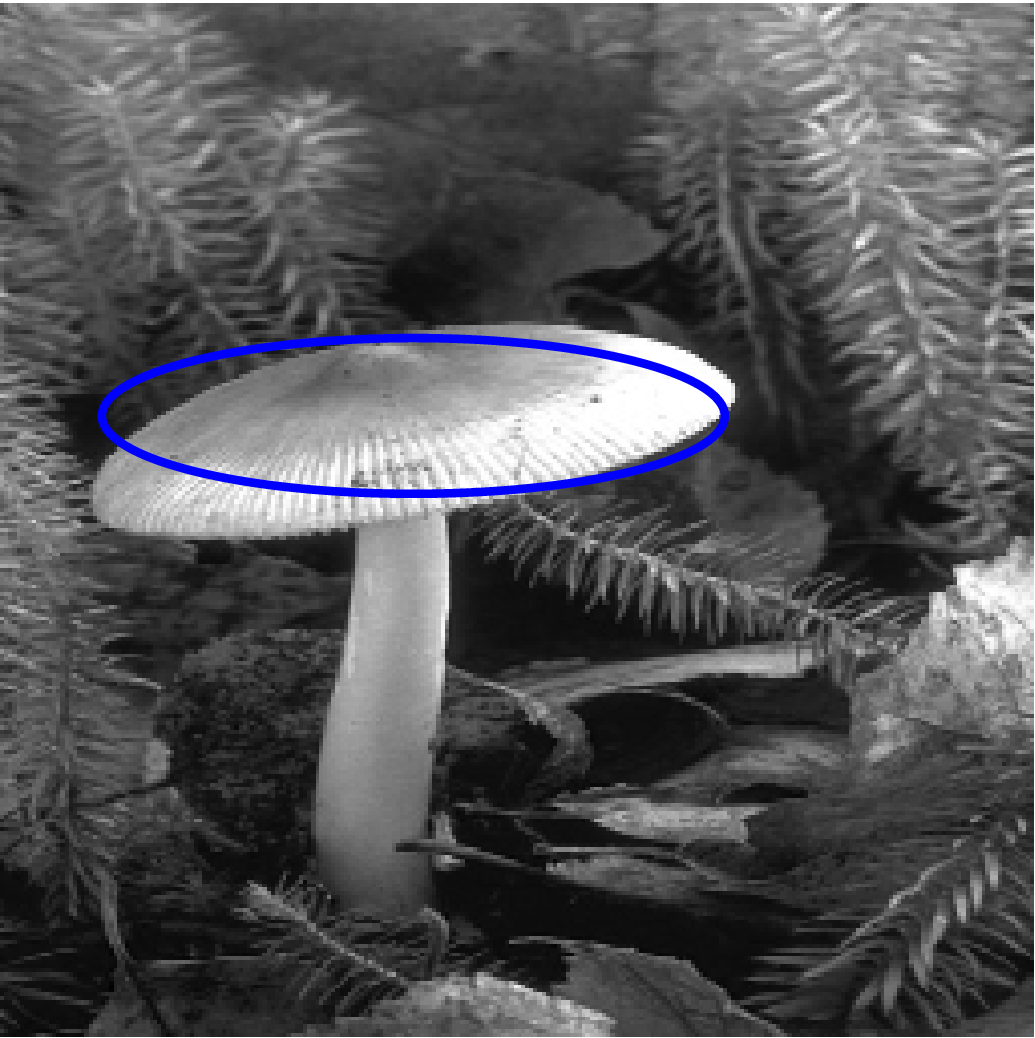}
\includegraphics[width=1.13in,height=1.13in]{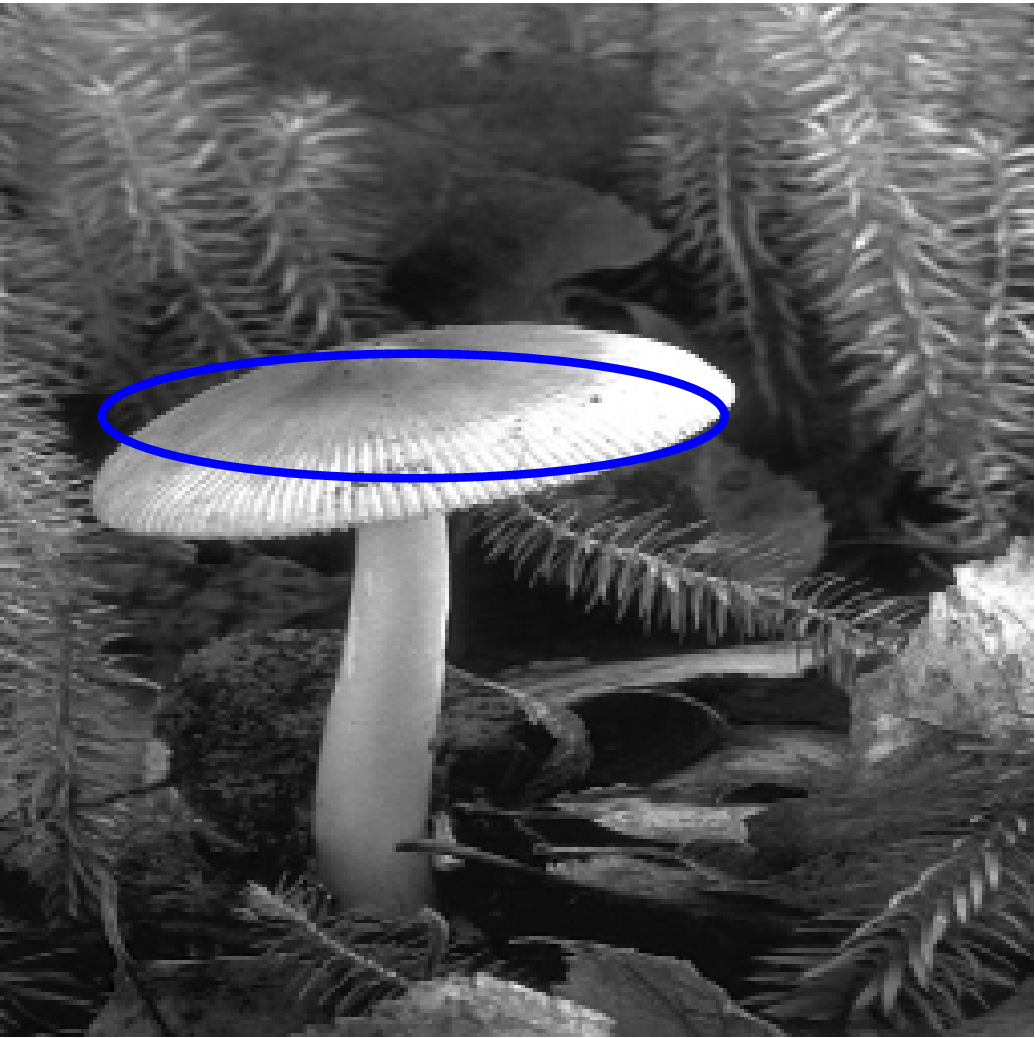}} 
\subfigure[\textcolor{black}{The corresponding segmentations by the convexity-preserving model. The running times are 46.5 s, 43.4 s, and 23.6 s.}]{
\includegraphics[width=1.13in,height=1.13in]{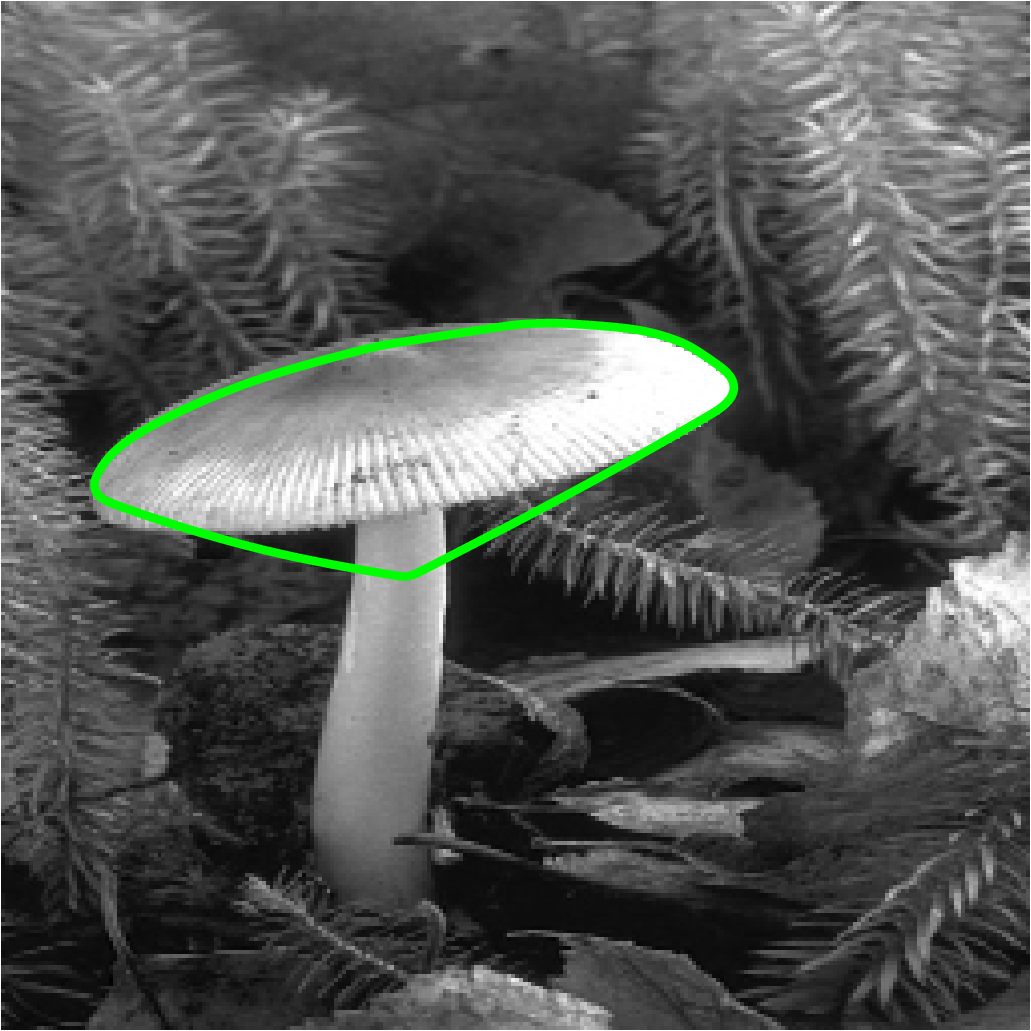}
\includegraphics[width=1.13in,height=1.13in]{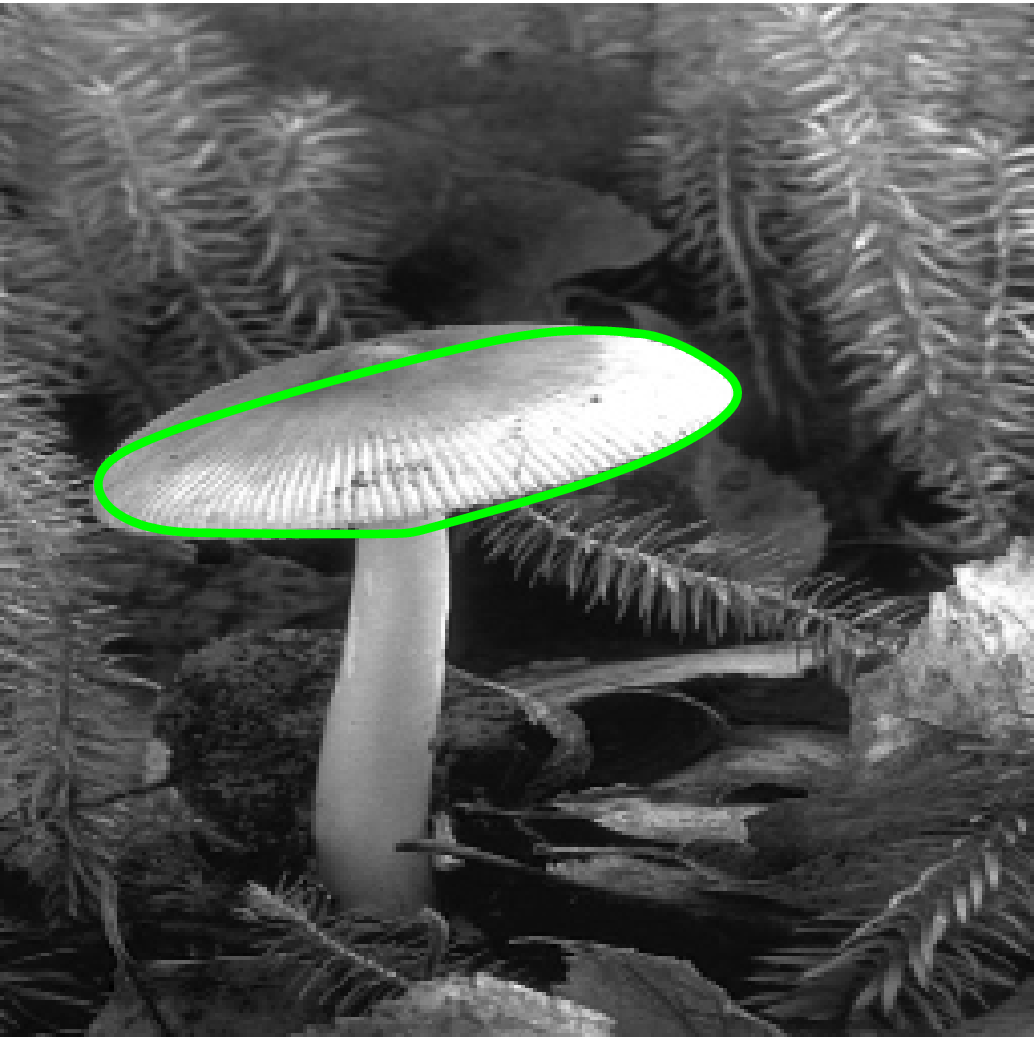}
\includegraphics[width=1.13in,height=1.13in]{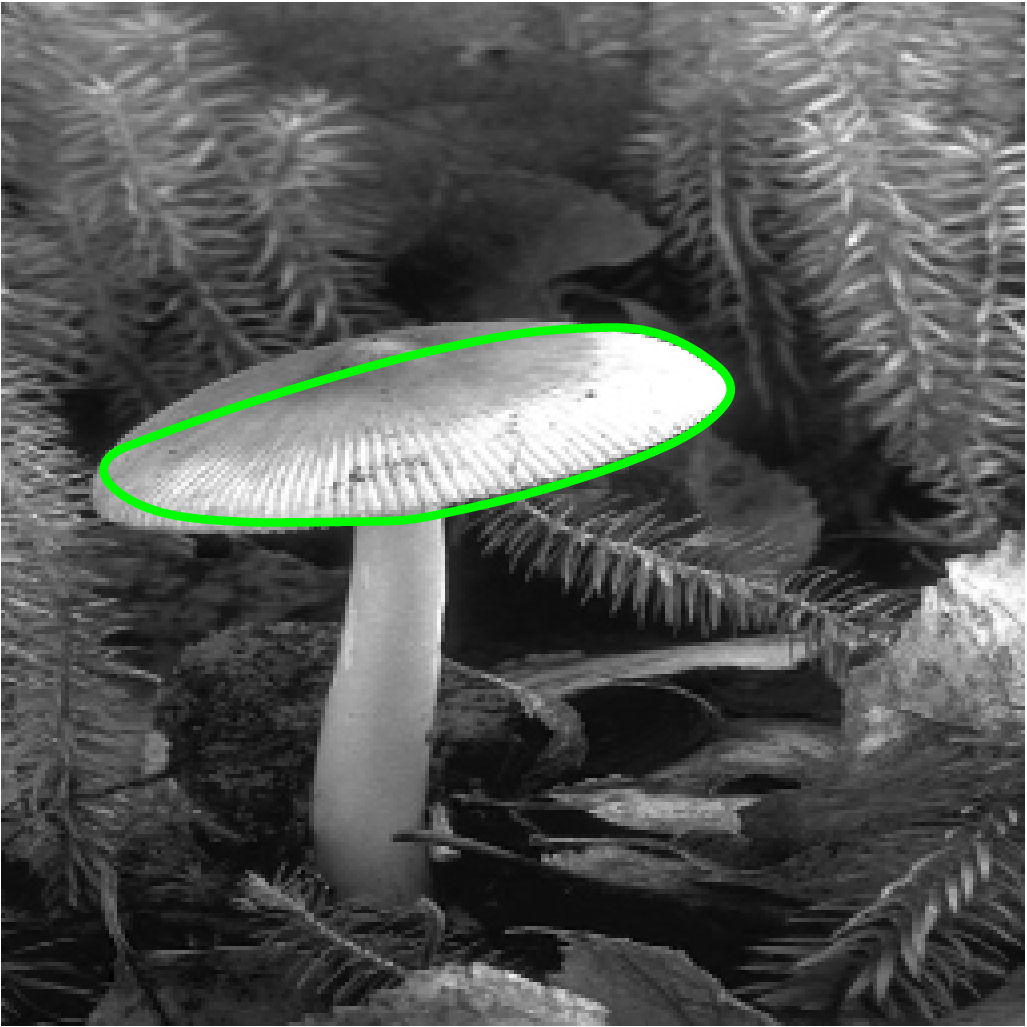}}
\caption{\textcolor{black}{Test on the initial contours for the convexity-preserving model \cite{zhang2021topology2}. The first row gives the different initial contours and the second row displays the corresponding segmentation results by the convexity-preserving model \cite{zhang2021topology2}. The running time is measured by seconds.}}\label{initial_contour}
\end{figure}

Next, we highlight the advantages of the proposed models \eqref{proposed_model2} and \eqref{proposed_model3} using two images, showcased in Fig. \ref{Exp2_fig3}(a) and \ref{Exp2_fig4}(a). Here, the parameter $\alpha$ in both models is set to $0.01$. For the first image (Fig. \ref{Exp2_fig3}(a)), as depicted in Fig. \ref{Exp2_fig3}(b), the CV model fails to produce an accurate segmentation due to occlusion, resulting in numerous outliers. To achieve precise segmentation, we employ a single center, yielding a star-shape segmentation as shown in Fig. \ref{Exp2_fig3}(d-e), derived from either the proposed model \eqref{proposed_model1} or the special case of the proposed model \eqref{proposed_model2}. Although this segmentation outperforms the CV model, further enhancement is attained by introducing an additional center, as demonstrated in Fig. \eqref{Exp2_fig3}(f). This illustrates that our proposed model \eqref{proposed_model3} can accurately segment intricate structures with occlusion by strategically setting center points. \textcolor{black}{While the convexity-preserving model \cite{zhang2021topology2} can also give a satisfied segmentation result (Fig. \ref{Exp2_fig3}(c)), its running time is a bit longer than the proposed model \eqref{proposed_model3}.} Moving to the second image (Fig. \ref{Exp2_fig4}(a)), the CV model again yields subpar segmentation (Fig. \ref{Exp2_fig4}(b)), mirroring previous cases. \textcolor{black}{The convexity-preserving model \cite{zhang2021topology2} also generates an unwanted segmentation (Fig. \ref{Exp2_fig4}(c)).} To improve the segmentation result, we employ our proposed model \eqref{proposed_model1}, resulting in a star-shape segmentation within the global restricted region. However, Fig. \ref{Exp2_fig4}(d) reveals inaccuracies stemming from the inhomogeneous intensity at the mushroom's root, which indicates that for this scenario, the global star-shape segmentation model \eqref{proposed_model1} is inadequate.
Addressing this, we use the proposed model \eqref{proposed_model3}, which can employ multiple centers and locally restrict the constraints (Fig. \ref{Exp2_fig4}(e)). From Fig. \ref{Exp2_fig4}(f), we can see that it leads to a satisfactory segmentation outcome. This illustrates that incorporating the centers and restricted regions suitably can significantly enhance the accuracy of the segmentation when dealing with inhomogeneous intensity or more complex structures. \textcolor{black}{Here, we further examine the performance of the convexity-preserving model \cite{zhang2021topology2} with different initial contour placements. As illustrated in Fig. \ref{initial_contour}, even when the initial contours are positioned in close proximity to the mushroom cap (a naturally convex domain), the segmentation results remain unsatisfactory. This observation reinforces the limitation of the convexity-preserving approach \cite{zhang2021topology2} in handling complex scenarios, where the average fitting term and convex constraint may conflict with accurate boundary delineation.}

\begin{figure}[htbp!]
\centering
\subfigure[Target image with the initial contour]{
\includegraphics[width=1.13in,height=1.13in]{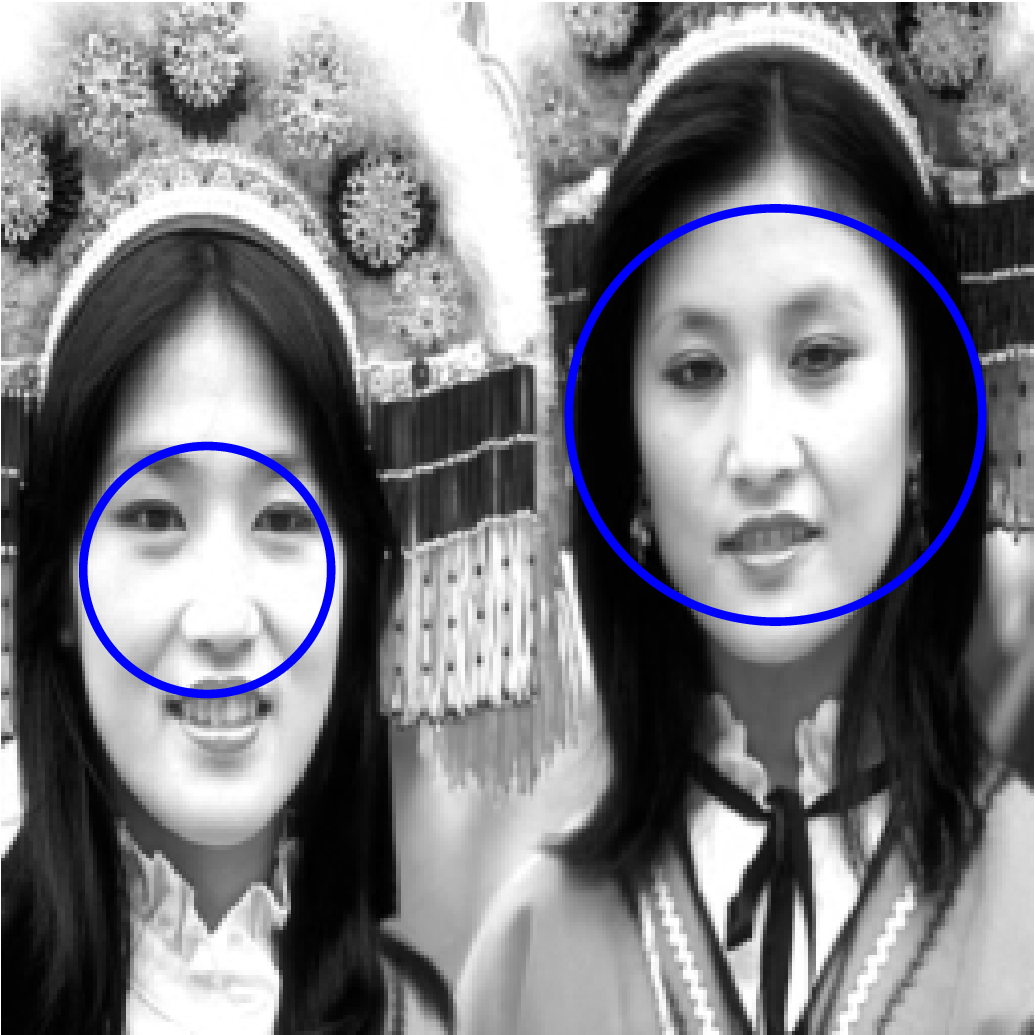}}
\hspace{0.5cm}
\subfigure[Segmentation by CV. \textcolor{black}{Running time 3.6 seconds.}]{
\includegraphics[width=1.13in,height=1.13in]{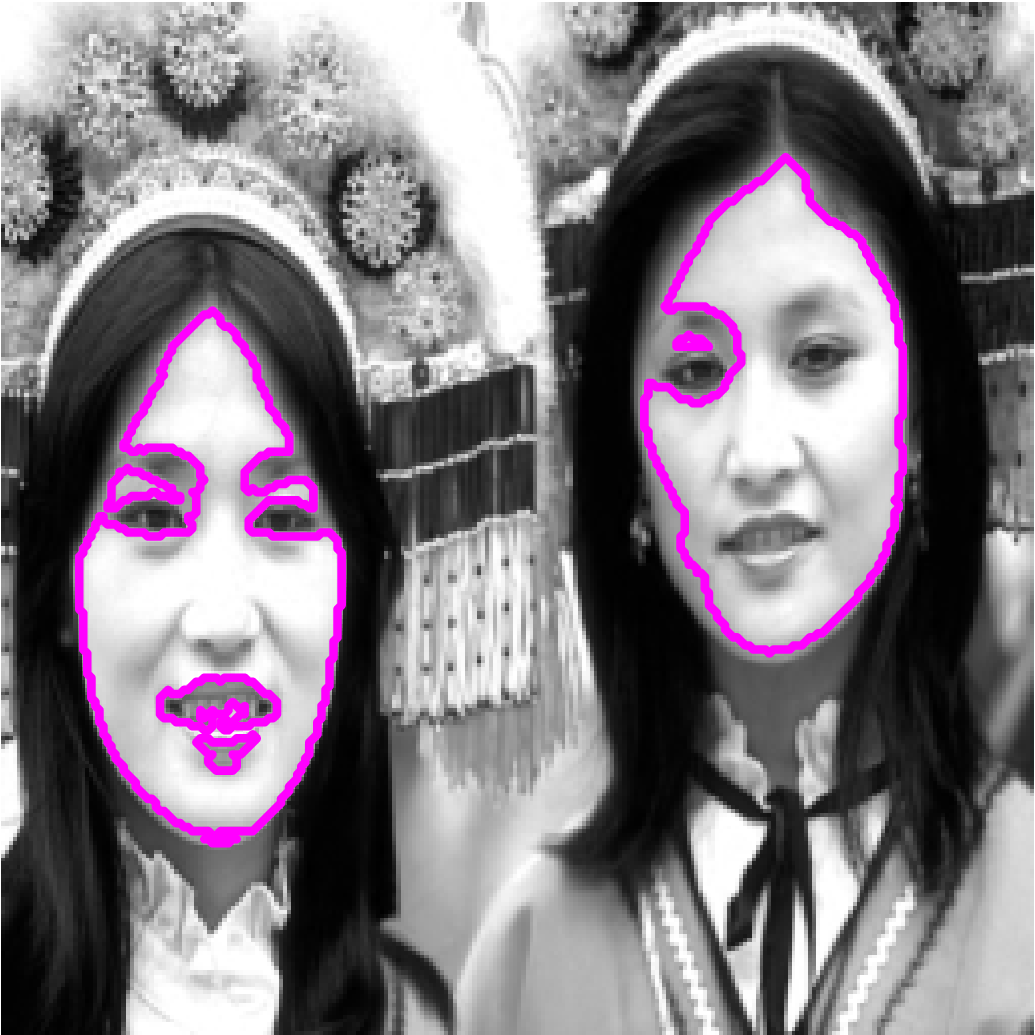}}
\hspace{0.5cm}
\subfigure[\textcolor{black}{Segmentation by the convexity-preserving model. Running time 32.6 seconds.}]{
\includegraphics[width=1.13in,height=1.13in]{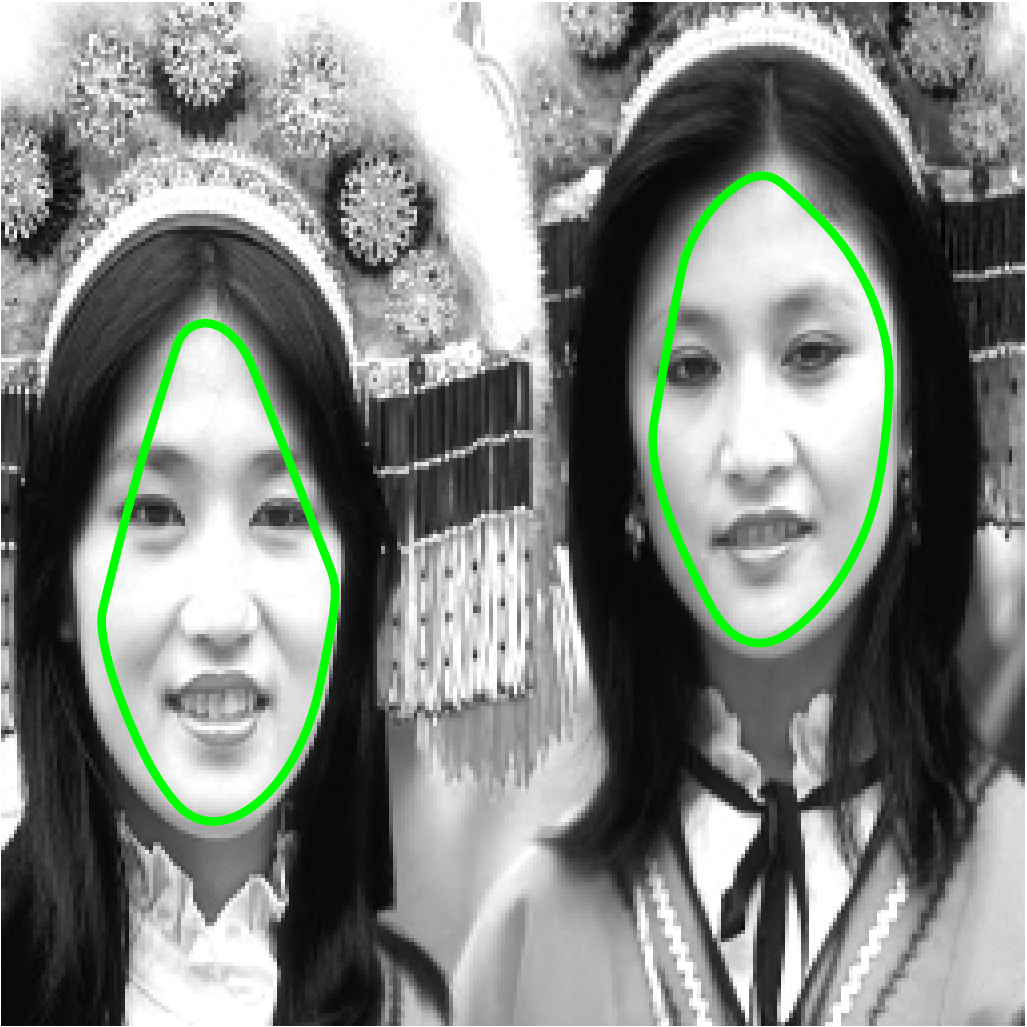}}
\hspace{0.5cm} \\
\subfigure[Segmentation by \eqref{proposed_model4} with one center. \textcolor{black}{Running time 2.2 seconds.}]{
\includegraphics[width=1.13in,height=1.13in]{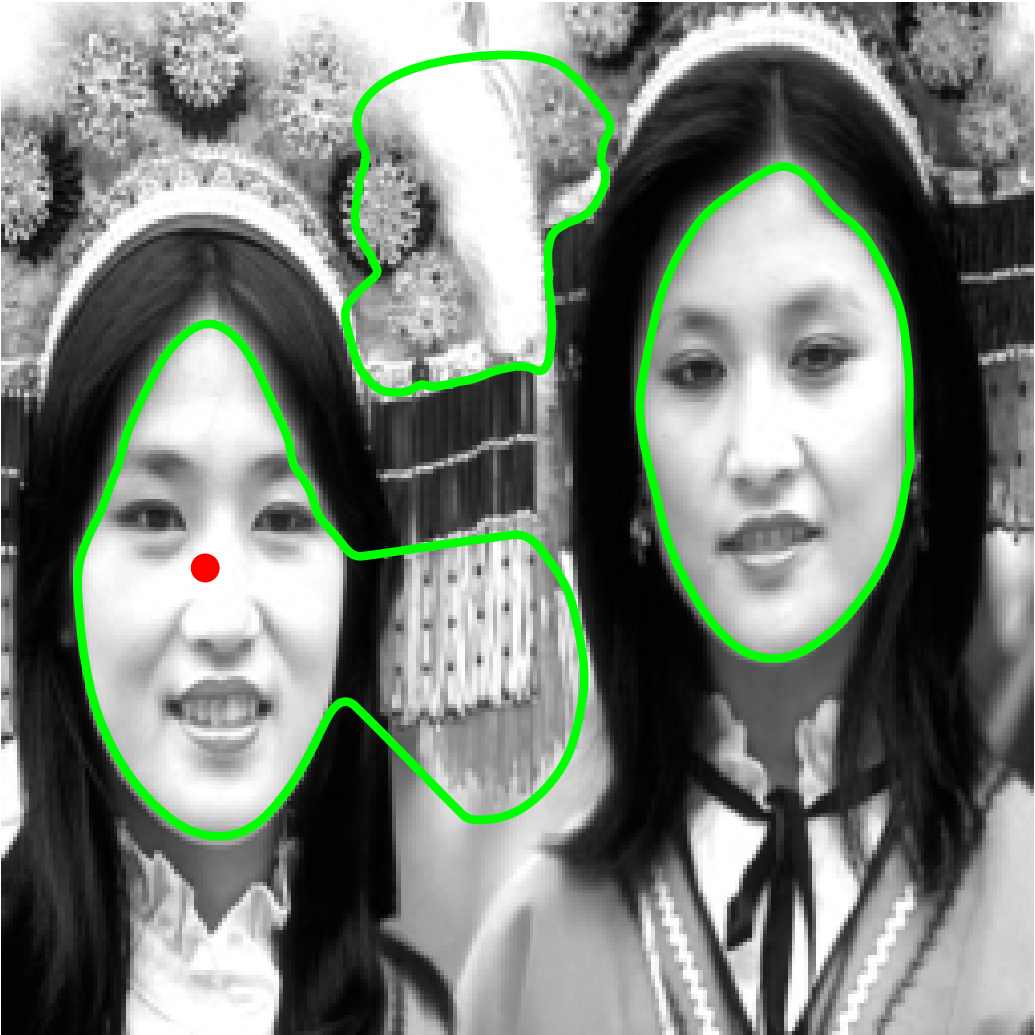}}
\hspace{0.5cm}
\subfigure[Segmentation by \eqref{proposed_model4} with one center. \textcolor{black}{Running time 1.8 seconds.}]{
\includegraphics[width=1.13in,height=1.13in]{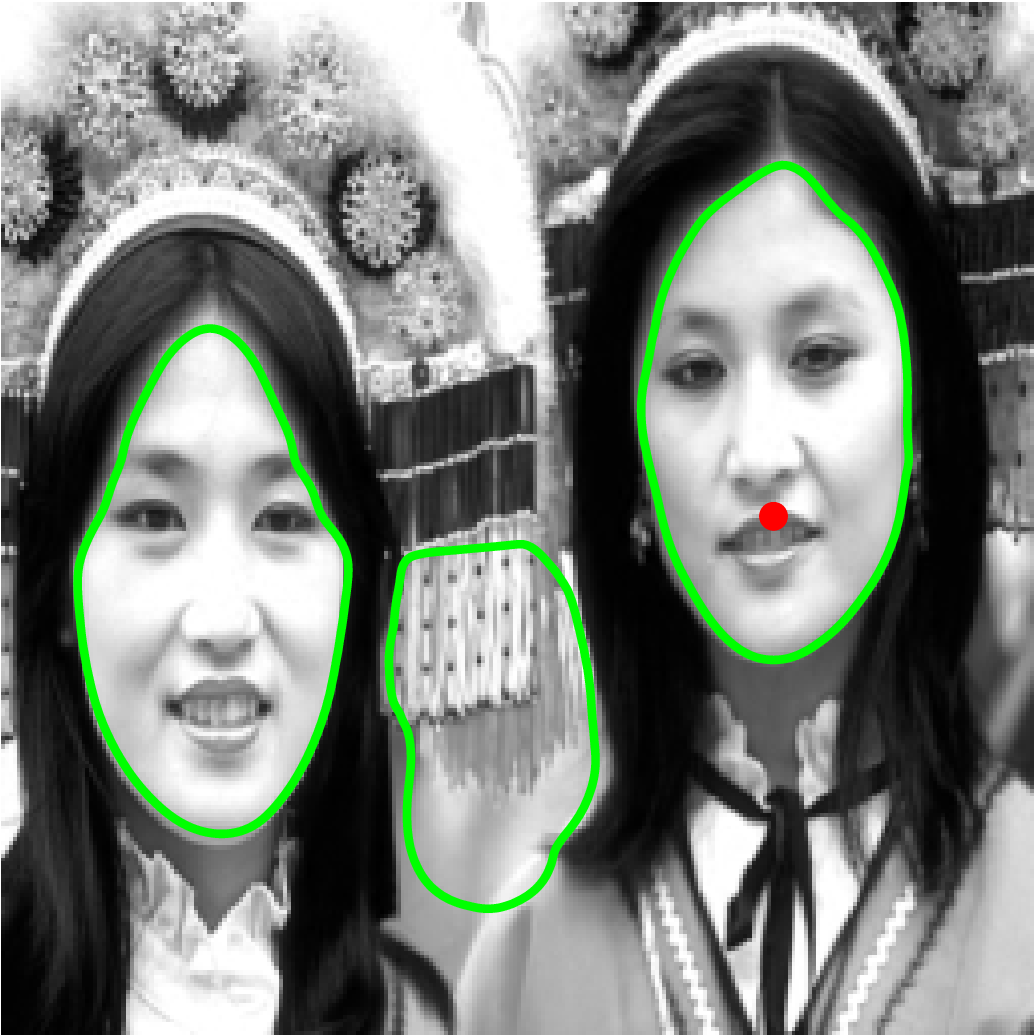}}
\hspace{0.5cm}
\subfigure[Segmentation by \eqref{proposed_model4} with two centers. \textcolor{black}{Running time 1.2 seconds.}]{
\includegraphics[width=1.13in,height=1.13in]{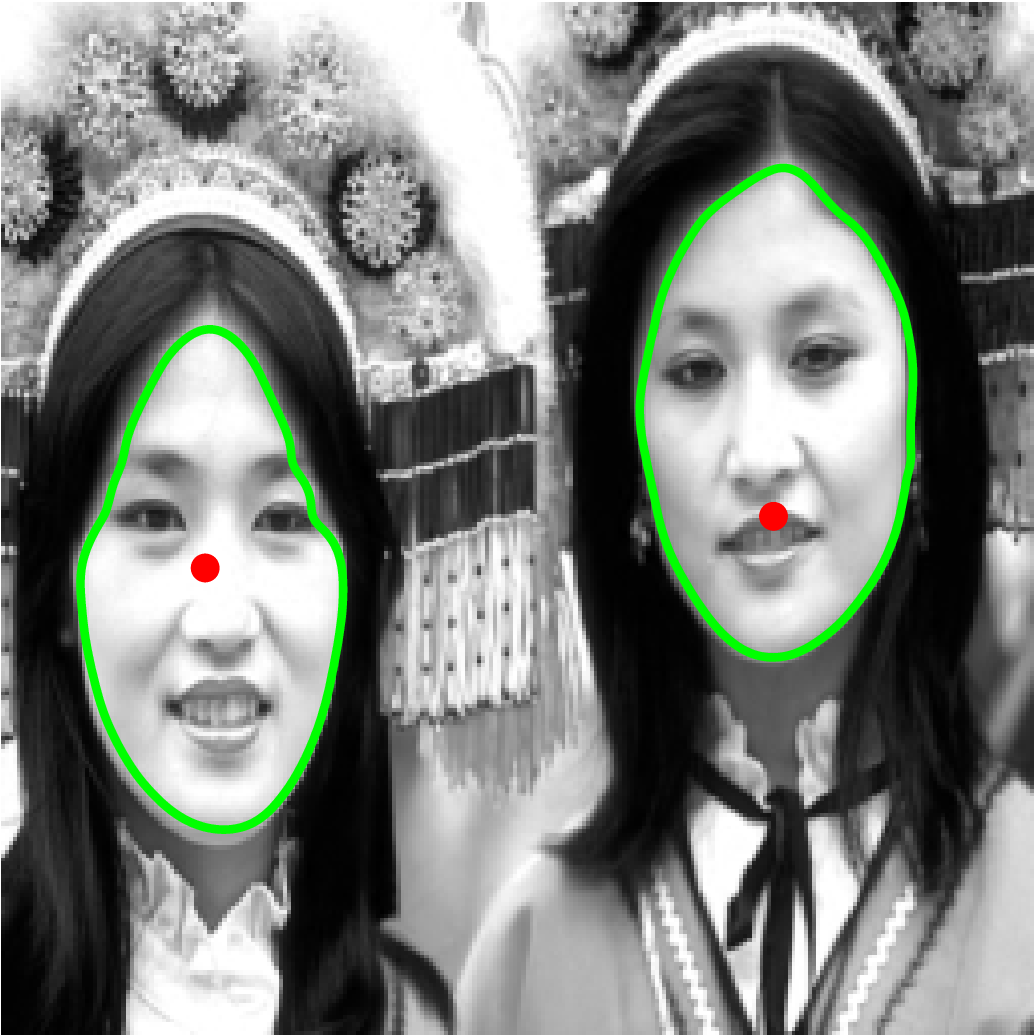}}
\caption{Test on the proposed model \eqref{proposed_model4}. (a) Target image with the initial contour. (b) Segmentation result by the CV model.
\textcolor{black}{(c) Segmentation result by the convexity-preserving model \cite{zhang2021topology2}.} (d) Segmentation result by the proposed model \eqref{proposed_model4} with one center (red point). (e) Segmentation result by the proposed model \eqref{proposed_model4} with one center (red point). (f) Segmentation result by the proposed model \eqref{proposed_model4} with two centers (red points).}\label{Exp2_fig5}
\end{figure}

Finally, we apply the proposed model \eqref{proposed_model4} using a previous example. Here, the parameter $\alpha$ is set to 0.02, and the results are shown in Fig. \ref{Exp2_fig5}. For this three-phase segmentation, initial contours are set as two circles (Fig. \ref{Exp2_fig5}(a)). The CV model produces a segmentation focusing on the bright parts but misses the overall connection (Fig. \ref{Exp2_fig5}(b)). Fig. \ref{Exp2_fig5}(d-f) display the results of the proposed model \eqref{proposed_model4} with a single center. In Fig. \ref{Exp2_fig5}(d), the left part forms a star-shape domain but includes an undesired region, while the right part consists of two components due to the lack of constraints ensuring that the region remains whole. Changing the center's position (Fig. \ref{Exp2_fig5}(f)) results in a well-segmented right part, but again, the left part contains two components. To address this, we enhance the proposed model \eqref{proposed_model4} by introducing two centers. This modification imposes constraints to maintain the desired star-shape domains, resulting in a satisfactory segmentation (Fig. \ref{Exp2_fig5}(e)). \textcolor{black}{For the convexity-preserving model \cite{zhang2021topology2}, while it  yields satisfactory segmentation results (Fig. \ref{Exp2_fig5}(c)), its computational efficiency is significantly inferior to that of the proposed model \eqref{proposed_model4}.}

\section{Conclusion}\label{Con}
In this paper, we introduce a novel star-shape segmentation framework leveraging registration techniques. By integrating the level set representation with a registration-based approach and imposing constraints on the deformed level set function, we formulate the star-shape segmentation model \eqref{proposed_model1}. Furthermore, we extend this model from a single center \eqref{proposed_model1} to accommodate specific region \eqref{proposed_model2}, multiple centers \eqref{proposed_model3}, and multiphase segmentation \eqref{proposed_model4}. By the advantage of the registration framework, we also incorporate the landmark constraints \eqref{proposed_model5}. The flexibility in selecting centers and restricted regions caters to diverse user requirements. To address the proposed model, we introduce an auxiliary variable and employ the alternating direction method of multipliers (ADMM). Subsequently, one subproblem is tackled using a modified Gauss-Newton method, while the other is resolved with a closed-form solution. Numerical experiments conducted on both synthetic and real images validate that our proposed models effectively achieves accurate star-shape segmentations.

%

%
 \section*{Conflict of interest}

 The authors declare that they have no conflict of interest.
 
 \section*{Data Availability}

The datasets used in this study are available from the first author on reasonable request.

\bibliographystyle{spmpsci}      
\bibliography{reference} 

\begin{thebibliography}{10}
\providecommand{\url}[1]{{#1}}
\providecommand{\urlprefix}{URL }
\expandafter\ifx\csname urlstyle\endcsname\relax
  \providecommand{\doi}[1]{DOI~\discretionary{}{}{}#1}\else
  \providecommand{\doi}{DOI~\discretionary{}{}{}\begingroup
  \urlstyle{rm}\Url}\fi

\bibitem{Bae2010a}
Bae, E., Yuan, J., Tai, X.C.: Global minimization for continuous multiphase
  partitioning problems using a dual approach.
\newblock International Journal of Computer Vision \textbf{92}(1), 112--129
  (2011)

\bibitem{Bae2013a}
Bae, E., Yuan, J., Tai, X.C.: Simultaneous convex optimization of regions and
  region parameters in image segmentation models.
\newblock Innovations for Shape Analysis: Models and Algorithms pp. 421--438
  (2013)

\bibitem{boyd2011distributed}
Boyd, S., Parikh, N., Chu, E.: Distributed optimization and statistical
  learning via the alternating direction method of multipliers.
\newblock Now Publishers Inc (2011)

\bibitem{broit1981optimal}
Broit, C.: Optimal registration of deformed images.
\newblock Ph.D. thesis, University of Pennsylvania, USA (1981)

\bibitem{burger2013hyperelastic}
Burger, M., Modersitzki, J., Ruthotto, L.: A hyperelastic regularization energy
  for image registration.
\newblock SIAM Journal on Scientific Computing \textbf{35}(1), B132--B148
  (2013)

\bibitem{cai2013two}
Cai, X., Chan, R., Zeng, T.: A two-stage image segmentation method using a
  convex variant of the mumford--shah model and thresholding.
\newblock SIAM Journal on Imaging Sciences \textbf{6}(1), 368--390 (2013)

\bibitem{caselles1997geodesic}
Caselles, V., Kimmel, R., Sapiro, G.: Geodesic active contours.
\newblock International Journal of Computer Vision \textbf{22}(1), 61--79
  (1997)

\bibitem{chan2018topology}
Chan, H.L., Yan, S., Lui, L.M., Tai, X.C.: Topology-preserving image
  segmentation by beltrami representation of shapes.
\newblock Journal of Mathematical Imaging and Vision \textbf{60}(3), 401--421
  (2018)

\bibitem{chan2002active}
Chan, T., Sandberg, B., Vese, L.: Active contours without edges for textured
  images.
\newblock CAM report pp. 02--28 (2002)

\bibitem{chan2005level}
Chan, T., Zhu, W.: Level set based shape prior segmentation.
\newblock In: 2005 IEEE Computer Society Conference on Computer Vision and
  Pattern Recognition (CVPR'05), vol.~2, pp. 1164--1170. IEEE (2005)

\bibitem{chan2007some}
Chan, T.F., Moelich, M., Sandberg, B.: Some recent developments in variational
  image segmentation.
\newblock Image Processing Based on Partial Differential Equations pp. 175--210
  (2007)

\bibitem{chan2000active}
Chan, T.F., Sandberg, B.Y., Vese, L.A.: Active contours without edges for
  vector-valued images.
\newblock Journal of Visual Communication and Image Representation
  \textbf{11}(2), 130--141 (2000)

\bibitem{chan2001active}
Chan, T.F., Vese, L.A.: Active contours without edges.
\newblock IEEE Transactions on Image Processing \textbf{10}(2), 266--277 (2001)

\bibitem{chen2022geodesic}
Chen, D., Mirebeau, J.M., Shu, M., Tai, X., Cohen, L.D.: Geodesic models with
  convexity shape prior.
\newblock IEEE Transactions on Pattern Analysis and Machine Intelligence
  \textbf{45}(7), 8433--8452 (2023)

\bibitem{christensen1996deformable}
Christensen, G.E., Rabbitt, R.D., Miller, M.I., et~al.: Deformable templates
  using large deformation kinematics.
\newblock IEEE transactions on image processing \textbf{5}(10), 1435--1447
  (1996)

\bibitem{cremers2003towards}
Cremers, D., Sochen, N., Schn{\"o}rr, C.: Towards recognition-based variational
  segmentation using shape priors and dynamic labeling.
\newblock In: International Conference on Scale-Space Theories in Computer
  Vision, pp. 388--400. Springer (2003)

\bibitem{droske2004variational}
Droske, M., Rumpf, M.: A variational approach to nonrigid morphological image
  registration.
\newblock SIAM Journal on Applied Mathematics \textbf{64}(2), 668--687 (2004)

\bibitem{elnakib2011medical}
Elnakib, A., Gimel’farb, G., Suri, J.S., El-Baz, A.: Medical image
  segmentation: a brief survey.
\newblock In: Multi Modality State-of-the-Art Medical Image Segmentation and
  Registration Methodologies, pp. 1--39. Springer (2011)

\bibitem{fischer2002fast}
Fischer, B., Modersitzki, J.: Fast diffusion registration.
\newblock Contemporary Mathematics \textbf{313}, 117--128 (2002)

\bibitem{fischer2003curvature}
Fischer, B., Modersitzki, J.: Curvature based image registration.
\newblock Journal of Mathematical Imaging and Vision \textbf{18}(1), 81--85
  (2003)

\bibitem{gavlasova2006wavelet}
Gavlasov{\'a}, A., Proch{\'a}zka, A., Mudrov{\'a}, M.: Wavelet based image
  segmentation.
\newblock In: Proc. of the 14th Annual Conference Technical Computing, Prague,
  pp. 1--7 (2006)

\bibitem{gorelick2016convexity}
Gorelick, L., Veksler, O., Boykov, Y., Nieuwenhuis, C.: Convexity shape prior
  for binary segmentation.
\newblock IEEE transactions on Pattern Analysis and Machine Intelligence
  \textbf{39}(2), 258--271 (2016)

\bibitem{gould2009region}
Gould, S., Gao, T., Koller, D.: Region-based segmentation and object detection.
\newblock Advances in Neural Information Processing Systems \textbf{22} (2009)

\bibitem{gui2017medical}
Gui, L., Li, C., Yang, X.: Medical image segmentation based on level set and
  isoperimetric constraint.
\newblock Physica Medica \textbf{42}, 162--173 (2017)

\bibitem{haber2006intensity}
Haber, E., Modersitzki, J.: Intensity gradient based registration and fusion of
  multi-modal images.
\newblock In: International Conference on Medical Image Computing and
  Computer-Assisted Intervention, pp. 726--733. Springer (2006)

\bibitem{haber2007intensity}
Haber, E., Modersitzki, J.: Intensity gradient based registration and fusion of
  multi-modal images.
\newblock Methods of Information in Medicine \textbf{46}(03), 292--299 (2007)

\bibitem{hodneland2009four}
Hodneland, E., Tai, X.C., Gerdes, H.H.: Four-color theorem and level set
  methods for watershed segmentation.
\newblock International journal of computer vision \textbf{82}, 264--283 (2009)

\bibitem{kamencay2013novel}
Kamencay, P., Zachariasova, M., Hudec, R., Jarina, R., Benco, M., Hlubik, J.: A
  novel approach to face recognition using image segmentation based on spca-knn
  method.
\newblock Radioengineering \textbf{22}(1), 92--99 (2013)

\bibitem{kass1988snakes}
Kass, M., Witkin, A., Terzopoulos, D.: Snakes: Active contour models.
\newblock International Journal of Computer Vision \textbf{1}(4), 321--331
  (1988)

\bibitem{kaymak2019brief}
Kaymak, {\c{C}}., U{\c{c}}ar, A.: A brief survey and an application of semantic
  image segmentation for autonomous driving.
\newblock Handbook of Deep Learning Applications pp. 161--200 (2019)

\bibitem{le2011combined}
Le~Guyader, C., Vese, L.A.: A combined segmentation and registration framework
  with a nonlinear elasticity smoother.
\newblock Computer Vision and Image Understanding \textbf{115}(12), 1689--1709
  (2011)

\bibitem{lee2016landmark}
Lee, Y.T., Lam, K.C., Lui, L.M.: Landmark-matching transformation with large
  deformation via n-dimensional quasi-conformal maps.
\newblock Journal of Scientific Computing \textbf{67}(3), 926--954 (2016)

\bibitem{li2007active}
Li, B., Acton, S.T.: Active contour external force using vector field
  convolution for image segmentation.
\newblock IEEE transactions on Image Processing \textbf{16}(8), 2096--2106
  (2007)

\bibitem{luo2020level}
Li, L., Luo, S., Tai, X.C., Yang, J.: A level set representation method for
  n-dimensional convex shape and applications.
\newblock Communications in Mathematical Research \textbf{37}(2), 180--208
  (2021)

\bibitem{lie2006variant}
Lie, J., Lysaker, M., Tai, X.C.: A variant of the level set method and
  applications to image segmentation.
\newblock Mathematics of Computation \textbf{75}(255), 1155--1174 (2006)

\bibitem{liu2022deep}
Liu, J., Wang, X., Tai, X.C.: Deep convolutional neural networks with spatial
  regularization, volume and star-shape priors for image segmentation.
\newblock Journal of Mathematical Imaging and Vision \textbf{64}(6), 625--645
  (2022)

\bibitem{luo2022convex}
Luo, S., Tai, X.C., Glowinski, R.: Convex object (s) characterization and
  segmentation using level set function.
\newblock Journal of Mathematical Imaging and Vision \textbf{64}(1), 68--88
  (2022)

\bibitem{luo2019convex}
Luo, S., Tai, X.C., Huo, L., Wang, Y., Glowinski, R.: Convex shape prior for
  multi-object segmentation using a single level set function.
\newblock In: Proceedings of the IEEE International Conference on Computer
  Vision, pp. 613--621 (2019)

\bibitem{maes1997multimodality}
Maes, F., Collignon, A., Vandermeulen, D., Marchal, G., Suetens, P.:
  Multimodality image registration by maximization of mutual information.
\newblock IEEE transactions on Medical Imaging \textbf{16}(2), 187--198 (1997)

\bibitem{mangan1999partitioning}
Mangan, A.P., Whitaker, R.T.: Partitioning 3d surface meshes using watershed
  segmentation.
\newblock IEEE Transactions on Visualization and Computer Graphics
  \textbf{5}(4), 308--321 (1999)

\bibitem{modersitzki2009fair}
Modersitzki, J.: FAIR: flexible algorithms for image registration, vol.~6.
\newblock SIAM (2009)

\bibitem{mumford1989optimal}
Mumford, D., Shah, J.: Optimal approximations by piecewise smooth functions and
  associated variational problems.
\newblock Communications on Pure and Applied Mathematics \textbf{42}(5),
  577--685 (1989)

\bibitem{pock2008convex}
Pock, T., Schoenemann, T., Graber, G., Bischof, H., Cremers, D.: A convex
  formulation of continuous multi-label problems.
\newblock In: European Conference on Computer Vision, pp. 792--805. Springer
  (2008)

\bibitem{pohle2001segmentation}
Pohle, R., Toennies, K.D.: Segmentation of medical images using adaptive region
  growing.
\newblock In: Medical Imaging 2001: Image Processing, vol. 4322, pp.
  1337--1346. SPIE (2001)

\bibitem{roberts2019convex}
Roberts, M., Chen, K., Irion, K.L.: A convex geodesic selective model for image
  segmentation.
\newblock Journal of Mathematical Imaging and Vision \textbf{61}(4), 482--503
  (2019)

\bibitem{siu2020image}
Siu, C.Y., Chan, H.L., Ming~Lui, R.L.: Image segmentation with partial
  convexity shape prior using discrete conformality structures.
\newblock SIAM Journal on Imaging Sciences \textbf{13}(4), 2105--2139 (2020)

\bibitem{strekalovskiy2011generalized}
Strekalovskiy, E., Cremers, D.: Generalized ordering constraints for multilabel
  optimization.
\newblock In: 2011 International Conference on Computer Vision, pp. 2619--2626.
  IEEE (2011)

\bibitem{tai2023potts}
Tai, X., Li, L., Bae, E.: The potts model with different piecewise constant
  representations and fast algorithms: a survey.
\newblock In: Handbook of Mathematical Models and Algorithms in Computer Vision
  and Imaging: Mathematical Imaging and Vision, pp. 1--41. Springer (2023)

\bibitem{tai2007image}
Tai, X.C., Christiansen, O., Lin, P., Skj{\ae}laaen, I.: Image segmentation
  using some piecewise constant level set methods with mbo type of projection.
\newblock International Journal of Computer Vision \textbf{73}(1), 61--76
  (2007)

\bibitem{tai2007level}
Tai, X.C., Hodneland, E., Weickert, J., Bukoreshtliev, N.V., Lundervold, A.,
  Gerdes, H.H.: Level set methods for watershed image segmentation.
\newblock In: Scale Space and Variational Methods in Computer Vision: First
  International Conference, SSVM 2007, Ischia, Italy, May 30-June 2, 2007.
  Proceedings 1, pp. 178--190. Springer (2007)

\bibitem{thiruvenkadam2008segmentation}
Thiruvenkadam, S.R., Chan, T.F., Hong, B.W.: Segmentation under occlusions
  using selective shape prior.
\newblock SIAM Journal on Imaging Sciences \textbf{1}(1), 115--142 (2008)

\bibitem{tremeau1997region}
Tremeau, A., Borel, N.: A region growing and merging algorithm to color
  segmentation.
\newblock Pattern recognition \textbf{30}(7), 1191--1203 (1997)

\bibitem{veksler2008star}
Veksler, O.: Star shape prior for graph-cut image segmentation.
\newblock In: European Conference on Computer Vision, pp. 454--467. Springer
  (2008)

\bibitem{wei2018newregion}
Wei, K., Yin, K., Tai, X.C., Chan, T.F.: New region force for variational
  models in image segmentation and high dimensional data clustering.
\newblock Annals of Mathematical Sciences and Applications \textbf{3}(1),
  255--286 (2018)

\bibitem{yan2020convexity}
Yan, S., Tai, X.C., Liu, J., Huang, H.Y.: Convexity shape prior for level
  set-based image segmentation method.
\newblock IEEE Transactions on Image Processing \textbf{29}, 7141--7152 (2020)

\bibitem{yi2012image}
Yi, F., Moon, I.: Image segmentation: A survey of graph-cut methods.
\newblock In: 2012 International Conference on Systems and Informatics
  (ICSAI2012), pp. 1936--1941. IEEE (2012)

\bibitem{Yuan2010}
Yuan, J., Bae, E., Tai, X.C.: A study on continuous max-flow and min-cut
  approaches.
\newblock In: 2010 IEEE Computer Society Conference on Computer Vision and
  Pattern Recognition, pp. 2217--2224. IEEE (2010)

\bibitem{Yuan2010a}
Yuan, J., Bae, E., Tai, X.C., Boykov, Y.: A continuous max-flow approach to
  potts model.
\newblock In: 2010: 11th European Conference on Computer Vision, pp. 379--392.
  Springer (2010)

\bibitem{yuan2014spatially}
Yuan, J., Bae, E., Tai, X.C., Boykov, Y.: A spatially continuous max-flow and
  min-cut framework for binary labeling problems.
\newblock Numerische Mathematik \textbf{126}, 559--587 (2014)

\bibitem{yuan2012efficient}
Yuan, J., Qiu, W., Ukwatta, E., Rajchl, M., Sun, Y., Fenster, A.: An efficient
  convex optimization approach to 3d prostate mri segmentation with generic
  star shape prior.
\newblock In: Proc. Med Image Comput.-Assisted Intervent. Conf. Prostate
  Segment. Challenge 2012, pp. 82--89 (2012)

\bibitem{Yuan2012b}
Yuan, J., Ukwatta, E., Tai, X.C., Fenster, A., Schnoerr, C.: A fast global
  optimization-based approach to evolving contours with generic shape prior.
\newblock CAM report pp. 1--17 (2012)

\bibitem{zaitoun2015survey}
Zaitoun, N.M., Aqel, M.J.: Survey on image segmentation techniques.
\newblock Procedia Computer Science \textbf{65}, 797--806 (2015)

\bibitem{zhang2018novel}
Zhang, D., Chen, K.: A novel diffeomorphic model for image registration and its
  algorithm.
\newblock Journal of Mathematical Imaging and Vision \textbf{60}(8), 1261--1283
  (2018)

\bibitem{zhang20203d}
Zhang, D., Chen, K.: 3d orientation-preserving variational models for accurate
  image registration.
\newblock SIAM Journal on Imaging Sciences \textbf{13}(3), 1653--1691 (2020)

\bibitem{zhang2021topology}
Zhang, D., Lui, L.M.: Topology-preserving 3d image segmentation based on
  hyperelastic regularization.
\newblock Journal of Scientific Computing \textbf{87}(3), 1--33 (2021)

\bibitem{zhang2021topology2}
Zhang, D., Tai, X.C., Lui, L.M.: Topology- and convexity-preserving image
  segmentation based on image registration.
\newblock Applied Mathematical Modelling \textbf{100}, 218--239 (2021)

\bibitem{zhang2015variational}
Zhang, J., Chen, K.: Variational image registration by a total fractional-order
  variation model.
\newblock Journal of Computational Physics \textbf{293}, 442--461 (2015)

\bibitem{zhao1996variational}
Zhao, H.K., Chan, T., Merriman, B., Osher, S.: A variational level set approach
  to multiphase motion.
\newblock Journal of computational physics \textbf{127}(1), 179--195 (1996)

\end{thebibliography}


\appendix
\section{Details of $g$ and $M$}\label{appendix1}
The first and second order derivatives of $H_{\epsilon}(v)$ are $H'_{\epsilon}(v) = \frac{\epsilon}{\pi}\frac{1}{v^2+\epsilon^2}$ and $H''_{\epsilon}(v) = \frac{\epsilon}{\pi}\frac{-2v}{(v^2+\epsilon^2)^2}$, respectively. Set $s_{i,j} = f_{1}(\bm{x}^{i,j})-f_{2}(\bm{x}^{i,j})$ and $k=i+(j-1)\times n$. Then we have
\begin{itemize}
\item $g = g_1+\sigma g_2$, $g_1 = \begin{pmatrix} g_{11} \\g_{12}\end{pmatrix}$, $g_{11},g_{12}\in \mathbb{R}^{n^2\times 1}$, $(g_{11})_{k} = s_{i,j}H'_{\epsilon}(\phi_0(\bm{y}^{i,j}))\partial_{y_{1}} \phi_0(\bm{y}^{i,j})$, $(g_{12})_{k} = s_{i,j}H'_{\epsilon}(\phi_0(\bm{y}^{i,j}))\partial_{y_{2}} \phi_0(\bm{y}^{i,j})$,
$g_2 = \begin{pmatrix} g_{21} \\g_{22}\end{pmatrix}$, $g_{21},g_{22}\in \mathbb{R}^{n^2\times 1}$, $g_{21} = \mathrm{Diag}(v_1)W^{t}S$, $g_{22} = \mathrm{Diag}(v_2)W^{t}S$, $(v_{l})_k = \partial_{y_{l}} \phi_0(\bm{y}^{i,j}), l=1,2$;
\item $M = M_1+\sigma M_2$, $M_1 = \begin{pmatrix} M_{11} & M_{12} \\ M_{12} & M_{13}\end{pmatrix}$, $M_{11}, M_{12}, M_{13}$ are all $n^2\times n^2$ diagonal matrices, $M_2 = \begin{pmatrix} M_{21} & M_{22} \\ M_{22} & M_{23}\end{pmatrix}$, $M_{21}, M_{22}, M_{23}$ are all $n^2\times n^2$ matrices, 
\begin{equation*}
\begin{split}
(M_{11})_{k,k}  &= s_{i,j}H''_{\epsilon}(\phi_0(\bm{y}^{i,j}))\partial _{y_{1}}\phi_0(\bm{y}^{i,j})\partial_{y_{1}}\phi_0(\bm{y}^{i,j})+s_{i,j}H'_{\epsilon}(\phi_0(\bm{y}^{i,j}))\partial_{y_{1}y_1}\phi_0(\bm{y}^{i,j}), \\
(M_{12})_{k,k}  &= s_{i,j}H''_{\epsilon}(\phi_0(\bm{y}^{i,j}))\partial_{y_{1}}\phi_0(\bm{y}^{i,j})\partial_{y_{2}} \phi_0(\bm{y}^{i,j})+s_{i,j}H'_{\epsilon}(\phi_0(\bm{y}^{i,j}))\partial_{y_{1}y_{2}}\phi_0(\bm{y}^{i,j}), \\
(M_{13})_{k,k} & = s_{i,j}H''_{\epsilon}(\phi_0(\bm{y}^{i,j}))\partial_{y_{2}} \phi_0(\bm{y}^{i,j})\partial_{y_{2}}\phi_0(\bm{y}^{i,j})+s_{i,j}H'_{\epsilon}(\phi_0(\bm{y}^{i,j}))\partial_{y_{2}y_{2}}\phi_0(\bm{y}^{i,j}), \\
M_{21} &= \mathrm{Diag}(v_1)W^{t}W\mathrm{Diag}(v_1)+ \mathrm{Diag}(\mathrm{Diag}(u_{1})W^{t}S),\ (u_1)_{k} =  \partial_{y_{1}y_{1}}\phi_0(\bm{y}^{i,j}),\\
M_{22} &= \mathrm{Diag}(v_1)W^{t}W\mathrm{Diag}(v_2)+ \mathrm{Diag}(\mathrm{Diag}(u_{2})W^{t}S),\ (u_2)_{k} =  \partial_{y_{1}y_{2}}\phi_0(\bm{y}^{i,j}),\\
M_{23} &= \mathrm{Diag}(v_2)W^{t}W\mathrm{Diag}(v_2)+ \mathrm{Diag}(\mathrm{Diag}(u_{3})W^{t}S),\ (u_3)_{k} =  \partial_{y_{2}y_{2}}\phi_0(\bm{y}^{i,j}). 
\end{split}
\end{equation*}

\end{itemize}

\section{Modification of $M$}\label{appendix2}
From Appendix \ref{appendix1}, we can see that the property of $M$ depends on the level set function $\phi_0$. In this paper, we mainly consider a simple case, $\phi_{0}(\bm{x}) = r^2-(x_1-b_1)^2-(x_2-b_2)^2$, namely, $\phi_{0}(\bm{x})$ is a level set function of a disk with center $(b_1,b_2)$ and radius $r$. In this case, due to $H'_{\epsilon}(\phi(\bm{y}^{i,j}))>0$, to make $M$ semipositive definite, we do the following modification:
\begin{equation*}
\begin{split}
(M_{11})_{k,k}  &= 4\max(s_{i,j}H''_{\epsilon}(\phi_0(\bm{y}^{i,j})),0)(y_1^{i,j}-b_1)^2+2\max(s_{i,j},0)H'_{\epsilon}(\phi_0(\bm{y}^{i,j})), \\
(M_{12})_{k,k}  &= 4\max(s_{i,j}H''_{\epsilon}(\phi_0(\bm{y}^{i,j})),0)(y_1^{i,j}-b_1)(y_2^{i,j}-b_2), \\
(M_{13})_{k,k}  &= 4\max(s_{i,j}H''_{\epsilon}(\phi_0(\bm{y}^{i,j})),0)(y_2^{i,j}-b_2)^2+2\max(s_{i,j},0)H'_{\epsilon}(\phi_0(\bm{y}^{i,j})), \\ 
M_{21} &= \mathrm{Diag}(v_1)W^{t}W\mathrm{Diag}(v_1),\\
M_{22} &= \mathrm{Diag}(v_1)W^{t}W\mathrm{Diag}(v_2),\\
M_{23} &= \mathrm{Diag}(v_2)W^{t}W\mathrm{Diag}(v_2). 
\end{split}
\end{equation*}

\end{document}